\documentclass[10pt]{article} 
\usepackage[preprint]{tmlr}

\usepackage{microtype}
\usepackage{graphicx}
\usepackage{subcaption}
\usepackage{booktabs} 
\usepackage{multirow}
\usepackage[pagebackref=true]{hyperref}
\usepackage{url}
\usepackage{amssymb}
\usepackage{mathtools}
\usepackage{amsthm}
\usepackage{xcolor}
\usepackage[]{todonotes}
\usepackage{titletoc}

\usepackage[normalem]{ulem}
\useunder{\uline}{\ul}{}

\usepackage{pifont}
\newcommand{\cmark}{\ding{51}}%
\newcommand{\xmark}{\ding{55}}%
\usepackage{xcolor}

\usepackage{amsmath,amsfonts,bm}

\def\eqref#1{equation~\ref{#1}}

\def\1{\bm{1}}

\DeclareMathAlphabet{\mathsfit}{\encodingdefault}{\sfdefault}{m}{sl}
\SetMathAlphabet{\mathsfit}{bold}{\encodingdefault}{\sfdefault}{bx}{n}

\usepackage[capitalize,noabbrev]{cleveref}

\title{ScaGNN: a Graph Neural Network for Multiple Scattering Simulations}

\author{\name Rémi Marsal \email remi.marsal@ensta.fr \\
\addr U2IS, ENSTA, \\ 
Institut Polytechnique de Paris
\AND
\name Stéphanie Chaillat \email stephanie.chaillat@ensta.fr \\
\addr Laboratoire POEMS, CNRS, INRIA, ENSTA, \\ Institut Polytechnique de Paris
\AND
\name Alexandre Chapoutot \email alexandre.chapoutot@ensta.fr\\
\addr U2IS, ENSTA, \\
Institut Polytechnique de Paris
}

\def\month{MM}  
\def\year{YYYY} 
\def\openreview{\url{https://openreview.net/forum?id=XXXX}} 

\begin{document}

\maketitle

\begin{abstract}
The boundary element method (BEM) provides an efficient numerical framework for solving multiple scattering problems in unbounded homogeneous domains.
By restricting the discretization to the domain boundaries, it substantially reduces computational complexity.
The procedure first consists in determining the solution trace on the boundaries of the domain by solving a boundary integral equation.
Then, the volumetric solution can be recovered at low computational cost using a boundary integral representation.
As the first step of the BEM represents the main computational bottleneck, we present ScaGNN, a learning-based approach designed to approximate the solution trace.
It relies on a graph neural network architecture that incorporates a dynamic adaptive edge sampling mechanism for selecting the most relevant interactions to model. 
Guided by intermediate predictions of expected error and edge length, this mechanism selects, at various stages of the forward pass, the most relevant distant interactions to model.
The proposed method is tailored to achieve linear complexity with the number of nodes in the input graph.
To train and evaluate our network, we present a benchmark consisting of several datasets with different types of multiple scattering problems. 
Our experiments show that our approach surpasses existing state-of-the-art learning-based methods on the considered tasks and investigate the generalization capabilities to settings with an increased number of obstacles and out-of-distribution obstacle shapes.
\href{https://github.com/LARIAD/ScaGNN}{github.com/LARIAD/ScaGNN}
\end{abstract}

\section{Introduction}

\begin{figure}[ht]
  \centering
  \includegraphics[width=0.5\linewidth]{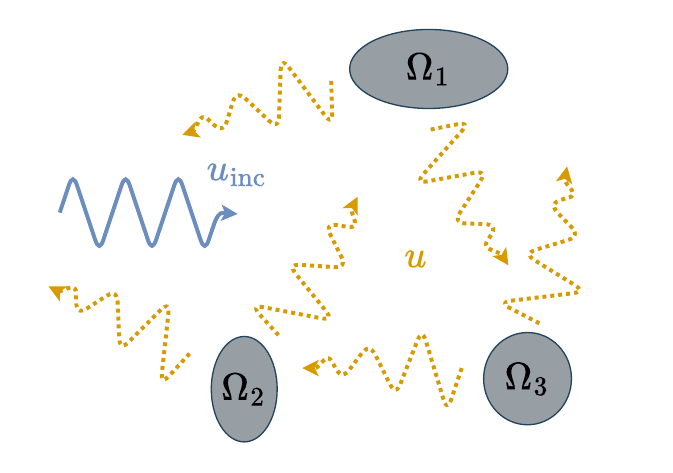}
\caption{Illustration of the resulting field $u$ (dashed arrows) from the scattering of an incoming wave $u_\mathrm{inc}$ (solid arrows) by three obstacles $\Omega_1$, $\Omega_2$ and $\Omega_3$.}
\label{fig:multiple_scattering}
\end{figure}

Multiple scattering problems, i.e., ``the interaction of fields with two or more obstacles''~\citet{martin2006}   \cref{fig:multiple_scattering},  arise in a wide range of applications, including acoustics, electromagnetics, and elasticity. 
The primary challenge in such problems stems from the interplay between the number of obstacles and their separation distances, both of which critically influence the complexity of multiple reflections.

The boundary element method (BEM) \citep{Bonnet1999} is an efficient numerical technique for solving linear partial differential equations (PDEs) such as those involved in multiple scattering. It reformulates the PDE as a boundary integral equation (BIE), with unknowns defined only on the problem boundaries, i.e., the surfaces of the obstacles. The method is decomposed into two steps. The first step is to evaluate the boundary unknowns by solving the BIE. Then, 
the second step consists in reconstructing the solution in the volumetric domain from the BIE solution using the integral representation formula.
By reducing the dimensionality of the problem, the BEM can achieve substantial gains in both computational efficiency and accuracy compared to the finite element method (FEM), especially for wave propagation in unbounded domains. However, despite these advantages, BEMs face significant challenges in multiple scattering problems due to the intricate physical interactions between obstacles, which may lead to a large number of iterations and consequently increased computational costs. These difficulties motivate the development of more efficient iterative strategies, preconditioning techniques or alternative approaches~\cite{Thierry2014}.

Within the BEM framework, the two computational stages have very different costs: solving the BIE on the boundaries is significantly more expensive than evaluating the boundary integral representation to obtain the volumetric solution. To reduce the overall computational cost, a few learning-based approaches have been proposed to simulate multiple scattering \citep{hao2021solving, nair2025multiple}.
Taking advantage of the recent development in neural network architectures for solving PDEs, these approaches rely either on discretizing the solution domain or on neural fields.
Discretization-based methods \citep{pfaff2020learning, zhdanov2025erwin} handle complex geometries and boundary conditions but inherit the computational cost of traditional solvers. 
Neural field approaches \citep{raissi2019physics} represent the solution as a continuous function, allowing inference at arbitrary points, but for complex geometries, they still rely on conditioning over discretized domains \citep{serrano2024aroma, alkin2024universal, alkin2025ab}.
Due to these limitations, learning-based multiple-scattering methods have been restricted to two-dimensional \citep{hao2021solving, nair2025multiple} problems only while our approach deals with 3D problems.

In this article, we propose ScaGNN, a learning-based alternative to traditional BIE solvers, whose estimated solutions are leveraged in boundary integral representations to simulate linear PDEs in the context of multiple scattering.
Learning the boundary solution instead of solving the BIE alleviates the BEM bottleneck while allowing fast evaluation of the volumetric solution in an infinite domain.
Previous methods \citep{lin2021binet,fang2024learning} apply the same principle to tackle problems involving a single continuous boundary. Their extension to multiple scattering configurations is not straightforward because the obstacles are represented by several disconnected boundary components.
To address this issue, we leverage a Graph Neural Network (GNN) that is applied to the obstacle meshes and which models distant interactions between nodes. 
According to the theory underlying the BEM, the dense graph of interactions must be accounted for, leading to a $\mathcal{O}(N^2)$ complexity, with $N$ the number of nodes.
Instead, we propose a dynamic adaptive edge sampling mechanism for creating sparse graphs that capture the relevant interactions, enabling the global approach to achieve linear complexity. This strategy connects nodes based on two criteria: their relative distance, and intermediate predictions of the expected errors, highlighting nodes involved in the strongest interactions.
To train and evaluate our method, we present a new benchmark comprising simulation datasets that focus on 3D exterior scattering problems for both Helmholtz and Laplace problems under Dirichlet or Neumann boundary conditions, as well as meaningful metrics to assess the estimated solution.
Our results show that our ScaGNN approach outperforms existing state-of-the-art learning-based methods designed for solving PDEs. We also investigate its ability to generalize to environments with up to three times as many obstacles per sample as in the training set, and to environments with out-of-distribution obstacle shapes.
To summarize, the main contributions of this work are as follows:

\begin{itemize}
  \item We present a novel learning-based approach for simulating Multiple Scattering phenomena using the BEM framework in which a GNN is leveraged as a surrogate model for replacing the BIE solver. This leads to a reduction of the simulation runtime by two orders of magnitude.
  \item We propose a strategy for selecting the distant interactions to be considered by the GNN, which accounts for both the node distance and intermediate node-wise error predictions. This allows reducing the computational cost, achieving linear complexity.
  \item We introduce a benchmark for training and evaluating multiple scattering surrogate models through three different 3D problems. It includes test sets for assessing generalization with additional obstacles and obstacles with out-of-distribution shapes.
\end{itemize}

\section{Preliminaries on the Boundary Element Method}

This section provides a comprehensive overview of the boundary element method (BEM) \citep{Bonnet1999} through the problem of multiple scattering~\citep{martin2006} of an incident wave $u_\text{inc}$. 
Let $\Omega = \bigcup_{i=1}^n \Omega_i \in \mathbb{R}^3$ be the union of $n$ closed bounded sets $\Omega_i$, $1\leq i \leq n$, representing obstacles that do not intersect.
We assume that each set $\Omega_i$ has a Lipschitz-continuous and piecewise-smooth boundary $\Gamma_i$ and let $\Gamma = \bigcup_{i=1}^n \Gamma_i$.
The homogeneous Helmholtz equation is given by:
\begin{equation}\label{eq:helmholtz}
    \left\{
    \begin{aligned}
         \mathcal{L} u &= 0 && \text{in } \mathbb{R}^3 \setminus \Omega \\
         u &= -u_\text{inc}   && \text{on } \Gamma
    \end{aligned}
    \right.
\end{equation}
where $\mathcal{L} = \Delta + k^2$ with $\Delta$ the Laplace operator and $k$ the wavenumber. We assume the Sommerfeld radiation condition: $\lim_{|\mathbf{x}|\rightarrow \infty} |\mathbf{x}|\big(\frac{\partial}{\partial|\mathbf{x}|} - \mathrm{i}k\big) u(\mathbf{x})=0$ is satisfied, which ensures that no energy is radiated from infinity, with $\mathrm{i}$ the imaginary unit. Therefore, the total field is given by $u_\text{tot}=u+u_\text{inc}$.

The BEM relies on a reformulation of \eqref{eq:helmholtz} as a boundary integral equation (BIE). The key ingredient in this reformulation is the Green's function $G$, defined as the solution of $\mathcal{L}G = \delta$ where $\delta$ denotes the Dirac delta function. 
For our problem, the variational form of the BIE can be defined as follows:
\begin{equation}\label{eq:bie}
    \int_\Gamma u(\mathbf{x}) q(\mathbf{x}) d\mathbf{x} = \int_{\Gamma \times \Gamma} G(\mathbf{x}-\mathbf{y})q(\mathbf{x})p(\mathbf{y})d\mathbf{x}d\mathbf{y}
\end{equation}
\begin{align*}     
      G \colon \mathbb{R}^3 \setminus \Omega \cup \Gamma &\to \mathbb{C}\\
      \mathbf{x} &\mapsto -\frac{e^{-\mathrm{i}kr}}{4\pi r}, \quad \text{with } r=\|\mathbf{x}\|_2,
\end{align*}
where $q$ is a test function and $p$ is the unknown. The solution of this equation gives only the trace of the density $p$ on the boundary $\Gamma$. The second step of the method consists in applying  the boundary integral representation to compute the scattered field in the volume:
\begin{equation}\label{eq:repr}
    u(\mathbf{x}) = \mathcal{S}(p)(\mathbf{x}), \quad \mathbf{x}\in \mathbb{R}^3 \setminus \Omega
\end{equation}
where $\mathcal{S}(p)(\mathbf{x}) = \int_\Gamma G(\mathbf{x}-\mathbf{y})p(\mathbf{y})d\mathbf{y}$ is   the single layer potential operator.

The main computational cost of BEM arises from solving the BIE (\eqref{eq:bie}).
After discretization of $\Gamma$, \eqref{eq:bie} becomes a fully populated linear system, with storage and solution complexities relative to the number of points $N$ on the discretized obstacles of $O(N^2)$ and $O(N^3)$, respectively. 
Instead of direct solvers, iterative solvers like GMRES \citep{saad1986gmres} are often employed for BIEs. 
While fast BEMs such as those accelerated by the Fast Multipole Method (FMM) or hierarchical matrix techniques \citep{Darve2000, Chaillat2008, Chaillat2017} significantly reduce the computational cost per iteration by achieving linear or quasi-linear complexity, they do not, by themselves, resolve the intrinsic difficulties of multiple scattering problems. In particular, the number of iterations required for convergence remains a major bottleneck, especially in the case of strong interactions between obstacles.
Once the solution trace on the boundary is obtained, the volumetric field can be efficiently reconstructed using the boundary integral representation (\eqref{eq:repr}). 
This work presents a machine learning model to predict the boundary solution, specifically tailored to address the challenges of multiple scattering.
This significantly accelerates the most expensive part of the computation while still recovering the full volumetric solution.

\section{Related works}

\subsection{Learning and Boundary Element Method}

Recent machine-learning approaches have sought to replace the solution of the BIE (\eqref{eq:bie}) by a neural-network surrogate, as solving the BIE typically constitutes the most computationally expensive stage of the BEM. Most of these methods rely on conditional neural fields \citep{lin2021binet, lin2023bi, qu2024boundary}. In such models, the inputs consist of 2D coordinates of points on the surface, together with additional information specific to the problem, like surface shape parameters or boundary conditions.
To improve accuracy, \citet{fang2024learning} and \citet{meng2024solving} use a frequency representation for input values with a high variation range.
More complex geometries have also been addressed with GNNs in \citet{wang2025graph}.
However, all approaches discussed here are limited to single-surface problems, whereas this work focuses on scattering by multiple disjoint obstacles, which introduces additional challenges.

\subsection{Learning PDEs on Unstructured Data}

PDEs often require processing unstructured data, either because of complex domains (airfoils, geological formations) or to leverage adaptive meshing \citep{pfaff2020learning}.
Computer-vision-based approaches that are effective with grid-structured data \citep{wang2024cvit, colagrande2025linear} can be adapted to unstructured data with continuous convolutions \citep{ummenhofer2019lagrangian}, for instance.
Other works \citep{li2023fourier, li2023geometry} project the unstructured data onto a regular grid in the latent space before applying Fourier Neural Operator layers \citep{li2020fourier}.
As mentioned in the previous section, neural fields may be leveraged \citep{fang2024learning}, but they are limited by the capacity of the conditioning mechanism to address complex geometries.
Many GNN approaches have been developed \citep{sanchez2020learning, li2018learning, li2020neural, li2020multipole}.
One of the most popular, MeshGraphNet \citep{pfaff2020learning}, has been extended several times to handle multiple resolutions using static offline \citep{fortunato2022multiscale, cao2023efficient, ripken2023multiscale} or adaptive dynamic downsampling \citep{deng2024evomesh}.
However, these approaches maintain a constant latent dimension across scales. Following U-Net \citep{ronneberger2015u}, our architecture expands the dimension of low-resolution representations.
More recently, multiple transformer-based methods have achieved top performance on numerous benchmarks, most of which focus on reducing the quadratic computation complexity of attention.
Thus, several approaches propose efficient attention mechanisms \citep{li2022transformer, hao2023gnot, xiao2023improved}.
Other approaches propose to apply transformers locally. 
In particular, the Transolver architecture \citep{li2022transformer,luo2025transolver++} performs attention on learnable slices of flexible shape of the input data.
Erwin \citep{zhdanov2025erwin} and GOAT \citep{wen2025geometry} cluster the input data in hierarchical balls.
\citet{zhdanov2025erwin} and \citet{janny2023eagle} combine both message-passing and transformers to effectively process information both on local and global scales.
As shown in \citep{zhdanov2025erwin}, attention-based point cloud processing approaches \citep{wu2024point} may be relevant for learning PDEs with unstructured data.
In this article, based on our knowledge of the BEM, we propose an efficient graph neural network architecture to approximate the boundary solution in the context of multiple scattering problems that outperforms other learning-based approaches designed for solving PDEs on unstructured data.

\subsection{Graph Adaptation for GNN}
In the BEM, all pairwise interactions between nodes of the obstacle meshes must be considered, leading to a quadratic computational complexity with respect to the number of nodes.
This implies that our GNN architecture must process dense graphs, where the number of edges grows quadratically with the number of mesh nodes. As a result, the computational cost becomes prohibitive for input meshes containing more than a few thousand nodes, even after downsampling.
Consequently, an efficient edge selection strategy is crucial to retain the most relevant interactions while keeping the graph computationally tractable.

In the GNN literature, many approaches have been proposed to adapt the input graph topology in order to improve performance.
Rewiring strategies have been proposed to mitigate the oversquashing phenomenon by adding edges that alleviate communication bottlenecks in the graph \cite{topping2021understanding, gutteridge2023drew}.
These approaches are not directly applicable in our setting, since nodes are already densely connected by construction.
In contrast, DropEdge \cite{rong2019dropedge} and DropMessage \cite{fang2023dropmessage} dynamically modify the graph topology to reduce oversmoothing and overfitting. However, the edge selection is purely random: edges are independently dropped according to a Bernoulli distribution, which may result in highly unbalanced node connectivity when the dropping rate is large.  
Several GNN methods propose dedicated remeshing techniques tailored to the task at hand.
In MeshGraphNet \citep{pfaff2020learning}, the authors  employ adaptive remeshing strategies based on local remeshing techniques \citep{narain2012adaptive} for cloth simulation. This approach requires a groundtruth sizing field to guide the remeshing process.
To simulate collisions between objects with GNNs, \citet{pfaff2020learning, yu2024learning} create contact edges between nodes from distinct objects when their distance is below a prescribed threshold. 
To adapt edge selection to any task, graph sparsification learning approaches \cite{rathee2021learnt, luo2021learning, saha2023learning, qian2023probabilistically} benefit from recent advances in differentiable sampling of discrete random variables \cite{jang2016categorical, maddison2016concrete}.
These methods learn edge selection probabilities and use them to determine which edges should be retained or removed. However, when applied to dense graphs, they require considering all possible node pairs, resulting in a computational complexity of $\mathcal{O}(N^2)$, where $N$ denotes the number of nodes.
To address the specific challenges posed by dense graphs in BEMs, we introduce a computationally efficient edge sampling strategy based on the edge length and predictions of the expected error, as it highlights nodes involved in strong interactions.

 \section{ScaGNN Method}

\begin{figure*}[ht]
  \includegraphics[width=\textwidth]{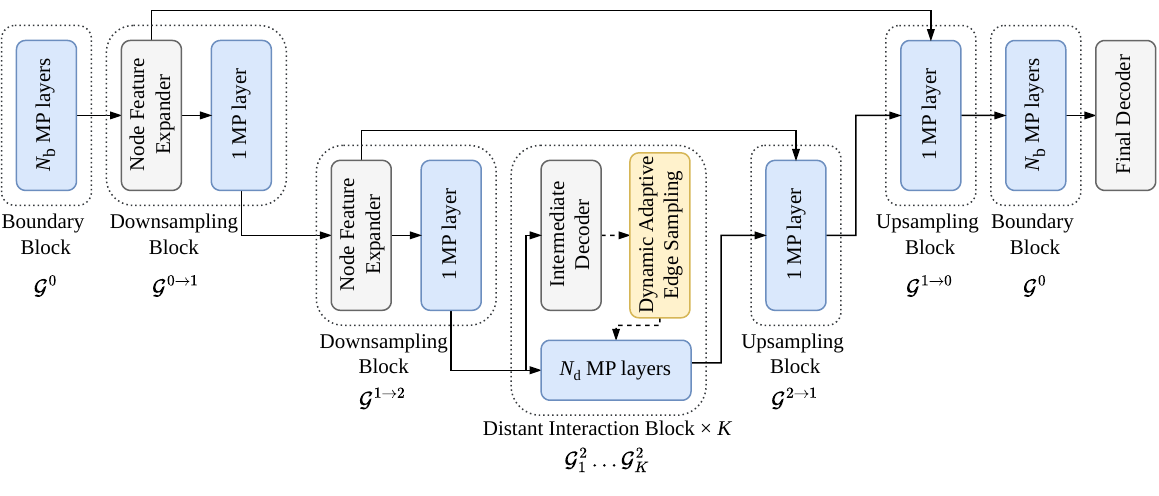}
  \caption{Illustration of ScaGNN architecture with $L=3$ hierarchical levels. The node and edge encoders have not been represented for better readability. The graph representation processed by each MP layer is given below the corresponding block.
  The architecture first consists of a Boundary Block with $N_\mathrm{b}$ MP layers applied to the Boundary Graph $\mathcal{G}^0$, then two Downsampling Blocks. At the lowest resolution, $K$ Distant Interaction Blocks are applied. They are composed of an intermediate decoder whose outputs are used by our dynamic adaptive edge sampling to select the edges of the Distant Interaction Graph $\mathcal{G}^2_1$, ..., $\mathcal{G}^2_{K}$ and $N_\mathrm{d}$ MP layers. Finally, two Upsampling Blocks and a Boundary Block of $N_\mathrm{b}$ MP layers are applied to the Boundary Graph $\mathcal{G}^0$ before the final decoder. Gradient flows through solid arrows, not dashed arrows.}
  \label{fig:benet_3lvl}
\end{figure*}

In this section, we present our ScaGNN method, illustrated in \cref{fig:benet_3lvl}.
In the context of multiple scattering, it is essential to represent both the geometry of obstacles and their interaction accurately.
For these reasons, we adopt a GNN architecture with edge features following MeshGraphNet~\citep{pfaff2020learning}, since it allows expressive modeling of interactions between two nodes.
The dense interaction graph induced by the BEM is computationally prohibitive to process directly. To reduce the number of interactions to model, we leverage a hierarchical GNN inspired by MuS-GNN \citep{lino2022multi}, which aggregates nodes locally, and we introduce a dynamic adaptive edge sampling mechanism to select the most relevant long-range interactions.
In the following, we first define the different graph structures used to represent obstacles, the mappings between successive resolution levels, and the interactions between distant nodes.
We then present the different components of the GNN architecture, including the encoders, processor, and decoders.
Finally, we present the dynamic adaptive edge sampling strategy together with the loss functions used to train the entire framework.

\paragraph{Graphs definition} Let $\mathcal{G}^0 = (V^0, E^0)$ denote the Boundary Graph with nodes $V^0$ and bidirected edges $E^0$. It is composed of M disconnected components  $\mathcal{G}^0_{\Gamma_m} = (V^0_{\Gamma_m}, E^0_{\Gamma_m})$, each corresponding to the mesh of the boundary $\Gamma_m$ of the obstacle $\Omega_m$, $1\leq m\leq M$.
We construct a multiscale hierarchy of $L$ node levels by building an octree for each obstacle. At each level $0<\ell<L$, we obtain a node set $V^\ell$ with $N_\ell$ nodes whose positions form a subset of those at the previous level, i.e., $p(V^\ell) \subset p(V^{\ell-1})$, where $p(V)$ denotes the set of node positions associated with $V$.
We also define directed downsampling edges $E^{\ell-1\rightarrow \ell}$
and directed upsampling edges $E^{\ell\rightarrow \ell-1}$, which connect nodes between consecutive levels $\ell-1$ and $\ell$.
The downsampling edges define the Downsampling Graph
$\mathcal{G}^{\ell-1\rightarrow \ell} = (V^{\ell-1}, E^{\ell-1\rightarrow \ell})$.
Similarly, the upsampling edges define the Upsampling Graph $\mathcal{G}^{\ell\rightarrow \ell-1} = (V^{\ell-1} , E^{\ell\rightarrow \ell-1})$.
To efficiently model long-range interactions between nodes,
we create $K$ Distant Interaction Graph $\mathcal{G}^{L-1}_k = (V^{L-1}, E^{L-1}_k)$, $1\leq k \leq K$, where $E^{L-1}_k$ denotes a set of directed edges.
In contrast to the predefined graph edges, the edges in $E^{L-1}_k$
are selected dynamically during the forward pass using the adaptive edge sampling strategy.
All graphs are associated with node features and edge features.
Examples of these graph structures for two spherical obstacles are shown in~\cref{app:graphs}.

\paragraph{Encoders} 
First, a node encoder initializes the features for all the nodes in $V^0$.
It consists of a two-layer MLP that takes the boundary conditions as input(more details in \cref{app:inputs}) and has an output dimension of $d_0$.
Then, each edge set $E^0$, $E^{\ell-1\rightarrow \ell}$ and $E^{\ell\rightarrow \ell-1}$, $0<\ell<L$, is associated with a dedicated edge encoder that initializes its edge features.
The edge features in each set $E^{L-1}_k$, $1\leq k\leq K$, are initialized successively using a shared edge encoder during the forward pass, after the edge sets have been generated by our dynamic adaptive edge sampling strategy.
The edge encoders are two-layer MLPs with output dimension of  $d_0$ for edges in $E^0$, $d_\ell$ for edges in $E^{\ell-1\rightarrow \ell}$ and $E^{\ell-1\rightarrow \ell}$, and $d_{L-1}$ for edges in $E^{L-1}_k$. They take as inputs a sinusoidal encoding of the edge lengths \citep{vaswani2017attention}, the normalized direction of edges and additional features related to the boundary conditions (more details in \cref{app:inputs}).

\paragraph{Processor} The processor is composed of multiple message-passing (MP) layers that operate on the graphs 
previously defined (see \cref{app:message_passing} for more details on MP layers).
The first step is a Boundary Block which consists of $N_\mathrm{b}$ MP layers that propagate information through
the Boundary Graph $\mathcal{G}^0$.
It is followed by $L-1$ Downsampling Blocks, each associated with a Downsampling Graph $\mathcal{G}^{\ell-1\rightarrow \ell}$, $0<\ell<L$. 
While MuS-GNN \citep{lino2022multi} maintains a fixed latent dimension across scales, our architecture aligns with the U-Net practice \citep{ronneberger2015u} of widening the feature space at coarser levels, allowing richer representations at lower resolutions.
To this end, the node features of $\mathcal{G}^{\ell-1\rightarrow \ell}$ are first initialized by the Node Feature Expander, a single-layer MLP projecting the node features output by the previous Downsampling Blocks (or by the Boundary Block when $\ell-1=0$)
from a dimension $d_{\ell-1}$ to a dimension $d_\ell$.
Then, a MP layer is applied. 
At the lowest resolution, $K$ Distant Interaction Blocks are successively executed on the corresponding Distant Interaction Graphs $\mathcal{G}^{L-1}_k$, $1\leq k \leq K$.
Each Distant Interaction Block is composed of an intermediate decoder, a dynamic adaptive edge sampling mechanism to create the edge set $E^{L-1}_k$, and $N_\mathrm{d}$ MP layers.
The node features of the $k$-th Distant Interaction Block are initialized using the output features of the preceding Distant Interaction Blocks (or the final Downsampling Block when $k=0$).
Thereafter, $L-1$ Upsampling Blocks are applied, each operating on an Upsampling Graph $\mathcal{G}^{\ell\rightarrow \ell-1}$, and composed of a MP layer that reduces the feature dimension from $d_\ell$ to $d_{\ell-1}$.
The node features of $\mathcal{G}^{\ell\rightarrow \ell-1}$ are initialized with the node features output by the preceding Upsampling Block (or by the last Distant Interaction Block if $\ell=L-1$) for the nodes that are also processed by this block, i.e., for the nodes in $V^\ell$.
Otherwise, the remaining node features of $\mathcal{G}^{\ell\rightarrow \ell-1}$, those associated with the nodes $V^{\ell-1} \setminus V^\ell$, are initialized with node features of the Downsampling Graph $\mathcal{G}^{\ell-1\rightarrow \ell}$.
Finally, a Boundary Block of $N_\mathrm{b}$ MP layers is applied to the Boundary  Graph $\mathcal{G}^0$. The node features are initialized from the node features produced by the last Upsampling Block.

\paragraph{Decoders}
Our GNN architecture includes several decoders: $K$ intermediate decoders, one in each Distant Interaction Block, and the final decoder after the last Boundary Block.
The intermediate decoders are two-layer MLPs operating on the node features computed by the last MP layer before each Distant Interaction Block. 
The $k$-th intermediate decoder outputs two vectors $\hat{\mathbf{y}}_{L-1,k}\in \mathbb{R}^{N_{L-1}\times d_\mathrm{f}}$ and $\hat{\mathbf{e}}_k\in \mathbb{R}^{N_{L-1}\times d_\mathrm{f}}$ corresponding to the approximate solution of the BIE (\eqref{eq:bie}) and the expected error associated with this prediction for each node in $V^{L-1}$.
The predicted errors are then used to sample the edges of the different graphs $\mathcal{G}^{L-1}_k$, $1\leq k \leq K$, while the predictions of the BIE solution are leveraged to generate the supervision signals for the error predictions (see the following sections). 
The final decoder is a linear layer that takes the node features of the last message-passing layer as input and outputs the vector $\hat{\mathbf{y}}_0 \in \mathbb{R}^{d_\mathrm{f}\times N_0}$  corresponding to the approximate solution of the BIE (\eqref{eq:bie}) on the boundary discretization of $\Gamma$.
For Laplace problems, $d_\mathrm{f}=1$, and for Helmholtz problems, $d_\mathrm{f}=2$ for the real and imaginary parts of the BIE solution.

\paragraph{Losses}

For each intermediate decoder $1\leq k\leq K$, the predictions $\hat{\mathbf{y}}_{L-1,k}$ are supervised with the groundtruth BIE solution $\mathbf{y}^*_{L-1} \in \mathbb{R}^{d_\mathrm{f}\times N_{L-1}}$ restricted to the low-resolution nodes $V^{L-1}$.
The predictions $\hat{\mathbf{e}}_k$ are trained to match the groundtruth error $\mathbf{e}^*_k \in \mathbb{R}^{d_\mathrm{f}\times N_{L-1}}$ at the intermediate decoder $k$.
It is given by
\begin{equation}
    \mathbf{e}^*_k = |\hat{\mathbf{y}}_{L-1,k} - \mathbf{y}^*_{L-1}|
\end{equation}
where $|\,.|$ is the element-wise absolute value.
The groundtruth error $\mathbf{e}^*_k$ is treated as a fixed target; therefore, gradients are not backpropagated through it.
The predictions $\hat{\mathbf{y}}_0$ of the final decoder are supervised by the groundtruth BIE solution on the discretization of the boundary $\Gamma$,
$\mathbf{y}^*_0 \in \mathbb{R}^{d_\mathrm{f}\times N_0}$.
All predictions (errors and BIE solutions) are compared with their respective groundtruth using the Huber loss $\mathcal{L}_\mathrm{huber}$ \citep{huber1992robust}, and the resulting losses are summed.
In the total loss $\mathcal{L}_\mathrm{total}$, the contributions of the intermediate decoders are weighted with hyperparameter $0<\gamma<1$ so that decoders appearing earlier in the GNN architecture have a smaller impact on the overall loss:
\begin{equation}
    \mathcal{L}_\mathrm{total} = \mathcal{L}_\mathrm{huber}(\hat{\mathbf{y}}_0, \mathbf{y}^*_0) + \sum_{k=0}^{K-1}\gamma^{K - k}\Big(\mathcal{L}_\mathrm{huber}(\hat{\mathbf{y}}_{L-1,k}, \mathbf{y}^*_{L-1}) + \mathcal{L}_\mathrm{huber}(\hat{\mathbf{e}}_k, \mathbf{e}_k^*)\Big).
\end{equation}

\paragraph{Dynamic Adaptive Edge Sampling}\label{sec:edge_sampling}
Since each node interacts with every other node, the number of interactions grows quadratically with the number of nodes. Consequently, processing such dense obstacle graphs with a GNN becomes computationally prohibitive, even after reducing the mesh resolution.
To make our method tractable, we propose a dynamic adaptive edge sampling. It selects a subset of relevant interactions to create the edge sets $E^{L-1}_k$, $1\leq k \leq K$, for each Distant Interaction Graph.
By restricting each node in a graph $\mathcal{G}^{L-1}_k$ to a fixed set of $N_\mathrm{e}$ incoming edges, i.e., each node serves as a destination node only $N_\mathrm{e}$ times, we achieve linear complexity with respect to the number of nodes.
The choice of the source nodes is based on two criteria: the distance to the destination node (i.e., the edge length) and the expected error predicted by the intermediate decoder of the current Distant Interaction Block. 
Let $n_\mathrm{d}$ be a node in $V^{L-1}$ with position  $p_{n_\mathrm{d}}$. To sample each of the $N_\mathrm{e}$ incoming edges of $n_\mathrm{d}$ in
$E^{L-1}_k$, $C$ candidate source nodes $\{n_c\}_{c=1}^C$ are uniformly drawn from $V^{L-1}$. The position of $n_c$ is $p_{n_c}$ and its expected error is $\hat{e}_{k, n_c} \in \mathbb{R}^{d_\mathrm{f}}_+$ (corresponding to the $n_c$-th index of error prediction $\hat{\mathbf{e}}_k$).
The candidate node minimizing the score function $f_\mathrm{score}^\alpha$ based on these two criteria is selected as the source node $n_\mathrm{s}$ for the corresponding edge:
\begin{equation}
    n_\mathrm{s} = \underset{\{n_c\}_{c=1}^C}{\mathrm{argmin}} f_\mathrm{score}^\alpha(p_{n_\mathrm{d}}, p_{n_c}, \hat{e}_{k, n_c})
\end{equation}
where the score function $f_\mathrm{score}^\alpha$ is defined as:
\begin{equation}
    f_\mathrm{score}^\alpha(p_{n_\mathrm{d}}, p_{n_c}, \hat{e}_{k, n_c}) = \frac{\|p_{n_\mathrm{d}} - p_{n_c}\|_2}{(\sum_ {d=1}^{d_\mathrm{f}}\hat{e}_{k, n_c, d})^\alpha}
\end{equation}
with $\hat{e}_{k, n_c, d}$, the $d$-th dimension of $\hat{e}_{k, n_c}$, $1 \leq d \leq d_\mathrm{f}$, and $\alpha \in \mathbb{R}$ is a hyperparameter controlling the relative sensitivity of the score function to the error and the distance between the two nodes.
The predicted errors indicate nodes involved in the most complex or difficult to model interactions (see \cref{app:expected_errors}).
Consequently, the selection of these nodes as source nodes ensures that the message-passing scheme captures the most significant interactions.
However, this criterion remains the same for every destination node $n_\mathrm{d}$.  
By  adding   the edge length, we tailor the criterion to each destination node.
The use of the edge length comes from the expression of Green's functions in the BIE (\eqref{eq:bie}). 
For Laplace and Helmholtz problems, the Green's function are given by $G(\mathbf{x})=-\frac{1}{4\pi r}$ and $G(\mathbf{x})=-\frac{e^{-\mathrm{i}kr}}{4\pi r}$, respectively, where $r=\|\mathbf{x}\|_2$, $\mathbf{x} \in \mathbb{R}^3 \setminus \Omega \cup \Gamma$.
The physical meaning is that interactions between obstacles are stronger at short distances, but long-range interactions remain important.
Importantly, selecting a source node from among $C$ uniformly sampled candidate nodes to generate the edges in the graphs $\mathcal{G}^{L-1}_k$, $1\leq k \leq K$, avoids computing the distance between all pairs of nodes, which significantly reduces the computational cost of our dynamic adaptive edge sampling.

\section{Experiments}
The performance of  ScaGNN is evaluated through extensive experiments on a new benchmark specially designed for this purpose. 
The results of this new method are compared against those of other state-of-the-art learning-based approaches for solving PDEs on unstructured data. This includes GNN-based methods like MeshGraphNet \citep{pfaff2020learning} (hereafter MGN) and MuS-GNN \citep{lino2022multi}, transformer-based methods such as Transolver \citep{wu2024transolver} and Transolver++ \citep{luo2025transolver++}, and a hybrid method that leverages both MP layers and transformers: Erwin \citep{zhdanov2025erwin}.
We also include the point cloud processing method Point Transformer v3 \citep{wu2024point} (hereafter PTv3).

\subsection{New ScaGNN Benchmark}

A new benchmark to evaluate and compare learning-based approaches for multiple scattering problems in unbounded homogeneous domains is proposed. 
We consider three exterior problems with different levels of complexity: a Laplace Dirichlet problem with standard boundary conditions, a Helmholtz Dirichlet problem with an incident wave emitted from a monopole source and a Helmholtz Neumann problem with an incident plane wave.
For each problem, the training set consists of samples containing three ellipsoidal obstacles. Four test sets are additionally considered: three datasets containing three, six, and nine ellipsoidal obstacles per sample, respectively, and one dataset containing three rounded parallelepiped obstacles per sample.
The obstacle locations and shape parameters together with the boundary condition parameters (e.g., the source location or the wavenumber when applicable) vary from one data sample to another.
For each sample, the groundtruth solution on the obstacle boundary is computed using the BEM.
We also introduce metrics dedicated to each problem that measure prediction errors.
Further details about our benchmark are provided in \cref{app:benchmark}. 

\subsection{Implementation Details}

Regarding our architecture, we set the number of levels to $L=3$ and 
the expansion rate in the Node Feature Expander to $2$.
This means that, for $0<\ell<L$, the latent dimension $d_\ell$ at level $\ell$ is given by $d_\ell=d_0 \times 2^{\ell}$. The latent dimension  at the first level is set to $d_0 = 64$.
The number $K$ of low-resolution graphs and Distant Interaction Block is set to 3 unless otherwise mentioned.
In each graph $\mathcal{G}^{L-1}_k$, $1\leq k \leq K$, each node is connected to $N_\mathrm{e}=20$ nodes to create the edge sets $E^{L-1}_k$, unless otherwise mentioned.
For modeling distant interactions, the number $C$ of candidate edges sampled per required edge is set to 2 and the exponent $\alpha$ in the score function $f_\mathrm{score}^\alpha$ is set to $1.0$, unless otherwise specified.
In the loss function $\mathcal{L}_\mathrm{total}$, we set $\gamma$  to $0.4$ and the parameter $\delta$ of all Huber losses is set to $1.0$.
The architecture details of the methods used for comparison are given in \cref{app:implem}.
Their hyperparameters have been tuned both (i) to maximize performance and (ii) to ensure the computational cost and number of parameters is similar or higher than those of our method.  The aim is to show that the observed performance gain can only be attributed to its contributions.
The  models are supervised with the groundtruth BIE solution using a Huber loss with parameter $\delta=1.0$.
Additional implementation details, including the hardware used and the training procedure, are provided in \cref{app:implem}.

\subsection{ScaGNN Performance}\label{sec:id_results}
\begin{table*}[ht]
\centering
\caption{Benchmark results on in-distribution test datasets comparing the prediction errors, number of parameters and computational cost (in FLOPs).}
\label{tab:results}
\resizebox{\textwidth}{!}{%
\begin{tabular}{|cc|c|c|c|c|cc|cc|}
\hline
\multicolumn{2}{|c|}{Architecture Type} &
\multirow{2}{*}{Method} &
\multirow{2}{*}{\begin{tabular}[c]{@{}c@{}}Number of\\ Parameters \end{tabular}} &
\multirow{2}{*}{FLOPs} &
  Laplace &
  \multicolumn{2}{c|}{Helmholtz Dirichlet} &
  \multicolumn{2}{c|}{Helmholtz Neumann} \\
  GNN &
  Transformer &
 &
 &
 &
  $\mathrm{Err}_\mathrm{rel}$ &
  $\mathrm{Err}_\mathrm{ampl}$ &
  $\mathrm{Err}_\mathrm{angle}$ &
  $\mathrm{Err}_\mathrm{ampl}$ &
  $\mathrm{Err}_\mathrm{angle}$ \\ \hline
& \cmark & PTv3 & 19.2 M & 16.8 G & 0.081 & 0.148 & 0.141 & 0.070 & 0.073 \\
& \cmark & Transolver   & 4.6 M & 38.4 G & 0.183 & 0.596 & 0.454 & 0.070 & 0.073 \\
& \cmark & Transolver++ & 4.2 M & 35.6 G & 0.195 & 0.635 & 0.466 & 0.103 & 0.122 \\
\cmark & \cmark & Erwin        & 7.9 M & 42.4 G & 0.119 & 0.218 & 0.175 & 0.074 & 0.076 \\
\cmark & & MGN & 5.3 M & 160.4 G & 0.173 & 0.286 & 0.181 & 0.068 & 0.068 \\
\cmark & & MuS-GNN      & 5.6 M & 91.5 G & 0.281 & 0.282 & 0.183 & 0.068 & 0.068 \\ \hline
\cmark & & Ours ($N_\mathrm{e} = 4$) & 5.6 M & 17.1 G & {\ul 0.070} & {\ul 0.122} & {\ul 0.117} & {\ul 0.055} & {\ul 0.054} \\
\cmark & & Ours ($N_\mathrm{e} = 20$) & 5.6 M & 29.6 G & \textbf{0.045} & \textbf{0.087} & \textbf{0.086} & \textbf{0.044} & \textbf{0.043} \\ \hline
\end{tabular}%
}
\end{table*}

\paragraph{In-distribution results} In \cref{tab:results}, we report performance in terms of prediction errors on the test datasets that follow the same data distribution as the training set, i.e., with three ellipsoidal obstacles per sample.
To demonstrate that the comparison is fair, we also report the number of parameters and the computational cost in FLOPs (see \cref{app:flops} for more details on the FLOPs measurement).
As our approach involves random processes, we report the mean of five evaluations, each with a different seed. We measure an average relative standard deviation of 0.18\% across all datasets and metrics, demonstrating the stability of our method even with a small number $N_\mathrm{e}$ of edges in the Distant Interaction Graph (see \cref{app:std} for the full details of relative standard deviation per dataset).

The results show the superiority of ScaGNN: regardless of the problem or the metric, our new method with $N_\mathrm{e} = 4$ and $N_\mathrm{e} = 20$ (i.e., small or large number of modeled interactions) outperform other learning-based approaches, whether their architecture is based on transformers, GNNs, or both, even if their number of parameters or computational cost far exceed ours.
Thus, our approach with $N_\mathrm{e} = 4$ and $N_\mathrm{e} = 20$ improves the performance of the best baseline, PTv3, by 19\% and 38\%, respectively.
The capacities of fully GNN architectures such as MGN and MuS-GNN are limited by design: since they cannot create edges between disconnected parts of the input graph, they are unable to model interactions between obstacles.
However, MGN and MuS-GNN architectures can sometimes achieve better performance than transformer-based methods.
While Transformers are, in principle, able to capture any interactions between two nodes, the attention mechanism does not allow rich modeling of the interactions between two nodes due to the limited number of attention heads (between 4 and 16). 
In contrast, our approach leverages MP layers to model interactions. Each one is simulated using a MLP with 64 to 256 latent dimensions.
Furthermore, most transformer-based approaches are tailored for PDE benchmarks such as \citep{pfaff2020learning,janny2023eagle} which typically operate on nearly regular meshes spanning the entire computational domain.
This is not the case for the PTv3 architecture, which is specifically designed for processing point clouds.
This may explain why PTv3 achieves the best performance among transformer-based approaches on our benchmark. The irregular distribution of points in our data, consisting of points sampled on the surfaces of distant objects, closely resembles that of point cloud datasets.
However, PTv3's good performance may also be attributed to its larger number of parameters.
In \cref{app:correlations}, we analyze the correlations between the ScaGNN prediction errors and several characteristics of the data samples, including the GMRES iteration count (a proxy for problem difficulty), the wavenumber and the obstacle dispersion.
Finally, qualitative results are given in \cref{app:quali}.

\begin{figure}[ht]
    \centering

    \begin{tabular}{c c c}

        \raisebox{1cm}{\rotatebox{90}{\textbf{Laplace Dirichlet}}} &
        \multicolumn{2}{c}{%
        \centering
        \begin{subfigure}[b]{0.9\textwidth}
            \centering
            \includegraphics[height=5cm]{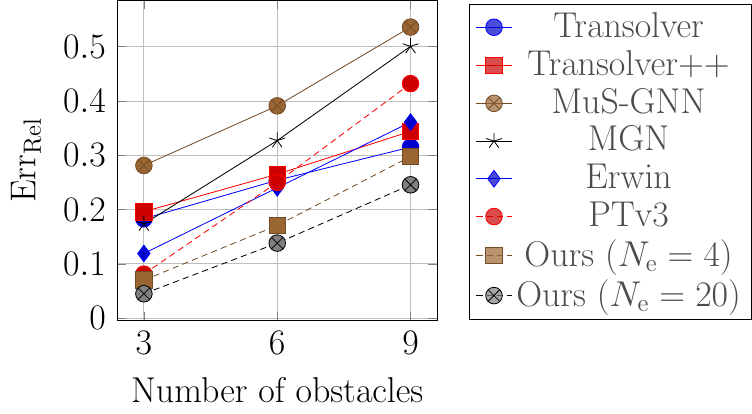}
        \end{subfigure}%
        } \\[12pt]

        \raisebox{1cm}{\rotatebox{90}{\textbf{Helmholtz Dirichlet}}} &
        \begin{subfigure}[b]{0.35\textwidth}
            \centering
            \includegraphics[height=5cm]{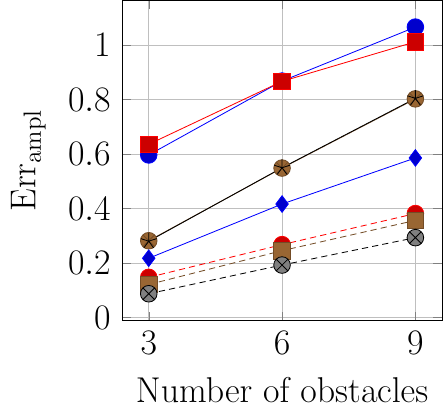}
        \end{subfigure} &
        \begin{subfigure}[b]{0.6\textwidth}
            \centering
            \includegraphics[height=5cm]{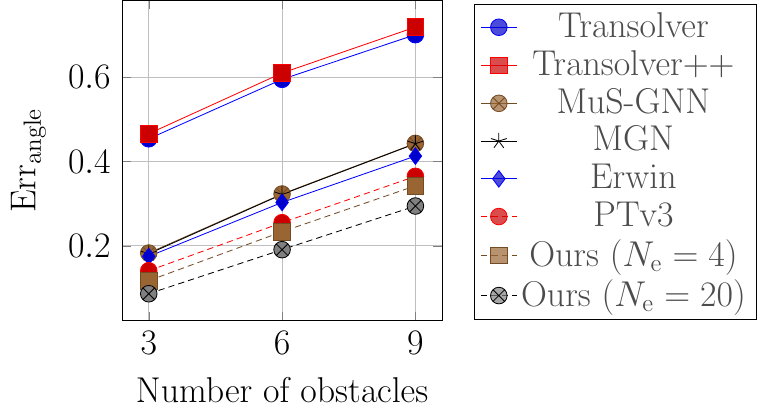}
        \end{subfigure} \\[12pt]

        \raisebox{1cm}{\rotatebox{90}{\textbf{Helmholtz Neumann}}} &
        \begin{subfigure}[b]{0.35\textwidth}
            \centering
            \includegraphics[height=5cm]{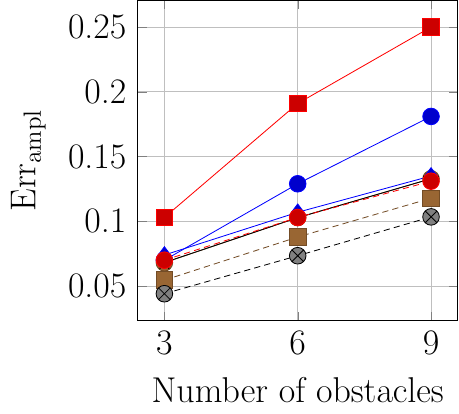}
        \end{subfigure} &
        \begin{subfigure}[b]{0.6\textwidth}
            \centering
            \includegraphics[height=5cm]{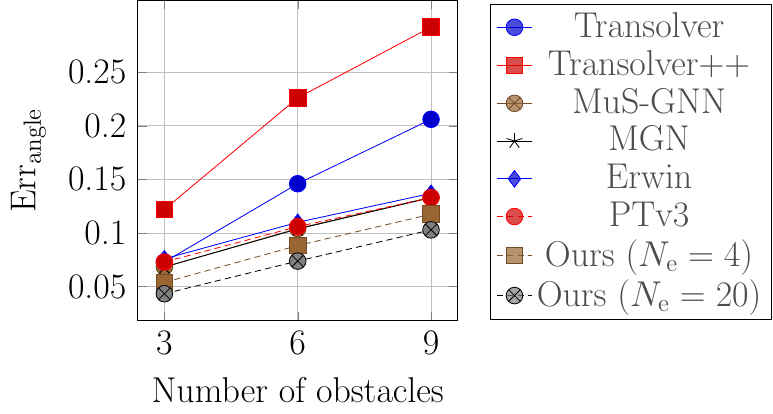}
        \end{subfigure}
    \end{tabular}
    \caption{Estimation errors as a function of the number of obstacles for the Laplace (top), the Helmholtz Dirichlet (middle) and the Helmholtz Neumann (bottom) problems, respectively.}
    \label{fig:generalization}
\end{figure}

\paragraph{Generalization capabilities with more obstacles} In \cref{fig:generalization}, we examine the generalization ability of models trained on datasets containing three obstacles per sample when evaluated on environments containing six and nine obstacles, for the different problems of the ScaGNN benchmark.
Since the number of distant edges per node is fixed to achieve linear scaling with respect to the size of the input mesh, increasing the number of obstacles at test time results in sparser Distant Interaction Graphs.
To continue identifying $N_\mathrm{e}$ relevant interactions per node from a larger set of potential edges, our dynamic adaptive edge sampling must become more selective.
The solution is to increase the number $C$ of candidate edges so as to compare more candidate edges when selecting each edge of a Distant Interaction Graph $\mathcal{G}^{L-1}_k$, $1\leq k\leq K$.
A thorough study of the impact of $C$ on performance for different numbers of obstacles (see \cref{app:num_cand_test}) suggests a heuristic for adjusting the value of $C$. When the number of obstacles is multiplied by $x$, $C$ becomes $C+x-1$.

The results in \cref{fig:generalization} show that performance decreases for all methods with the rise of the number of obstacles. 
This trend is expected due to the quadratic growth in the number of pairwise interactions as the number of obstacles increases, which substantially increases the problem complexity.
In contrast, all models are trained on simpler problems involving three obstacles and are designed to scale linearly with the input size.
Nevertheless, both ScaGNN with $N_\mathrm{e}=4$ and $N_\mathrm{e}=20$ remain superior to the other baselines.
Furthermore, interpolations of the results with linear regressions show that performance reduction remains limited for modest increases in obstacle count. For every new obstacle, ScaGNN prediction errors increase on average by $0.027$ and $0.024$ when $N_\mathrm{e}=4$ and $N_\mathrm{e}=20$, respectively, with a coefficient of determination $R^2 > 0.999$ for both settings. 
The analysis of ScaGNN stability in \cref{app:std} highlights that even when the number of obstacles increases, the relative standard deviations of the metrics remain very low.

\begin{table*}[ht]
\centering
\caption{Benchmark results on out-of-distribution obstacle shapes test sets.}
\label{tab:ood}
\begin{tabular}{|c|c|cc|cc|}
\hline
\multirow{2}{*}{Method} &
  Laplace &
  \multicolumn{2}{c|}{Helmholtz Dirichlet} &
  \multicolumn{2}{c|}{Helmholtz Neumann} \\
  &
  $\mathrm{Err}_\mathrm{rel}$ &
  $\mathrm{Err}_\mathrm{ampl}$ &
  $\mathrm{Err}_\mathrm{angle}$ &
  $\mathrm{Err}_\mathrm{ampl}$ &
  $\mathrm{Err}_\mathrm{angle}$ \\ \hline
PTv3 & 0.166 & 0.203 & 0.165 & 0.071 & 0.098 \\
Transolver   & 0.195 & 0.437 & 0.381 & 0.076 & 0.091 \\
Transolver++ & 0.206 & 0.458 & 0.390 & 0.087 & 0.120 \\
Erwin        & \textbf{0.148} & 0.211 & 0.199 & 0.088 & 0.116 \\
MGN & 0.211 & 0.214 & 0.158 & 0.079 & 0.081 \\
MuS-GNN      & 0.334 & 0.214 & 0.160 & 0.063 & 0.073 \\ \hline
Ours ($N_\mathrm{e} = 4$) & 0.150 & {\ul 0.147} & {\ul 0.120} & {\ul 0.058} & {\ul 0.069} \\
Ours ($N_\mathrm{e} = 20$) &
  {\ul 0.149} &
  \textbf{0.144} &
  \textbf{0.103} &
  \textbf{0.049} &
  \textbf{0.062} \\ \hline
\end{tabular}%
\end{table*}

\paragraph{Generalization capabilities with different obstacle shapes} In \cref{tab:ood}, we measure performance on out-of-distribution obstacle shapes that are parallelepipeds with rounded edges. As expected, we observe a performance degradation compared with  the same settings with ellipsoidal obstacles.  Nevertheless, ScaGNN performs on par with Erwin and remains superior to the other state-of-the-art approaches.

\begin{figure}[ht]
    \centering
    \includegraphics[width=0.6\textwidth]{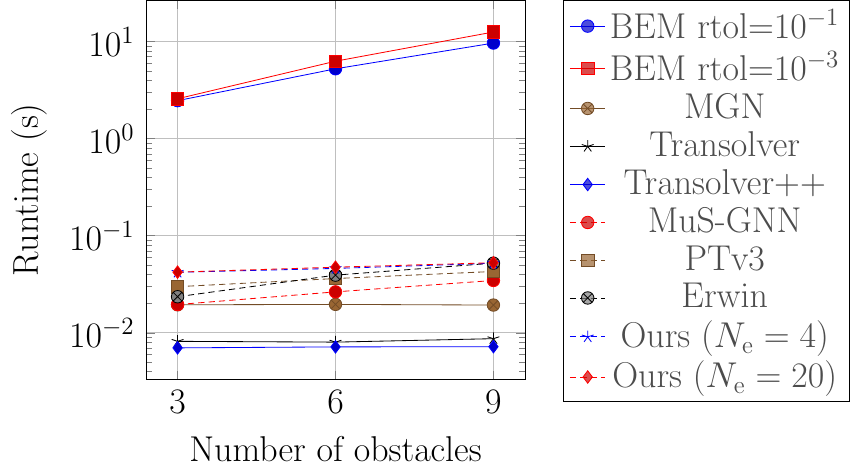}
    \caption{Runtime comparison with respect to the number of obstacles for learning-based methods and for the BEM, considering different convergence tolerance thresholds $\mathrm{rtol}$ for GMRES. The two curves obtained using our method, Ours ($N_e=4$) and Ours ($N_e=20$), overlap.}
    \label{fig:runtime}
\end{figure}

\paragraph{Runtime analysis}
In \cref{fig:runtime}, we study the runtime required to generate the solution traces on the boundaries for the Helmholtz Dirichlet problem with respect to the number of obstacles.
We compare all approaches on the same GPU, including learning-based methods and the BEM with different convergence tolerance thresholds $\mathrm{rtol}$ for GMRES. 
It is worth noting that due to the problem sizes examined in this work (the average number of nodes per data sample with nine obstacles is $N = 12,000$), we do not leverage the FMM-accelerated BEM on GPU.
This follows \citet{gumerov2019gpu}'s conclusion, which reports that for $N \lesssim 20,000$, the standard BEM on GPU remains faster than the FMM-accelerated variant.
With the exception of Transolver and Transolver++, which achieve the lowest runtime due to the small number of learnable slices and layers that we found to provide the best performance, the runtime of the other learning-based methods is similar. 
We observe that our method with $N_e=4$ and $N_e=20$ achieves the same runtime because the same architecture is processed and the extra computational cost when $N_e=20$ is parallelized by the GPU. 
More importantly, this study shows that the BEM is orders of magnitude slower than the learning-based approaches and the BEM's runtime increases faster with the number of obstacles than that of learning-based methods.
The BEM is also sensitive to the wavelength and to the relative position of obstacles, while the runtime of learning-based methods only depends on the number of nodes.
Finally, this study shows that learning-based methods constitute a viable alternative to the BEM in terms of runtime even when the BEM's convergence tolerance thresholds are very low, as in \cref{fig:runtime}.

\subsection{Ablation}

\begin{table}[]
\centering
\caption{Ablation}
\label{tab:ablation}
\resizebox{\textwidth}{!}{%
\begin{tabular}{|cccccc|c|c|cc|}
\hline
  \begin{tabular}[c]{@{}c@{}}Dimension at\\ the finest level\end{tabular} &
  \begin{tabular}[c]{@{}c@{}}Dimension\\ expansion factor\end{tabular} &
  $K$ &
  $N_\mathrm{d}$ &
  \begin{tabular}[c]{@{}c@{}}Intermediate\\ Predictions\end{tabular} &
  \begin{tabular}[c]{@{}c@{}}Edge\\ Sampling\end{tabular} &
  \# params &
  FLOPs &
  $\mathrm{Err}_\mathrm{ampl}$ &
  $\mathrm{Err}_\mathrm{angle}$ \\ \hline
184 & 1 & 1 & 6 & \xmark & Uniform  & 5.6 M & 79.2 G & 0.109          & 0.105          \\
64  & 2 & 1 & 6 & \xmark & Uniform  & 5.2 M & 29.0 G & 0.106          & 0.104          \\
32  & 3 & 1 & 6 & \xmark & Uniform  & 5.8 M & 26.3 G & 0.128          & 0.128          \\
64  & 2 & 3 & 2 & \xmark & Uniform  & 5.2 M & 29.4 G & {\ul 0.100}    & {\ul 0.099}    \\
64  & 2 & 3 & 2 & \cmark & Uniform  & 5.6 M & 29.6 G & {\ul 0.100}    & {\ul 0.099}    \\
64  & 2 & 3 & 2 & \cmark & Adaptive & 5.6 M & 29.6 G & \textbf{0.087} & \textbf{0.086} \\ \hline
\end{tabular}%
}
\end{table}

In \cref{tab:ablation}, we study the impact of the different components of our method for the exterior Helmholtz Dirichlet problem.
In the first three rows, we demonstrate that the computational cost can be significantly reduced by increasing the latent dimension at low resolution, while maintaining similar performance and number of parameters.
Although it is a common deep learning practice \citep{ronneberger2015u}, it is not applied in GNNs for PDE simulation \citep{lino2022multi,deng2024evomesh}.
This justifies the introduction of the Node Feature Expander module in our architecture.
However, we observe that the latent dimension at the finest level and the expansion factor must be carefully tuned to maximize performance for a given budget of parameters and computational cost in FLOPs. 
In our case, we found that a latent dimension of 64 at the finest level and an expansion factor of 2 is optimal.
The fourth row shows that $K > 1$, i.e. using multiple distinct Distant Interaction Graphs $\mathcal{G}^{L-1}_k$, $1\leq k\leq K$, for modeling distant interactions within a single forward pass, improves performance.
This can be explained by the greater diversity of interactions modeled by our GNN.
Moreover, the impact of increasing $K$ on computational cost is minimal; it is only due to the initialization of $K$ times as many edges.
It is worth noting that when K increases, the number of MP layers per Distant Interaction Block is adapted so that the total number of MP layers remains identical.

The last row highlights the gain achieved by selecting edges of the Distant Interaction Graph using our dynamic adaptive edge sampling, which relies on the intermediate predictions.
The slight increase in the number of parameters and in the computational cost is explained by the intermediate decoders introduced to predict the expected errors.
In the penultimate row, the same architecture trained with the intermediate predictions but with uniform sampling, instead of our dynamic adaptive edge sampling, does not improve performance. This demonstrates that the gain with our dynamic adaptive edge sampling is not due to the intermediate predictions; it can only be explained by the selection of more relevant distant edges.
More thorough ablations of hyperparameters are provided in the Appendix, focusing on the impact of the number of edges $N_\mathrm{e}$ per Distant  Interaction Graph (\cref{app:num_edges}), $\alpha$ in the score function $f_\mathrm{score}^\alpha$ (\cref{app:alpha_score}), the number $K$ of the Distant Interaction Graph (\cref{app:num_distant_graph}), and the weight $\gamma$ in the loss $\mathcal{L}_\mathrm{total}$ (\cref{app:gamma_loss}).

\section{Conclusion}

In this article, we present ScaGNN, a new learning-based method for simulating multiple scattering problems.
Such problems are well suited to BEMs, but can become computationally expensive.
Our approach proposes a new GNN architecture to approximate the solution of the BIE. 
In particular, the key contribution is a dynamic adaptive edge sampling mechanism that identifies the most relevant distant interactions to simulate, and creates edges accordingly.
The edge selection is based on two criteria: the edge length and the expected error predicted by intermediate decoders which highlights nodes involved in strong interactions.
By design, our approach achieveq linear complexity with respect to the number of nodes in the input obstacle meshes.
To evaluate the efficiency of this method, we introduce a novel benchmark of multiple scattering simulations including Laplace and Helmholtz problems.
Our results demonstrate that ScaGNN surpasses other state-of-the-art learning-based methods for solving PDEs and processing point clouds. Additionally, ScaGNN generalizes better to settings with a larger number of obstacles and to out-of-distribution obstacle shapes than the other baselines we compare against.
Our study focuses on problems with a small number of obstacles and with simple shapes due to the high cost to generate the training data.
Extending the approach to more challenging problems with a larger number of obstacles and more complex shapes is needed to validate its applicability to practical engineering problems.
To reduce the need for groundtruth, exploring self-supervised pretraining or physics-informed approaches could also be a relevant perspective.



\subsubsection*{Acknowledgments}
This work is supported by Agence de l’Innovation de Défense (AID) via Centre Interdisciplinaire d’Études pour la Défense et la Sécurité (CIEDS), through the APRO project.

\bibliography{main}
\bibliographystyle{tmlr}

\newpage

\appendix
\startcontents[appendices]
\printcontents[appendices]{l}{1}{\section*{Appendix Contents}}

\newpage

\section{Illustrations of Graph Representations}\label{app:graphs}

\begin{figure*}[ht]
\centering
\begin{subfigure}{.49\textwidth}
  \centering
  \includegraphics[width=0.9\linewidth]{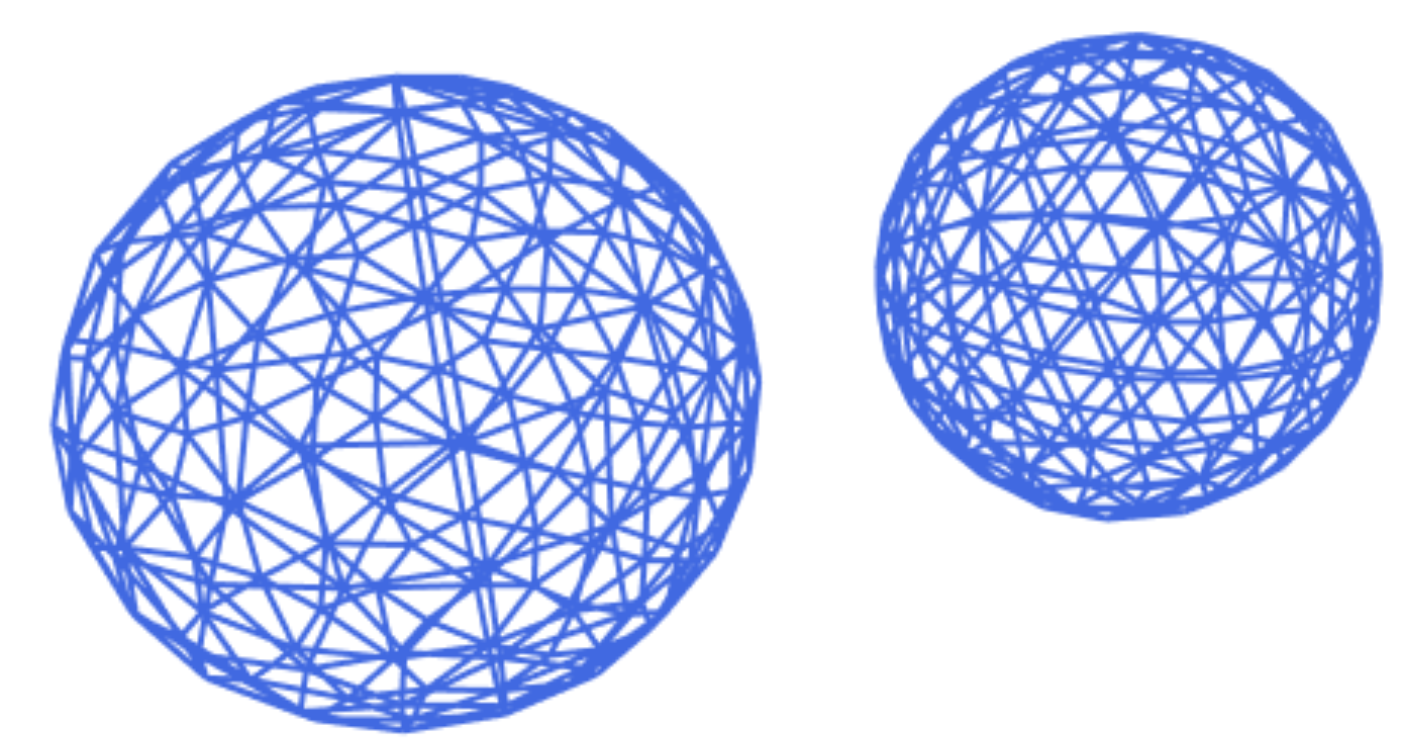}
  \caption*{$\mathcal{G}^0$}
  \label{fig:sub1}
\end{subfigure}%
\begin{subfigure}{.49\textwidth}
  \centering
  \includegraphics[width=0.9\linewidth]{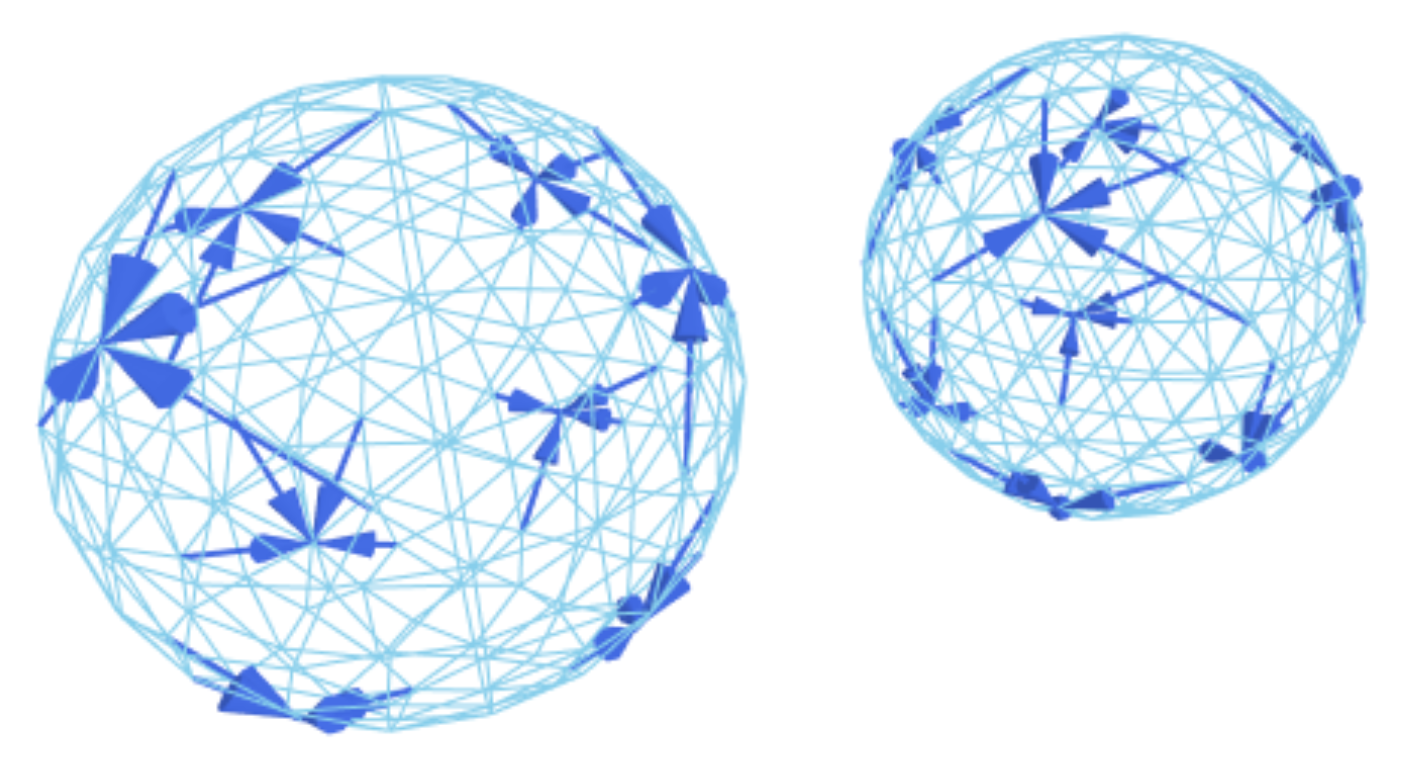}
  \caption*{$\mathcal{G}^{1\rightarrow 2}$}
  \label{fig:sub2}
\end{subfigure}
\begin{subfigure}{.49\textwidth}
  \centering
  \includegraphics[width=0.9\linewidth]{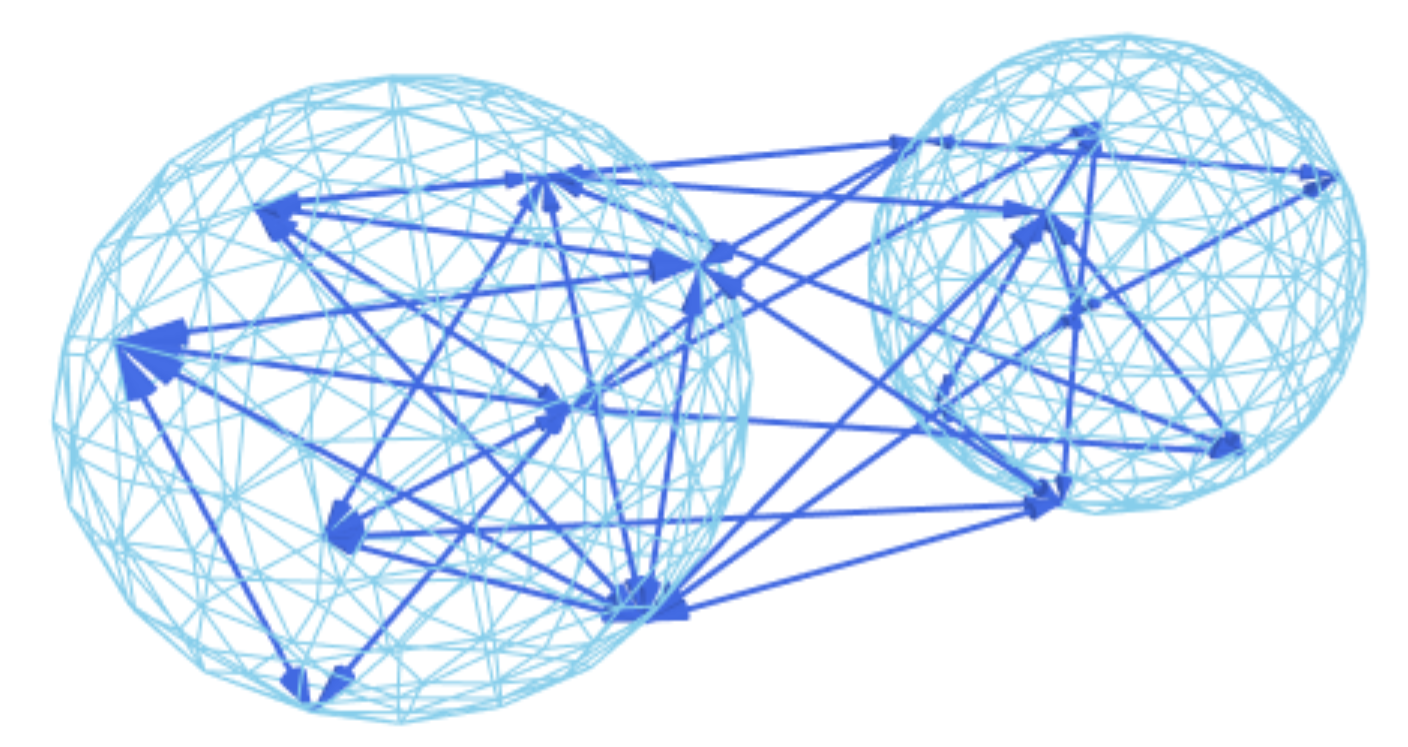}
  \caption*{$\mathcal{G}^2$}
  \label{fig:sub3}
\end{subfigure}
\begin{subfigure}{.49\textwidth}
  \centering
  \includegraphics[width=0.9\linewidth]{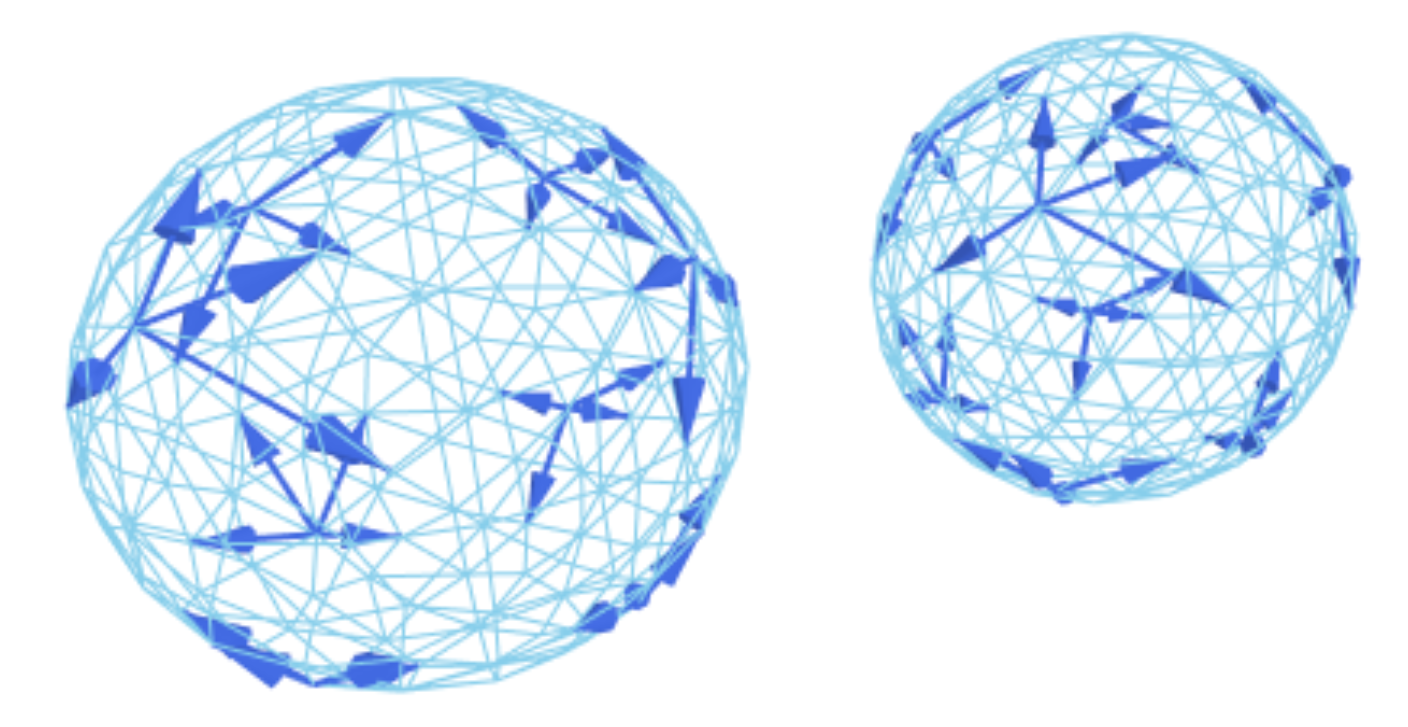}
  \caption*{$\mathcal{G}^{2\rightarrow 1}$}
  \label{fig:sub4}
\end{subfigure}
\caption{Illustration of some directed graph representations used in ScaGNN in the case of two obstacles. 
Graph edges are in dark blue, while the light blue corresponds to the obstacle meshes. $\mathcal{G}^0$: the Boundary Graph (arrows have been omitted for readability), $\mathcal{G}^{1\rightarrow 2}$: the Downsampling Graph from level 1 to 2, $\mathcal{G}^2$: a Distant  Interaction Graph and $\mathcal{G}^{2\rightarrow 1}$: the Upsampling Graph from level 2 to 1. We omit the Downsampling and Upsampling Graphs $\mathcal{G}^{0\rightarrow 1}$ and $\mathcal{G}^{1\rightarrow 0}$ between levels 0 and 1 which are similar to graphs $\mathcal{G}^{1\rightarrow 2}$ and $\mathcal{G}^{2\rightarrow 1}$ at a lower resolution.}
\label{fig:graphs}
\end{figure*}

\Cref{fig:graphs} illustrates of the different types of graph representations used in ScaGNN.

\section{Benchmark details}\label{app:benchmark}

\subsection{Addressed Problems}

This section presents our benchmark in detail. We address the following three multiple scattering problems defined by a PDE and boundary conditions:
\begin{enumerate}
    \item Exterior Laplace Dirichlet problem with standard boundary conditions:
    \begin{equation}
      \begin{cases}
        \Delta u(\mathbf{x}) = 0, \qquad &\mathbf{x} \in \mathbb{R}^3 \setminus \Omega \cup \Gamma \\
        u(\mathbf{x}) = - \Phi_0 -  \frac{\Phi_1}{\|\mathbf{x}-\mathbf{x}_0\|_2} - 2\Phi_2 \mathbf{v} \cdot \frac{\mathbf{x}-\mathbf{x}_0}{\|\mathbf{x}-\mathbf{x}_0\|_2}, \qquad &\mathbf{x} \in \Gamma \\
        u(x)\rightarrow 0, \qquad & \text{as }|x|\rightarrow \infty.
      \end{cases}
    \end{equation}
    where $\Phi_0$, $\Phi_1$ and $\Phi_2$ are three constants between $-1$ and 1 so that $|\Phi_0| + |\Phi_1| + |\Phi_2| = 1$, $\mathbf{x}_0 \in \mathbb{R}^3 \setminus \Omega \cup \Gamma$ and $\mathbf{v}$ is a unit vector in $\mathbb{R}^3$.

    \item Exterior Helmholtz Dirichlet problem with an incident wave of unit amplitude emitted from a monopole source. The Dirichlet boundary condition is therefore parametrized by the source location $\mathbf{x}_0 \in \mathbb{R}^3 \setminus \Omega \cup \Gamma$ and the wavenumber $k$:
    \begin{equation}
      \begin{cases}
        (\Delta + k^2)u(\mathbf{x}) = 0, \qquad &\mathbf{x} \in \mathbb{R}^3 \setminus \Omega \cup \Gamma \\
        u(\mathbf{x}) = -\frac{\mathrm{e}^{\mathrm{i}k\|\mathbf{x}-\mathbf{x}_0\|_2}}{\|\mathbf{x}-\mathbf{x}_0\|_2}, \qquad &\mathbf{x} \in \Gamma \\
        \text{+ Sommerfeld radiation condition.}
      \end{cases}
    \end{equation}
    
    \item Exterior Helmholtz Neumann problem with an incident plane wave of unit amplitude. The Neumann boundary condition is parametrized by the incident wave's direction $\mathbf{v}$ and the wavenumber $k$:
    \begin{equation}
      \begin{cases}
        (\Delta + k^2)u(\mathbf{x}) = 0, \qquad &\mathbf{x} \in \mathbb{R}^3 \setminus \Omega \cup \Gamma \\
        \frac{\partial u}{\partial \mathbf{n}} = - \mathrm{i}k \mathrm{e}^{\mathrm{i}k\mathbf{x}\cdot \mathbf{v}}, \qquad &\mathbf{x} \in \Gamma \\
        \text{+ Sommerfeld radiation condition.}
      \end{cases}
    \end{equation}
    where $\frac{\partial u}{\partial \mathbf{n}} = \nabla u(\mathbf{x}) \cdot \mathbf{n}$ is the normal derivative and $\mathbf{n}$ the normal vector to $\Gamma$ at $\mathbf{x}$.
\end{enumerate}

\subsection{Datasets}

\begin{figure*}
     \centering
     \begin{subfigure}[b]{0.4\textwidth}
         \centering
         \includegraphics[width=\textwidth]{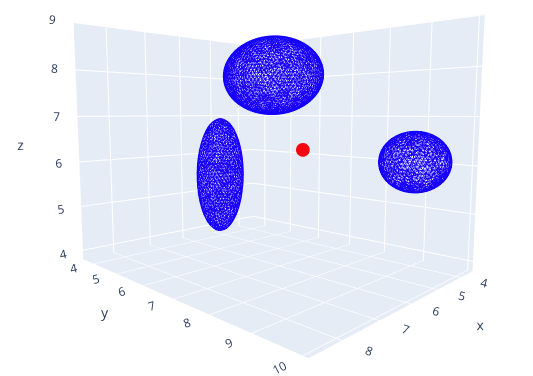}
      \caption{Ellipsoids Obstacles.}
     \end{subfigure}
     \hfill
     \begin{subfigure}[b]{0.5\textwidth}
         \centering
         \includegraphics[width=\textwidth]{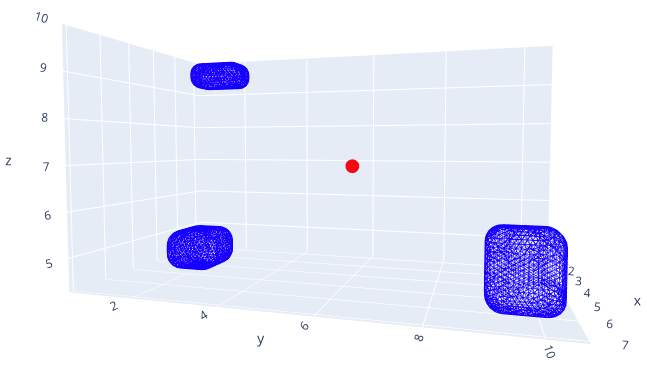}
      \caption{Parallelepipeds Obstacles.}
     \end{subfigure}
    \caption{Illustrations of dataset samples with obstacle meshes in blue and the source in red.}
    \label{fig:data_samples}
\end{figure*}

We generated a training dataset and several test datasets for each problem.
The training set samples contain three ellipsoidal obstacles, and we created test sets with three, six and nine ellipsoidal obstacles, as well as a test set with three rounded parallelepiped obstacles referred to as the OoD test set. General statistics of the training and test sets are given in \cref{tab:data_stats}.
Each dataset sample consists of the meshes of randomly sized and positioned, non-overlapping obstacles, randomly selected boundary condition parameters depending on the problem, and the corresponding boundary solution trace as labels for each mesh node.
The data samples' characteristics are provided in \cref{tab:data_characteristics}.
The meshes representing the obstacles and the trace solution were generated using the GMSH library \citep{geuzaine2009gmsh} and the BEMPP library \citep{bempp}. 
The BIE and the representation formulations used for each problem we address are given in \cref{tab:BIE_formulations}. The BIE solution is computed with GMRES \citep{saad1986gmres} with a convergence tolerance of $10^{-5}$ and double precision.
We also recorded the number of GMRES iterations required to converge for each sample to monitor their computational complexity. We provide an illustration of a dataset sample with three ellipsoids and one with three rounded parallelepipeds in \cref{fig:data_samples}

\begin{table}[ht]
\centering
\caption{General dataset statistics that are common for all problems}
\label{tab:data_stats}
\resizebox{\textwidth}{!}{%
\begin{tabular}{|c|c|cccc|}
\hline
\multirow{2}{*}{Datasets} & \multirow{2}{*}{Training Set} & \multicolumn{4}{c|}{Test Sets} \\
 &  & 3 obstacles & 6 obsatcles & 9 obstacles & OoD \\ \hline
Number of samples & 10k & 1k & 1k & 1k & 1k \\
Number of obstacles per samples & 3 & 3 & 6 & 9 & 3 \\
Total number of nodes & 41M & 4M & 8M & 12M & 3M \\
Obstacle shape & Ellipsoid & Ellipsoid & Ellipsoid & Ellipsoid & Rounded parallelepiped \\ \hline
\end{tabular}%
}
\end{table}

\begin{table}[ht]
\centering
\caption{Main characteristics of our datasets. Lengths are given without units.}
\label{tab:data_characteristics}
\resizebox{\textwidth}{!}{%
\begin{tabular}{cc}
\toprule
Environment size & $10\times 10\times 10$ \\
Edge length in obstacle meshes & 0.1 \\
Minimal distance between two obstacles & 0.1 \\
Ellipses semi-axes length (min -- max) & 0.3 -- 1.5 \\
Wavelength (min -- max) & 0.6 -- 6 \\
Rounded parallelepiped length, width, height (min -- max) & 0.6 -- 3 \\
Rounded parallelepiped rounded radius length (min -- max) & 0.3 -- max(parallelepiped length, width, height) \\
Laplace boundary condition constants $\Phi_0$, $\Phi_1$ and $\Phi_2$ (min -- max) & -1 -- 1, $|\Phi_0| + |\Phi_1| + |\Phi_2| = 1$ \\ 
GMRES rtol & $10^{-5}$ \\
\bottomrule
\end{tabular}%
}
\end{table}

\begin{table}[ht]
\centering
\caption{Formulations of the BIE and of the representations used to generate the dataset for each problem where $\mathrm{S}$ and $\mathrm{N}$ are the single-layer and the hypersingular boundary integral operators, respectively, while $\mathcal{S}$ and $\mathcal{D}$ are the single-layer and the double-layer potential operators, respectively.}
\begin{tabular}{|c|c|c|}
\hline
Problem             & BIE                                                   & Representation   \\ \hline
Laplace Dirichlet   & $\mathrm{S}p=u$                                      & $u=\mathcal{S}p$ \\ \hline
Helmholtz Dirichlet & $\mathrm{S}p=u$                                      & $u=\mathcal{S}p$ \\ \hline
Helmholtz Neumann   & $\mathrm{N}p=\frac{\partial u}{\partial \mathbf{n}}$ & $u=\mathcal{D}p$ \\ \hline
\end{tabular}
\label{tab:BIE_formulations}
\end{table}

\subsection{Metrics}

For evaluation, we assess the performance on the Laplace problem using the relative error of the trace, $\mathrm{Err}_\mathrm{rel}$. 
For the Helmholtz problems, we introduce two metrics: the relative error of the trace amplitude, $\mathrm{Err}_\mathrm{ampl}$, and the absolute error of the trace phase, $\mathrm{Err}_\mathrm{angle}$. 
The definitions of these metrics for a single sample are given by:

\begin{equation}
    \mathrm{Err}_\mathrm{rel} = \frac{\sum_{x \in \Gamma}\big|\hat{p}(x) - p^*(x)\big|}{\sum_{x \in \Gamma}|p^*(x)|}
\end{equation}

\begin{equation}
    \mathrm{Err}_\mathrm{ampl} = \frac{1}{\#\Gamma}\sum_{x \in \Gamma}\bigg|\frac{|\hat{p}(x)| - |p^*(x)|}{|p^*(x)|}\bigg|
\end{equation}

\begin{equation}
    \mathrm{Err}_\mathrm{angle} = \frac{1}{\#\Gamma}\sum_{x \in \Gamma}\mathrm{atan}2(\sin(\Delta p), \cos(\Delta p)), \quad \Delta p = \angle \hat{p}(x) - \angle p^*(x)
\end{equation}

where $p^*$ and $\hat{p}$ denote the ground-truth boundary trace obtained from the BEM and the neural network prediction, respectively, $\#$ indicates the cardinality of a set, and $\angle$ stands for the angle of a complex number.
These metrics are then averaged over all samples in a dataset.

\subsection{Hardware details}

The data was generated on a 36 Intel Core i9-10980XE (3.00GHz) CPUs.
The time required to generate our training datasets with 10000 samples depends on the problem and the BIE formulation. Generating the Laplace Dirichlet, Helmholtz Dirichlet and Helmholtz Neumann training datasets took 12, 36 and 96 hours, respectively. 

\section{Message-Passing}\label{app:message_passing}

MP is applied to  a graph G=(V, E). It consists of a edge feature update \eqref{eq:egde_update} followed by a node feature update \eqref{eq:node_update}:
\begin{equation}\label{eq:egde_update}
    f_{e_{kl}}' = \phi^e(f_{e_{kl}}, f_{v_k}, f_{v_l})
\end{equation}
\begin{equation}\label{eq:node_update}
    f_{v_k}' = \phi^n(f_{v_k}, \sum_k f_{e_{kl}}')
\end{equation}
where $\phi^e$ and $\phi^n$ are MLPs and $f_{e_{kl}}$ are the features of edge $e_{kl} \in E$ connecting node $v_k \in V$ to node $v_l \in V$ of features $f_{v_k}$ and $f_{v_l}$, respectively.

\section{Input details}\label{app:inputs}

\Cref{tab:laplace_inputs,tab:helmholtz_dirichlet_inputs,tab:helmholtz_neumann_inputs} provide the details of the encoder inputs for the Laplace Dirichlet, the Helmholtz Dirichlet and the Helmholtz Neumann problems, respectively. 
It is worth noting that transformer-based baselines require the absolute position of each input node in order to locate them relative to one another.
For this reason, in the Helmholtz Neumann problems with a plane wave as the incident field, the absolute node positions are given with respect to the average node position.
In pure GNN approaches, this information is not mandatory since the relative position between each node is directly encoded through the edge features. However, for the sake of fairness, this input is used with all methods.
For the Laplace Dirichlet and the Helmholtz Dirichlet problems, the node absolute coordinates must be provided regardless of the architecture due to $\mathbf{x}_0$ in the boundary conditions.

\begin{table}[h]
\centering
\caption{Inputs of the node and edge encoders, and their corresponding dimensionalities for the Laplace Dirichlet problems.}
\begin{tabular}{lp{10cm}c}
\toprule
\textbf{Encoder} & \textbf{Input Feature} & \textbf{Dimensions} \\
\midrule
\multirow{5}{*}{Node encoder} 
 & Sinusoidal encoding of the distance between the current node and $\mathbf{x}_0$ & 128 \\
 & Normalized direction of $\mathbf{x}_0$ relative to the current node & 3 \\
 & Normalized direction $\mathbf{v}$ & 3 \\
 & Each term of the boundary condition computed at the current node position & 3 \\
\midrule
\multirow{2}{*}{Edge encoder} 
 & Sinusoidal encoding of the edge length & 128 \\
 & Normalized direction of the edge & 3 \\
\bottomrule
\end{tabular}
\label{tab:laplace_inputs}
\end{table}

\begin{table}[h]
\centering
\caption{Inputs of the node and edge encoders, and their corresponding dimensionalities for the Helmholtz Dirichlet problems.}
\begin{tabular}{lp{10cm}c}
\toprule
\textbf{Encoder} & \textbf{Input Feature} & \textbf{Dimensions} \\
\midrule
\multirow{5}{*}{Node encoder} 
 & Sinusoidal encoding of the distance between the current node and the source $\mathbf{x}_0$ & 128 \\
 & Normalized direction of the source $\mathbf{x}_0$ relative to the current node & 3 \\
 & Wavenumber $k$ of the sample & 1 \\
 & Sine and cosine of the angle of the incident wave with wavenumber $k$ coming from $\mathbf{x}_0$ & 2 \\
\midrule
\multirow{5}{*}{Edge encoder} 
 & Sinusoidal encoding of the edge length & 128 \\
 & Normalized direction of the edge & 3 \\
 & Wavenumber $k$ of the sample & 1 \\
 & Sine and cosine of the angle of an incident wave with wavenumber $k$ at the destination node coming from the source node & 2 \\
\bottomrule
\end{tabular}
\label{tab:helmholtz_dirichlet_inputs}
\end{table}

\begin{table}[h]
\centering
\caption{Inputs of the node and edge encoders, and their corresponding dimensionalities for the Helmholtz Neumann problems.}
\begin{tabular}{lp{10cm}c}
\toprule
\textbf{Encoder} & \textbf{Input Feature} & \textbf{Dimensions} \\
\midrule
\multirow{5}{*}{Node encoder} 
 & Normalized direction of $\mathbf{v}$ of the incident wave & 3 \\
 & Wavenumber $k$ of the sample & 1 \\
 & Sine and cosine of the angle of the incident wave with wavenumber $k$ & 2 \\
 & Sinusoidal encoding of the distance between the current node and the average of node positions & 128 \\
 & Normalized direction of the average node position & 3 \\
\midrule
\multirow{5}{*}{Edge encoder} 
 & Sinusoidal encoding of the edge length & 128 \\
 & Normalized direction of the edge & 3 \\
 & Wavenumber $k$ of the sample & 1 \\
 & Sine and cosine of the angle of an incident wave with wavenumber $k$ at the destination node coming from the source node & 2 \\
\bottomrule
\end{tabular}
\label{tab:helmholtz_neumann_inputs}
\end{table}

\section{Addition Implementation details}\label{app:implem}

The octree partitioning is performed using the implementation of \citet{ripken2023multiscale}. Every two levels are retained to construct the Downsampling and Upsampling Graphs. 

All models are trained from scratch for 100 epochs, with AdamW optimizer \citep{loshchilov2017decoupled}, a batch size of 16, a learning rate starting at $10^{-4}$ decreasing to $10^{-7}$ with a cosine scheduler, and gradient clipping by norm with a maximum value of $1.0$.
Data augmentation is employed by applying the same random rotation to both the input mesh node positions and the source location. 

All training experiments have been conducted on a single NVIDIA RTX 3090 GPU with  24GB memory. Training times range from 4 hours for Point Transformer v3, benefiting from FlashAttention acceleration \cite{dao2022flashattention}, to more than 18 hours for Transolver and Transolver++, due to gradient accumulation, which is required for batch sizes larger than 1. Otherwise, training time is 8 hours for ScaGNN, 11 hours for MuS-GNN, 12 hours for Erwin and 18 hours for MeshGraphNet. Note that FlashAttention cannot be used for training Erwin because its distance-based attention bias position encoding is essential for good performance but is not supported by FlashAttention.

The implementation details of the different state-of-the-art methods we compare against are given in \cref{tab:implem_other_methods}.
The hyperparameters have been tuned to maximize performance on the Helmholtz Dirichlet problem while maintaining similar or higher computational cost and number of parameters compared to our approach.
Due to the nature of the different architectures considered, one cannot simultaneously match the number of parameters and the computational cost of our method.

\begin{table}[ht]
\centering
\caption{Implementation details of the different methods we compare against in this paper.}
\label{tab:implem_other_methods}
\begin{tabular}{lll}
\toprule
\textbf{Model}                & \textbf{Parameter}          & \textbf{Value}               \\ \hline
MeshGraphNet \citep{pfaff2020learning}         & Processor depth    & 15                  \\
                     & Latent dimension   & 128                 \\ \midrule
Point Transformer V3 \citep{wu2024point} & Grid Size          & 0.2                 \\
                     & Encoder Latent dimensions  & (64, 128, 256) \\
                     & Encoder depths      & (2, 2, 6)        \\
                     & Encoder heads      & (4, 8, 16)      \\
                     & Encoder patch size & 2048                \\
                     & Decoder Latent dimensions  & (64, 128) \\
                     & Decoder depths      & (2, 2)           \\
                     & Decoder heads      & (4, 8)          \\
                     & Decoder patch size & 2048                \\
                     & Stride            & 2                   \\ \midrule
Transolver \citep{wu2024transolver}           & Latent dimension   & 256                 \\
                     & Number of layers   & 8                   \\
                     & MPL ratio          & 4                   \\ \midrule
Transolver++ \citep{luo2025transolver++}         & Latent dimension   & 256                 \\
                     & Number of layers   & 8                   \\
                     & MPL ratio          & 4                   \\ \midrule
Erwin \citep{zhdanov2025erwin}                & MPNN dim.          & 64                  \\
                     & Latent dimensions  & (64, 128, 256)      \\
                     & Window sizes        & (512, 512, 512)     \\
                     & Encoder depths      & (2, 2, 6)           \\
                     & Encoder heads      & (4, 8, 16)          \\
                     & Decoder depths      & (2, 2)              \\
                     & Decoder heads      & (4, 8)              \\
                     & Stride            & 2                   \\
                     & Distance-based attention bias            & Enabled \\ \midrule
MuS-GNN \citep{lino2022multi}                & MPNN dim.          & 128                  \\
                     & Scale number  & 3      \\
                     & Number of MPs at each level  & (4, 2, 4)      \\ \bottomrule
\end{tabular}%
\end{table}

\section{Stability Analysis}\label{app:std}

\begin{table*}[ht]
\centering
\caption{Relative standard deviation results over five inferences.}
\label{tab:std}
\begin{tabular}{|c|c|c|cc|cc|}
\hline
\multirow{2}{*}{Method} &
\multirow{2}{*}{Number of obstacles} &
  Laplace &
  \multicolumn{2}{c|}{Helmholtz Dirichlet} &
  \multicolumn{2}{c|}{Helmholtz Neumann} \\
  &
  &
  $\mathrm{Err}_\mathrm{rel}$ &
  $\mathrm{Err}_\mathrm{ampl}$ &
  $\mathrm{Err}_\mathrm{angle}$ &
  $\mathrm{Err}_\mathrm{ampl}$ &
  $\mathrm{Err}_\mathrm{angle}$ \\ \hline
Ours ($N_\mathrm{e} = 4$) & 3 & 0.29\% & 0.20\% & 0.21\% & 0.08\% & 0.12\% \\
Ours ($N_\mathrm{e} = 4$) & 6 & 0.14\% & 0.39\% & 0.11\% & 0.07\% & 0.08\% \\
Ours ($N_\mathrm{e} = 4$) & 9 & 0.18\% & 0.20\% & 0.20\% & 0.04\% & 0.03\% \\ \hline
Ours ($N_\mathrm{e} = 20$) & 3 & 0.31\% & 0.18\% & 0.21\% & 0.10\% & 0.09\% \\
Ours ($N_\mathrm{e} = 20$) & 6 & 0.24\% & 0.15\% & 0.15\% & 0.13\% & 0.09\% \\
Ours ($N_\mathrm{e} = 20$) & 9 & 0.03\% & 0.16\% & 0.07\% & 0.09\% & 0.08\% \\ \hline
\end{tabular}%
\end{table*}

In \cref{tab:std}, we provide the relative standard deviation over five inferences with different seeds to study the stability of our approach. The results show a relative standard deviation always lower than 0.3\% regardless of the problem studied, the number of obstacles or the number of connections per node $N_\mathrm{e}$ in the Distant Interaction Graph. 
This demonstrates the low variability of ScaGNN predictions despite the random processes involved in the dynamic adaptive edge sampling.

\section{FLOPs measurement}\label{app:flops}

FLOPs is measured using Lightning $\mathtt{measure\_flops}$ function \citep{falcon2019pytorch}. 
Point Transformer v3 \citep{wu2024point} enhanced conditional position encoding (xCPE) was re-implemented in native PyTorch \citep{pytorch2024pytorch} so its computational cost can be captured by $\mathtt{measure\_flops}$.
Our native PyTorch version of xCPE was only used for measuring Point Transformer v3 FLOPs.

\begin{figure*}[h]
    \centering
    \includegraphics[width=0.6\textwidth]{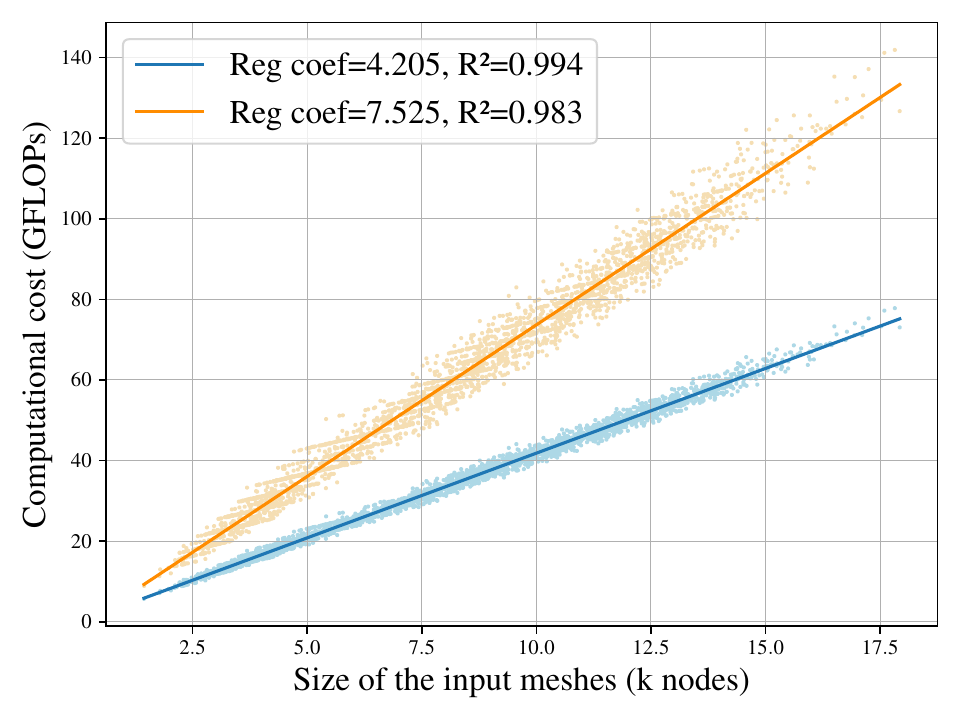}
    \caption{Computational cost of ScaGNN with $N_\mathrm{e} = 4$ (in blue) and $N_\mathrm{e} = 20$ (in orange) as a function of the number of nodes per sample in the test sets with 3, 6 and 9 obstacles.}
    \label{fig:flops}
\end{figure*}

\Cref{fig:flops} shows the computational cost of ScaGNN when $N_\mathrm{e} = 4$ (in blue) and $N_\mathrm{e} = 20$ (in orange) as a function of the number of nodes per sample in the test sets with 3, 6 and 9 obstacles.
Linear regression fits, with coefficients of determination close to 1, confirm the linear relation between the size of the input mesh and ScaGNN computational cost.

\section{Repartition of the Expected Error}\label{app:expected_errors}

\begin{figure*}[h]
     \centering
     \begin{subfigure}[b]{0.32\textwidth}
         \centering
         \includegraphics[width=\textwidth,trim={3cm 1cm 4.5cm 6cm},clip]{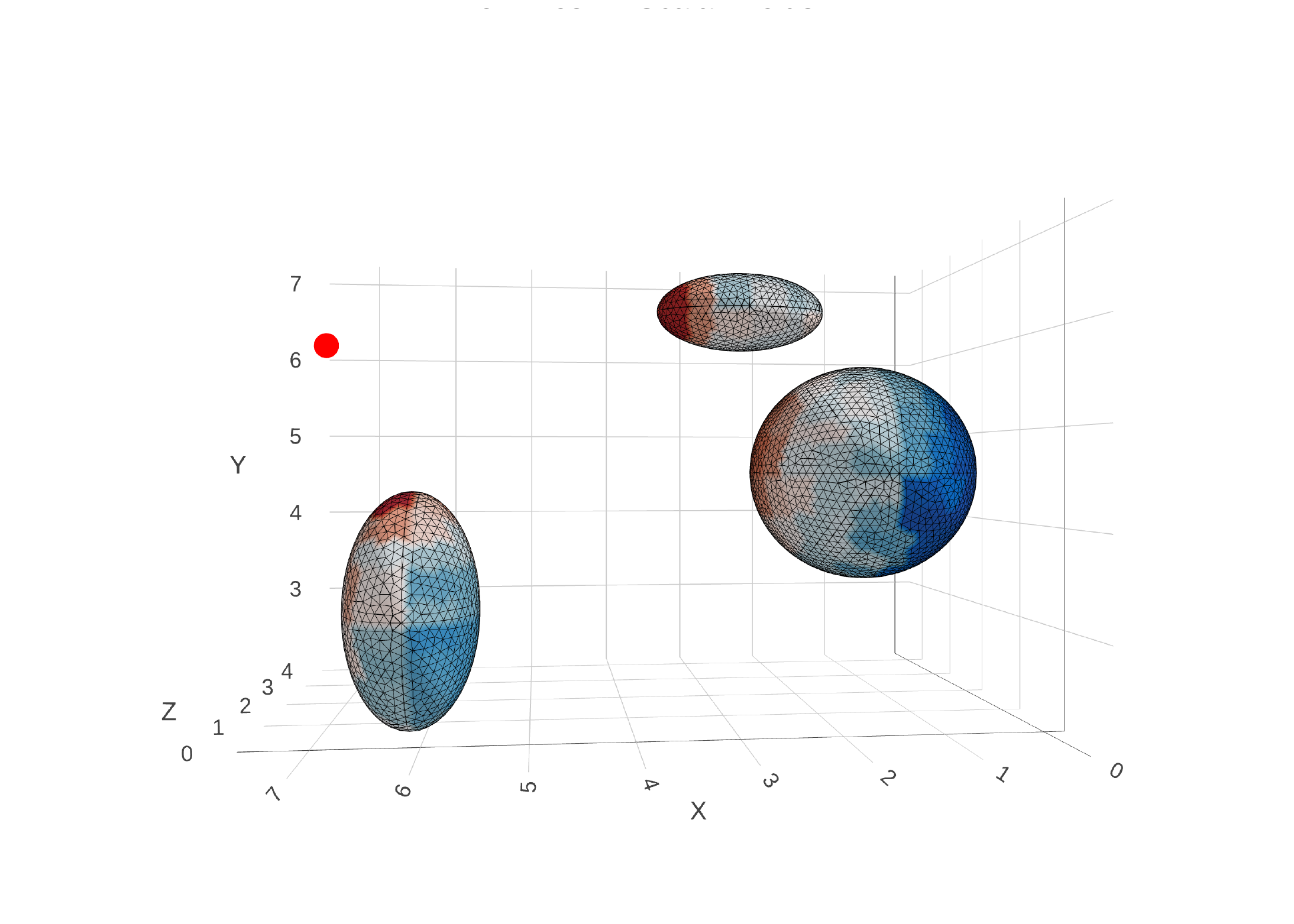}
     \end{subfigure}
     \begin{subfigure}[b]{0.32\textwidth}
         \centering
         \includegraphics[width=\textwidth,trim={3cm 1cm 4.5cm 6cm},clip]{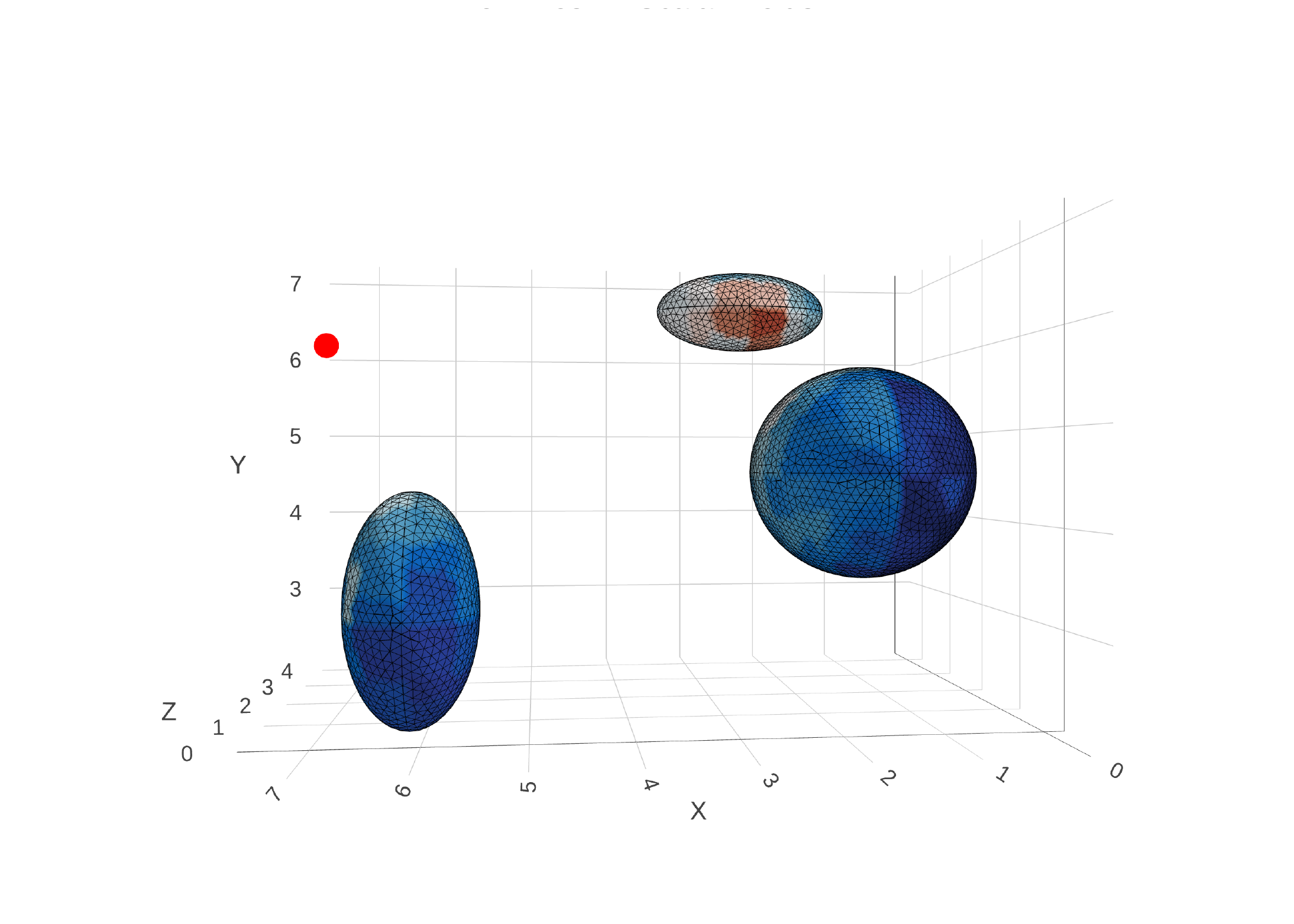}
     \end{subfigure}
     \begin{subfigure}[b]{0.32\textwidth}
         \centering
         \includegraphics[width=\textwidth,trim={3cm 1cm 4.5cm 6cm},clip]{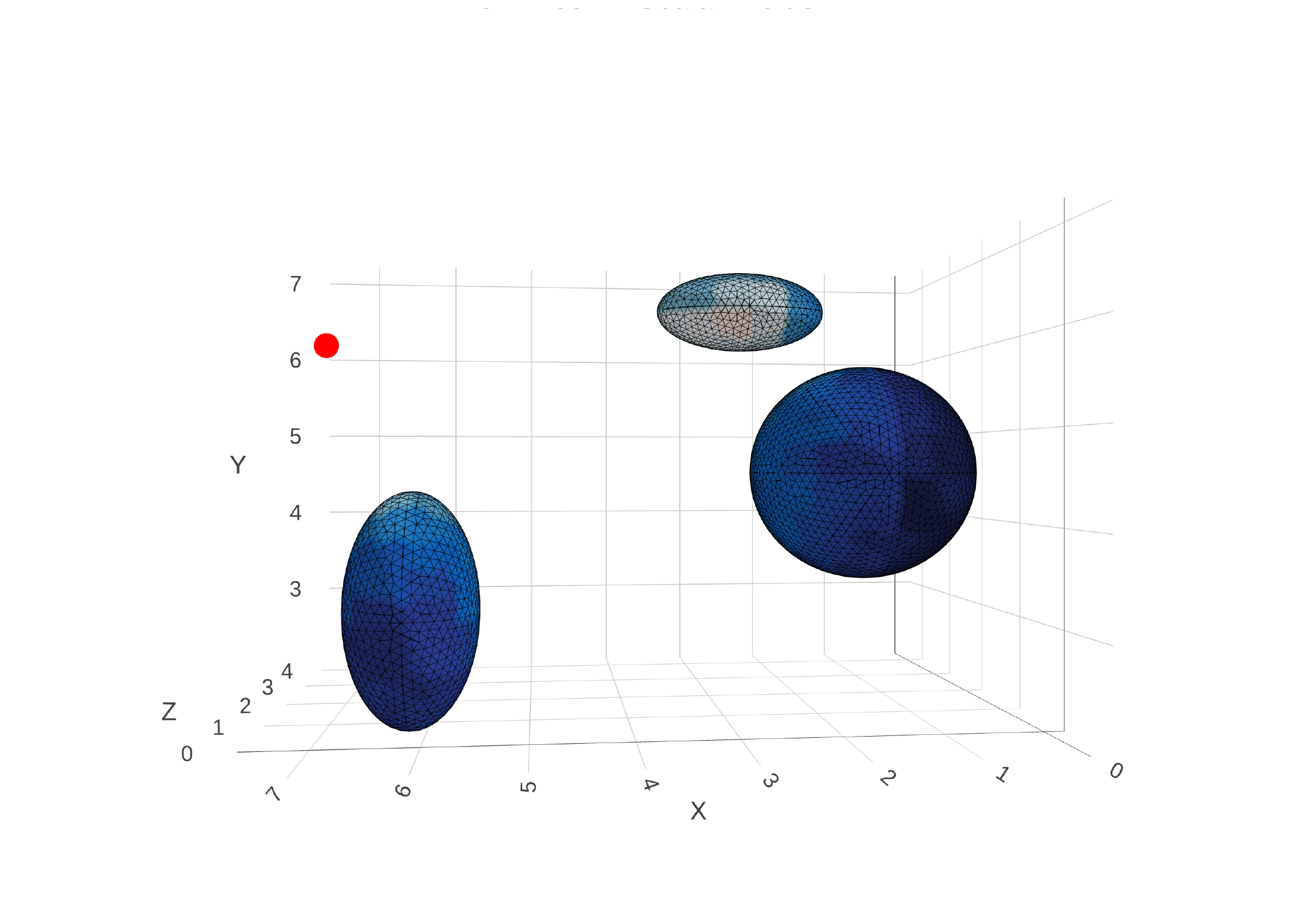}
     \end{subfigure}
     \begin{subfigure}[b]{\textwidth}
         \centering
         \includegraphics[width=0.5\textwidth,trim={0cm 18cm 0cm 0cm},clip]{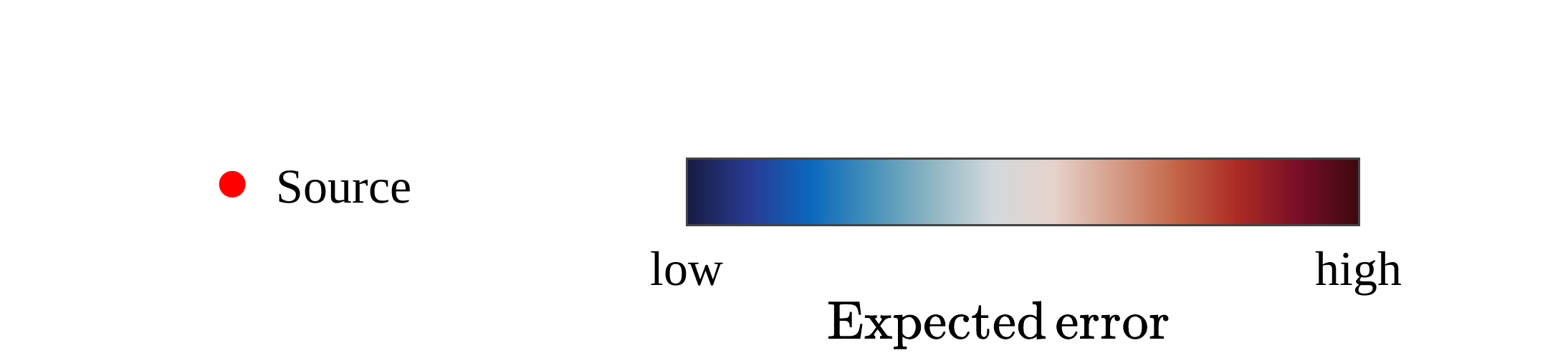}
    \end{subfigure}
    \caption{Expected error predicted by ScaGNN intermediate decoders for a sample from the Helmholtz Dirichlet test set. It is worth noting that the intermediate decoder predictions on the low-resolution graph have been interpolated to the original mesh.}
    \label{fig:expected_errors}
\end{figure*}

In this section, we analyse the repartition of the predicted errors on obstacles for the Helmholtz Dirichlet problem. 
Motivated by the underlying physics, in \cref{tab:expected_errors}, we report the correlations between the predicted errors for each intermediate decoder $1\leq k \leq K$ and different features that are expected to indicate strong or complex interactions. These features include the distance to the monopole source and the distance to the closest node from another obstacle.
The first intermediate predictions ($k=1$) are returned directly after the downsampling part of the GNN architecture when only local interactions have been modeled.
In this situation, no node is aware of the other obstacles, so the intermediate predictions mainly highlight the regions closest to the monopole source, where the prediction errors are largest.
For the subsequent intermediate predictions, Distant Interaction Blocks have already been applied, so each node on an obstacle is aware of the location of the other obstacles.
This information is used to refine error prediction, resulting in higher predicted errors in regions where obstacles are close to each other.
This behavior is also illustrated in \cref{fig:expected_errors}.
Both our quantitative and qualitative studies show that areas with higher predicted errors correspond to the nodes that are more likely to be involved in the strongest or   most complex interactions. 
Consequently, selecting edges based on error predictions favors the connection to such nodes.

\begin{table}[ht]
\centering
\caption{Correlations between the expected error predicted by each Distant Interaction Block and the inverse of the two distances: the source distance and the distance to the closest obstacle.}
\label{tab:expected_errors}
\begin{tabular}{c|ccc}
\begin{tabular}[c]{@{}c@{}} Index of the\\ Distant Interaction Block\end{tabular} & 1 & 2 & 3 \\ \hline
Inverse distance to the source & 0.65 & 0.46 & 0.50 \\
Inverse distance to the closest obstacle & 0.21 & 0.51 & 0.47 \\
\end{tabular}%
\end{table}

\section{Analysis of the model performance}\label{app:correlations}

\begin{table}[ht]
\centering
\caption{Correlations between the log mean absolute error and several dataset sample features.}
\label{tab:correlations}
\begin{tabular}{c|ccc}
 & Laplace & \begin{tabular}[c]{@{}c@{}}Helmholtz\\ Dirichlet\end{tabular} & \begin{tabular}[c]{@{}c@{}}Helmholtz\\ Neumann\end{tabular} \\ \hline
log number of GMRES iterations & -0.15 & 0.80 & 0.76 \\
Log wavenumber & -- & 0.81 & 0.81 \\
Obstacle dispersion & 0.06 & -0.23 & -0.11 \\
\end{tabular}%
\end{table}

In \cref{tab:correlations}, we examine and analyze, for each problem with three ellipsoidal obstacles, the correlation between the performance and various properties of dataset samples, including the log number of GMRES iterations, the log wavenumber, and the obstacle dispersion. 
We quantify dispersion as the maximum of the minimum distances between obstacle pairs for each dataset sample.
For wave propagation problems, the difficulty mainly depends on the wavenumber and the distance between obstacles, which both generate more complex reflections. 
Consequently, we anticipate that ScaGNN performance will decrease in these cases.
This is indeed what we observe: strong correlations between ScaGNN prediction errors and the wavenumber.
Regarding obstacle dispersion, the correlation with ScaGNN prediction errors is lower since most of the error is already explained by the wavenumber, but it is still present.
The aforementioned sources of difficulty for wave problems are also reflected by the number of GMRES iterations to generate the groundtruth solution, so we naturally notice important correlations between the number of GMRES iterations and the ScaGNN performance.
In comparison, ScaGNN prediction errors are weakly correlated with the number of GMRES iterations or with the obstacle dispersion for Laplace problems, which are not wave problems.

\section{Additional Ablations}

\subsection{Impact of the Number of Edge $N_\mathrm{e}$ per Distant  Interaction Graph} \label{app:num_edges}

\begin{figure}[h]
    \centering
    \begin{subfigure}[b]{0.37\textwidth}
        \centering
        \includegraphics[width=\textwidth]{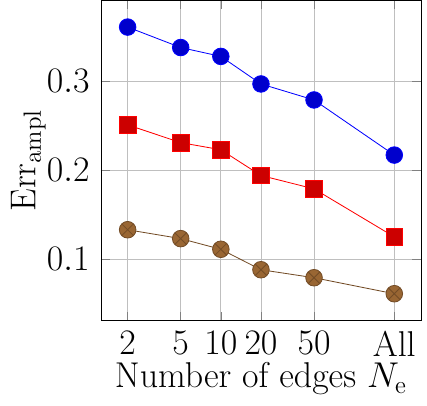}
    \end{subfigure}
   \hfill
    \begin{subfigure}[b]{0.58\textwidth}
        \centering
        \includegraphics[width=\textwidth]{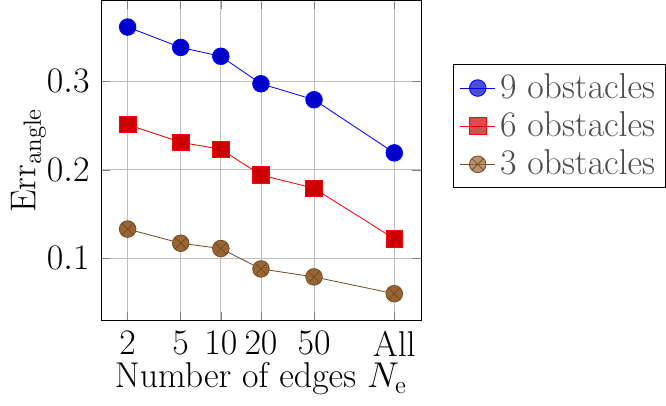}
    \end{subfigure}
    \caption{Estimation errors as a function of the number of edges $N_\mathrm{e}$ per Distant  Interaction Graph.``All'' means that the Distant  Interaction Graph is densely connected.}
    \label{fig:num_edges}
\end{figure}

In \cref{fig:num_edges}, we show how the error decreases when the number of edges per node in the Distant Interaction Graphs grows. We observe that the error reaches a lower bound when the Distant  Interaction Graph is densely connected.

\subsection{Impact of the Number $K$ of Distant Interaction Graph}\label{app:num_distant_graph}

\begin{figure}[h]
    \centering
    \begin{subfigure}[b]{0.37\textwidth}
        \centering
        \includegraphics[width=\textwidth]{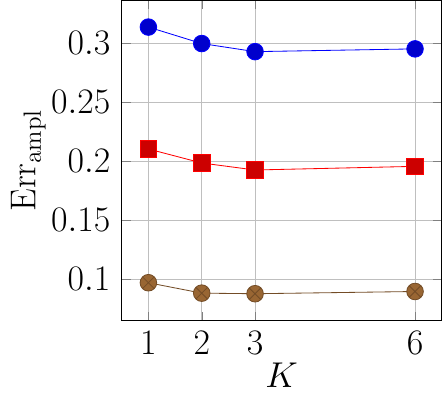}
    \end{subfigure}
   \hfill
    \begin{subfigure}[b]{0.58\textwidth}
        \centering
        \includegraphics[width=\textwidth]{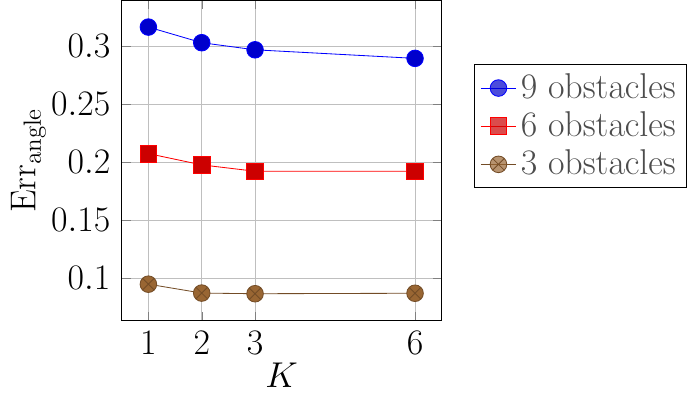}
    \end{subfigure}
    \caption{Estimation errors as a function of the number $K$ of distinct Distant Interaction Graphs processed in our GNN to model distant interactions.}
    \label{fig:num_distant_graphs}
\end{figure}

\Cref{fig:num_distant_graphs} illustrates how the number $K$ of Distant Interaction Graphs affects performance on the Helmholtz Dirichlet problem, with corresponding parameter counts and FLOPs reported in \cref{tab:K_ablation}.
As $K$ varies, we adjust the value of $N_\mathrm{d}$ so that the Distant Interaction Graphs are processed by $K\times N_\mathrm{d} = 6$ message-passing layers, restricting our study to $K \in\{1,2,3,6\}$.
The growth in parameter count and FLOPs with increasing $K$ stems from the initialization of additional edges and from the intermediate predictions.
$K=6$ does not always lead to the best performance since edge features are not carried over from one message-passing layer to the next, as each layer operates on a different Distant Interaction Graph.
Overall, $K=3$ offers a good trade-off between performance gain and the added cost in parameters and FLOPs, and is therefore the value we adopt for our main results.

\begin{table}[h]
\centering
\begin{tabular}{|c|cc|}
\hline
K & Number of parameters & FLOPs  \\ \hline
1 & 5.4 M                & 29.1 G \\
2 & 5.5 M                & 29.3 G \\
3 & 5.6 M                & 29.6 G \\
6 & 6.0 M                & 30.4 G \\ \hline
\end{tabular}%
\caption{Number of parameters and FLOPs for different values of K.}
\label{tab:K_ablation}
\end{table}

\subsection{Impact of the $\alpha$ Parameter in the Score Function $f_\mathrm{score}^\alpha$}\label{app:alpha_score}

\begin{figure}[h]
    \centering
    \begin{subfigure}[b]{0.37\textwidth}
        \centering
        \includegraphics[width=\textwidth]{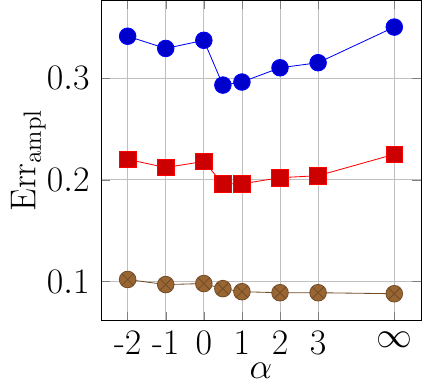}
    \end{subfigure}
   \hfill
    \begin{subfigure}[b]{0.58\textwidth}
        \centering
        \includegraphics[width=\textwidth]{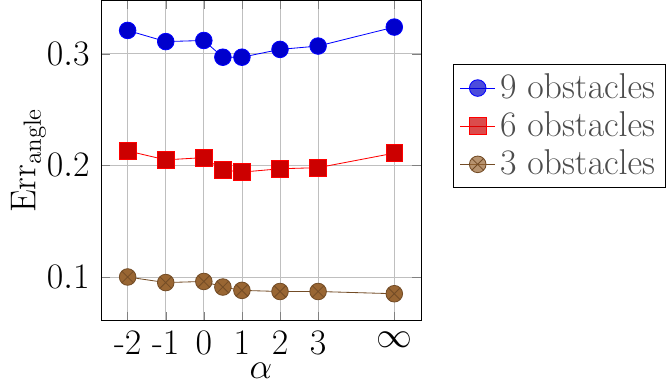}
    \end{subfigure}
    \caption{Estimation errors as a function of the $\alpha$ parameter in the score function $f_\mathrm{score}^\alpha$. When $\alpha= \infty$, the score function only depends on error predictions and the edge length is ignored.}
    \label{fig:alpha_score}
\end{figure}

In \cref{fig:alpha_score}, we study the impact of the $\alpha$ parameter in the score function $f_\mathrm{score}^\alpha$, which controls the relative importance given to the error predictions compared to the edge length when sampling distant edges.
The results for the Helmholtz Dirichlet problem highlight that a negative $\alpha$, i.e., favoring connections from nodes with low predicted error, leads to higher errors. This can be interpreted by the fact that such edges are involved in weak interactions (see \cref{app:expected_errors}), and therefore fail to bring relevant information to the destination node.
When the model is evaluated with the same number of obstacles as in the training set, $\alpha= \infty$ provides the best performance, which means that $f_\mathrm{score}^\alpha$ accounts only for the error predictions and the edge length is ignored.
However, with two or three times as many obstacles as in the training set, a lower value of $\alpha$ improves performance.
For the main results of this paper, we chose $\alpha=1.0$, which strikes a balance between the behaviors mentioned above.

\subsection{Impact of the weight $\gamma$ in the loss $\mathcal{L}_\mathrm{total}$}\label{app:gamma_loss}

\begin{figure}[h]
    \centering
    \begin{subfigure}[b]{0.33\textwidth}
        \centering
        \includegraphics[width=\textwidth]{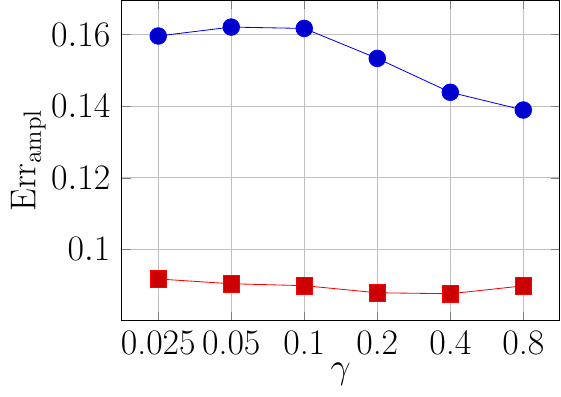}
    \end{subfigure}
   \hfill
    \begin{subfigure}[b]{0.64\textwidth}
        \centering
        \includegraphics[width=\textwidth]{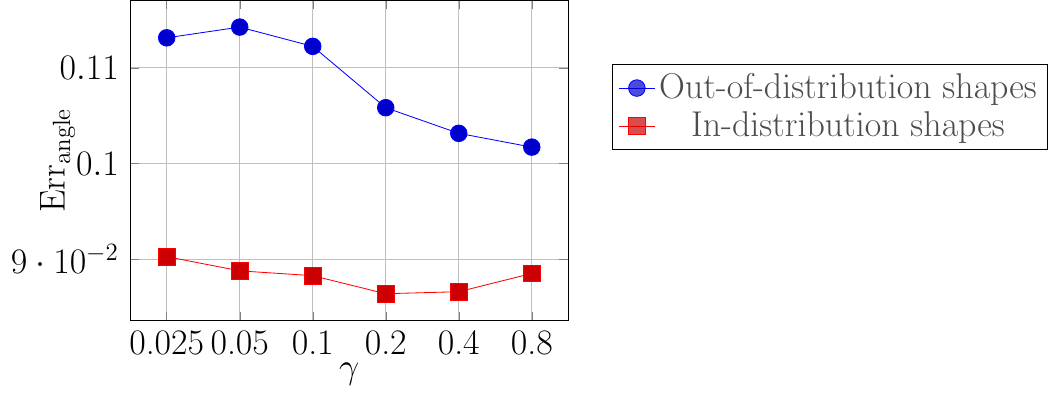}
    \end{subfigure}
    \caption{Estimation errors with three obstacles as a function of the $\gamma$ parameter that weights the intermediate predictions in the loss function $\mathcal{L}_\mathrm{total}$.}
    \label{fig:gamma_loss}
\end{figure}

\Cref{fig:gamma_loss} show the impact of the $\gamma$ parameter in the loss function $\mathcal{L}_\mathrm{total}$, which governs the weighting of the intermediate predictions.
The results for the Helmholtz Dirichlet problem with three obstacles highlight that $\gamma=0.2$ and $\gamma=0.4$ are optimal when the model is evaluated on in-distribution obstacle shapes.
Since we also observe lower error when $\gamma=0.4$ on out-of-distribution shapes, we keep this value for the main results of this work.

\subsection{Impact of the Number $C$ of Candidate Edges during Training}\label{app:num_cand_train}

\begin{figure}[h]
    \centering
    \begin{subfigure}[b]{0.45\textwidth}
        \centering
        \includegraphics[width=\textwidth]{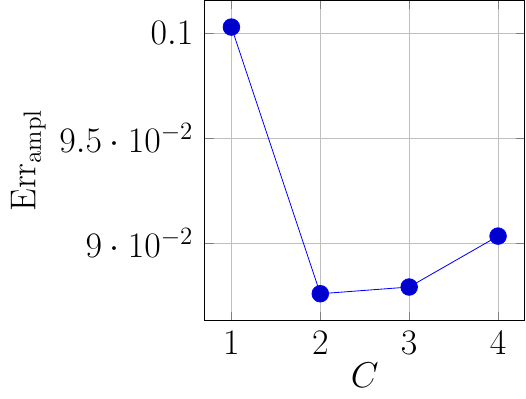}
    \end{subfigure}
   \hfill
    \begin{subfigure}[b]{0.45\textwidth}
        \centering
        \includegraphics[width=\textwidth]{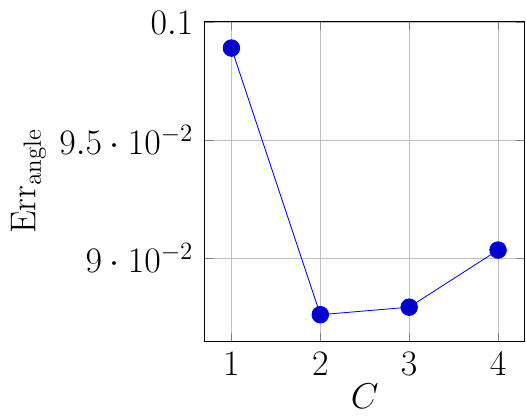}
    \end{subfigure}
    \caption{Estimation errors with three obstacles as a function of the number $C$ of candidate edges from which each edge in a Distant Interaction Graph $\mathcal{G}^{L-1}_k$, $1\leq k\leq K$, is selected. More specifically, $C$ varies during both training and testing.}
    \label{fig:num_cand_train}
\end{figure}

In \cref{fig:num_cand_train}, we examine the effect of varying the number $C$ of candidate edges from which each edge of a Distant Interaction Graph $\mathcal{G}^{L-1}_k$, $1\leq k\leq K$, is selected during training. We obtain optimal results with $C=2$.

\subsection{Scaling the Number $C$ of Candidate Nodes with the Number of Obstacles at Test Time}\label{app:num_cand_test}

\begin{figure}[h]
    \centering
    \begin{subfigure}[b]{0.37\textwidth}
        \centering
        \includegraphics[width=\textwidth]{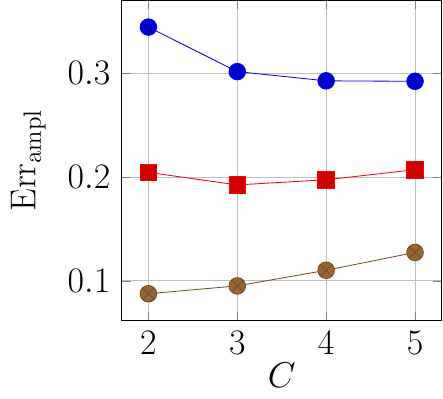}
    \end{subfigure}
   \hfill
    \begin{subfigure}[b]{0.58\textwidth}
        \centering
        \includegraphics[width=\textwidth]{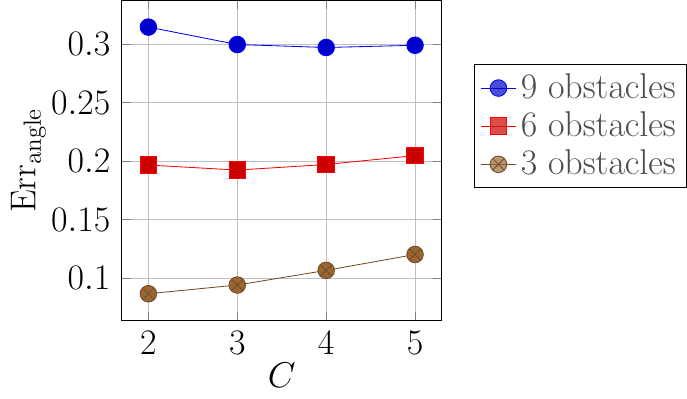}
    \end{subfigure}
    \caption{Estimation errors as a function of the number $C$ of candidate edges for selecting each edge in the Distant Interaction Graph $\mathcal{G}^{L-1}_k$, $1\leq k\leq K$. Here, $C$ varies at test time while the evaluated model has been trained with $C=2$.}
    \label{fig:num_cand_test}
\end{figure}

As the number of interactions grows quadratically with the number of nodes but the number of edges per node in the Distant Interaction Graph $\mathcal{G}^{L-1}_k$, $1\leq k\leq K$, is fixed, edge selection in our adaptive edge sampling must become more stringent.
To ensure that the most relevant edges are retained, we increase the number of candidate edges $C$ as the number of obstacles increases.
\Cref{fig:num_cand_test} illustrates that when our ScaGNN method is evaluated with more obstacles than seen in the training set, performance can be optimized by increasing the number $C$ of candidate edges at test time while the model has been trained with $C=2$. 
This observation motivates a simple heuristic for scaling $C$ with the number of obstacles: when the obstacle count is multiplied by a factor $x$, $C$ should be increased by $x-1$.

\section{Qualitative results}\label{app:quali}

\subsection{Boundary solution}

\begin{figure}[htbp]
    \centering

    \begin{tabular}{c c c}
        &
        \textbf{Intensity} &
        \textbf{Relative error} \\[4pt]

        \raisebox{0.5cm}{\rotatebox{90}{\textbf{Point Transformer v3}}} &
        \begin{subfigure}[b]{0.38\textwidth}
            \centering
            \includegraphics[width=\linewidth,trim={3cm 1cm 4.5cm 5cm},clip]{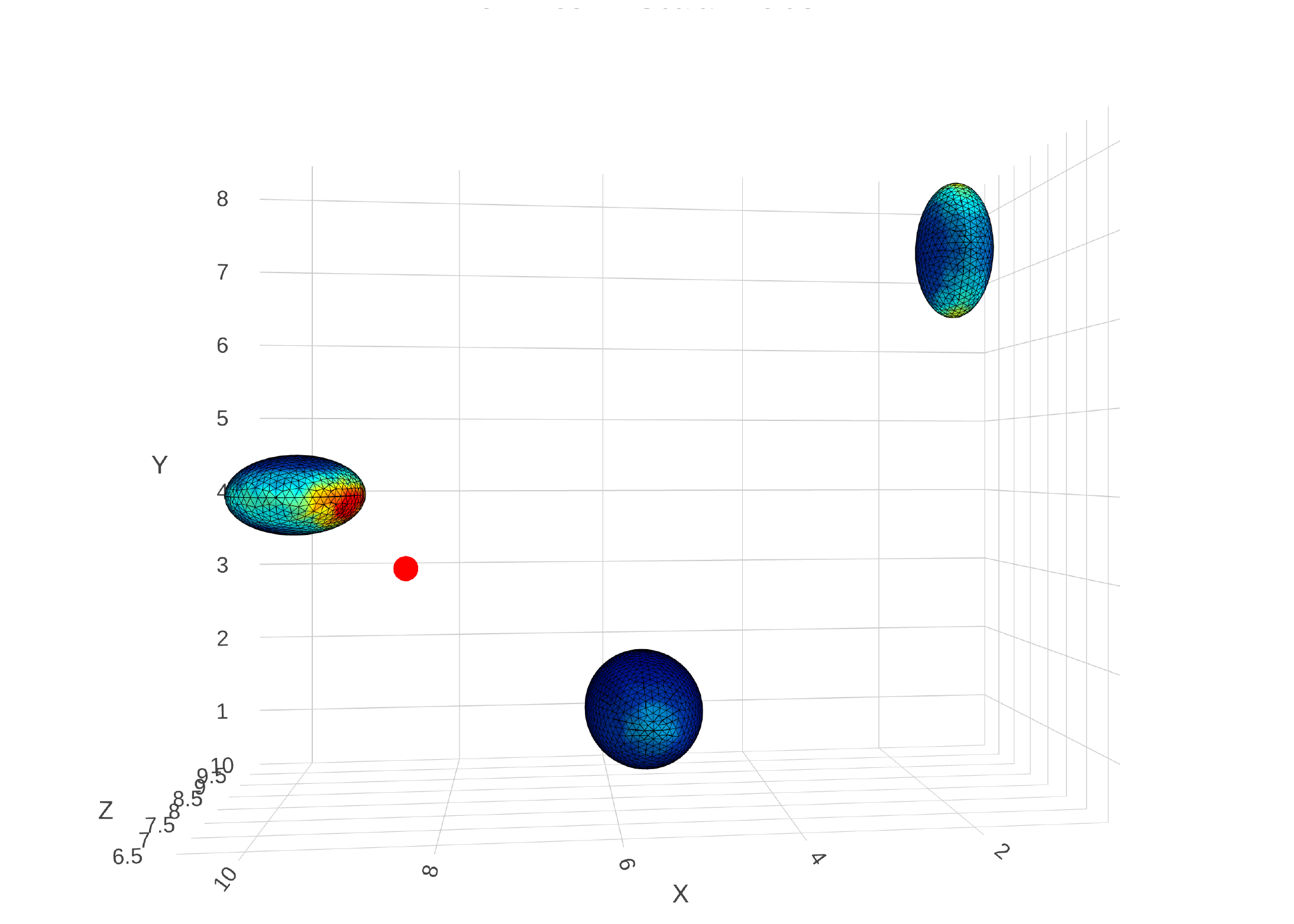}
        \end{subfigure} &
        \begin{subfigure}[b]{0.38\textwidth}
            \centering
            \includegraphics[width=\linewidth,trim={3cm 1cm 4.5cm 5cm},clip]{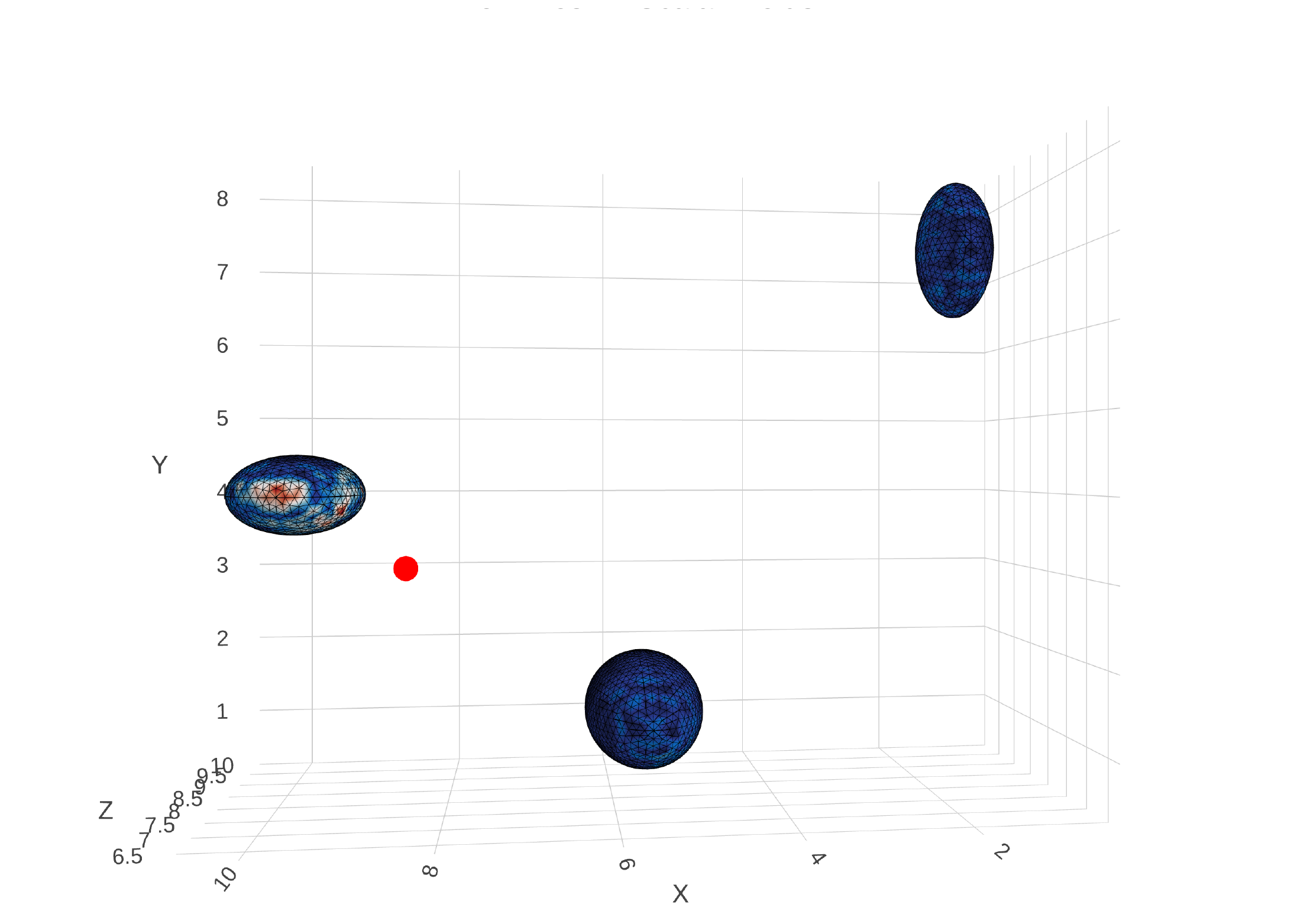}
        \end{subfigure} \\[12pt]

        \raisebox{1.5cm}{\rotatebox{90}{\textbf{ScaGNN}}} &
        \begin{subfigure}[b]{0.38\textwidth}
            \centering
            \includegraphics[width=\linewidth,trim={3cm 1cm 4.5cm 5cm},clip]{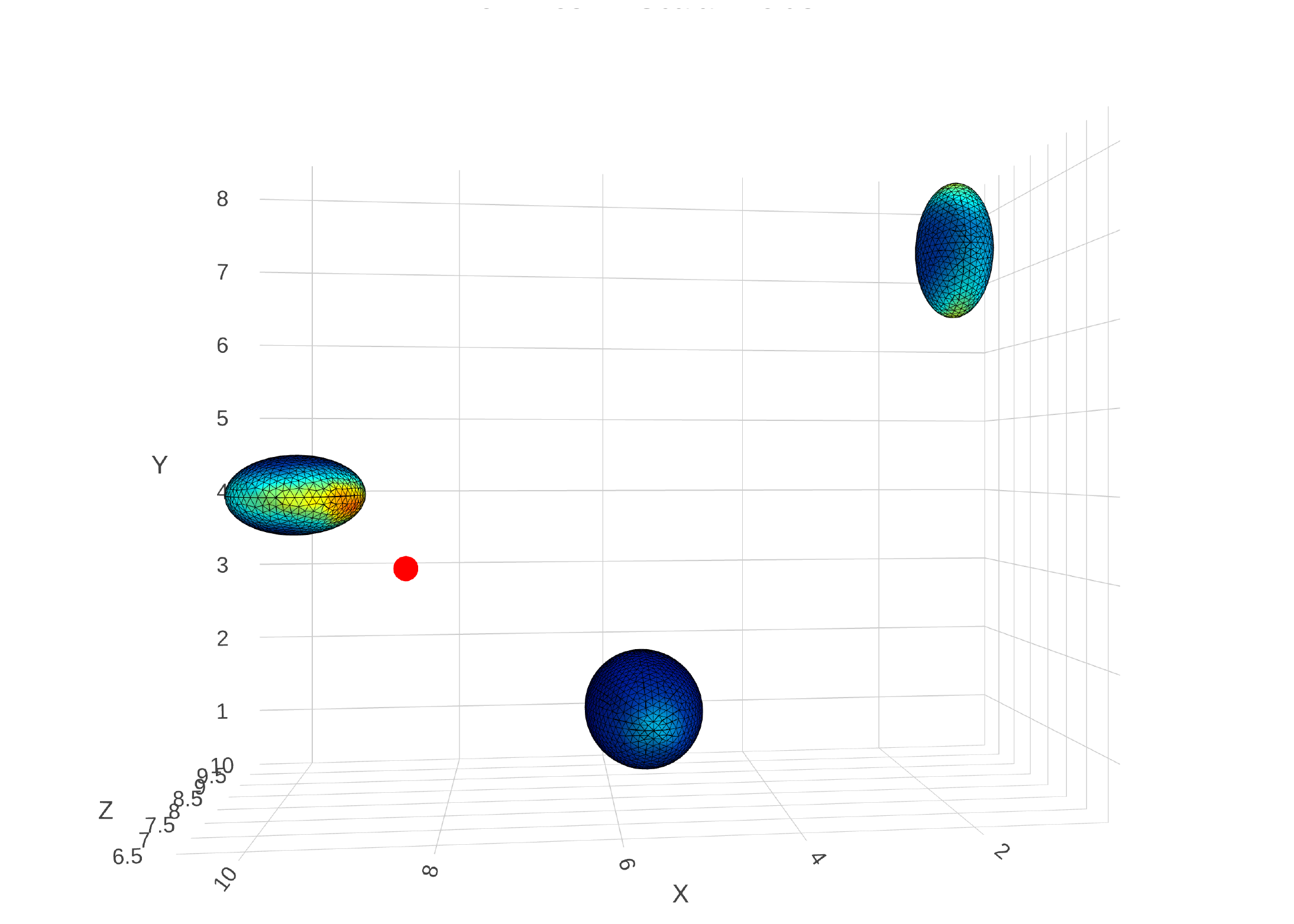}
        \end{subfigure} &
        \begin{subfigure}[b]{0.38\textwidth}
            \centering
            \includegraphics[width=\linewidth,trim={3cm 1cm 4.5cm 5cm},clip]{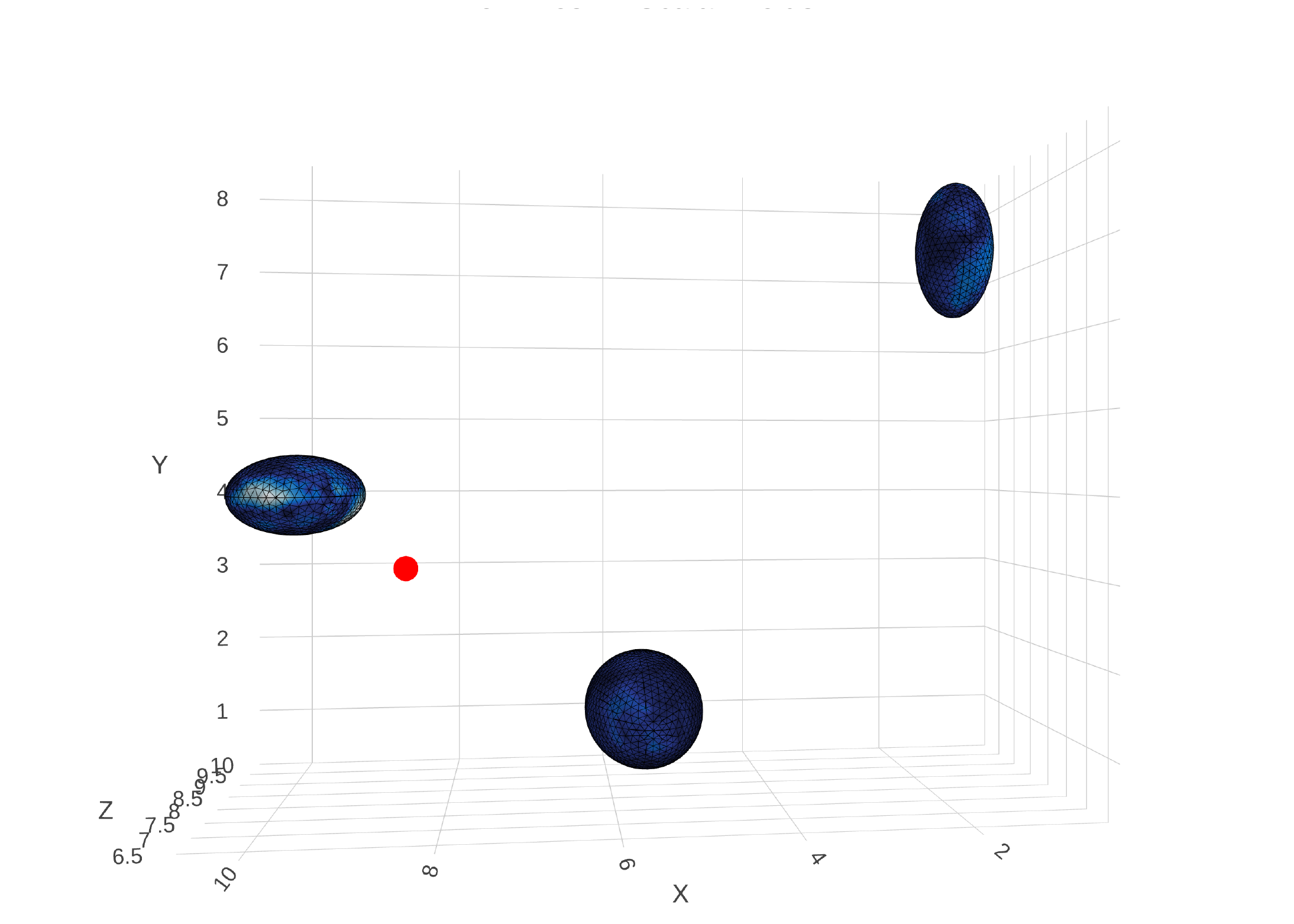}
        \end{subfigure} \\[12pt]

        \raisebox{1.2cm}{\rotatebox{90}{\textbf{Groundtruth}}} &
        \begin{subfigure}[b]{0.38\textwidth}
            \centering
            \includegraphics[width=\linewidth,trim={3cm 1cm 4.5cm 5cm},clip]{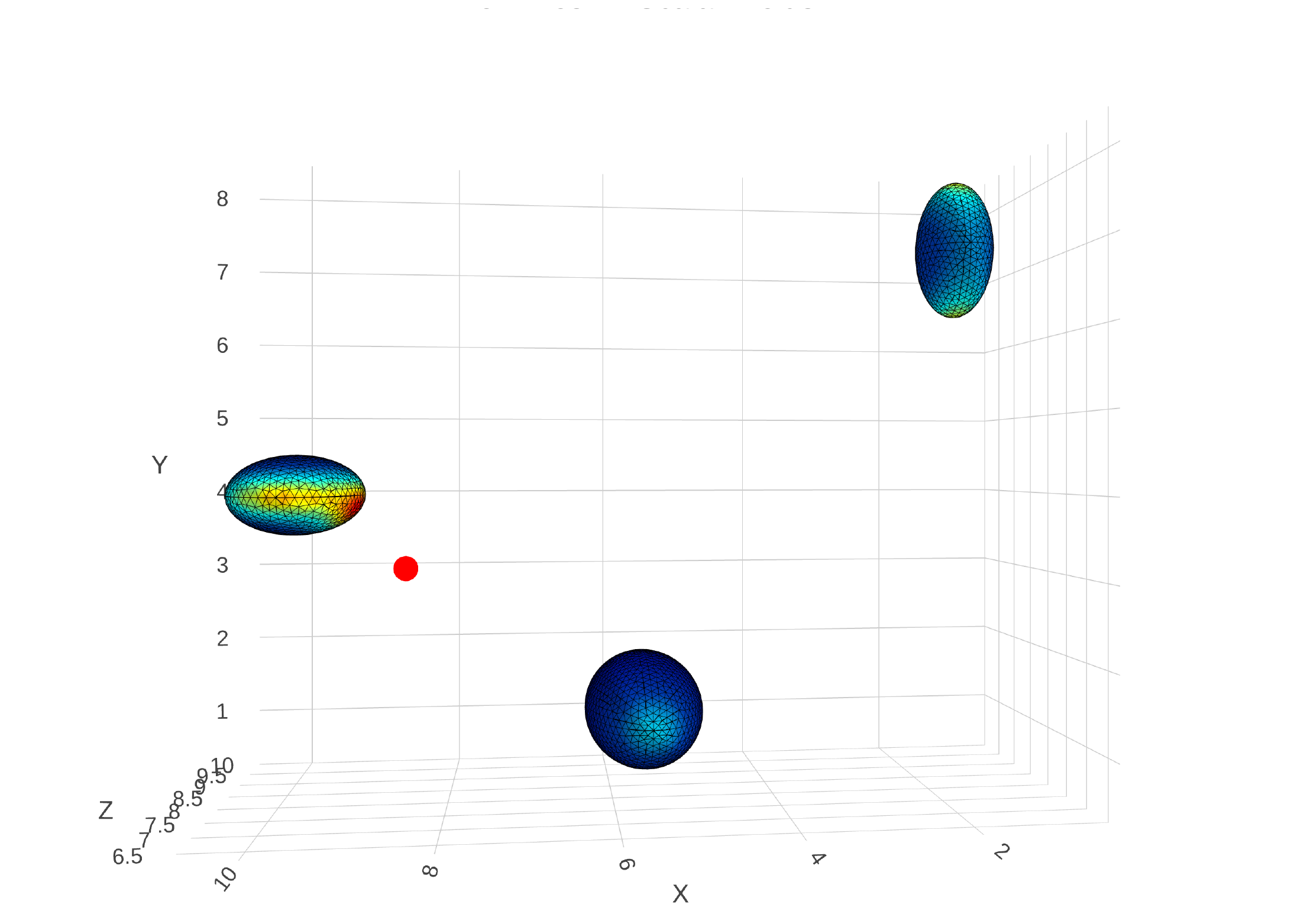}
        \end{subfigure} &
        \begin{subfigure}[b]{0.38\textwidth}
            \centering
            \hspace*{0.8cm}\includegraphics[width=\linewidth,trim={5cm 5cm 0cm 0cm},clip]{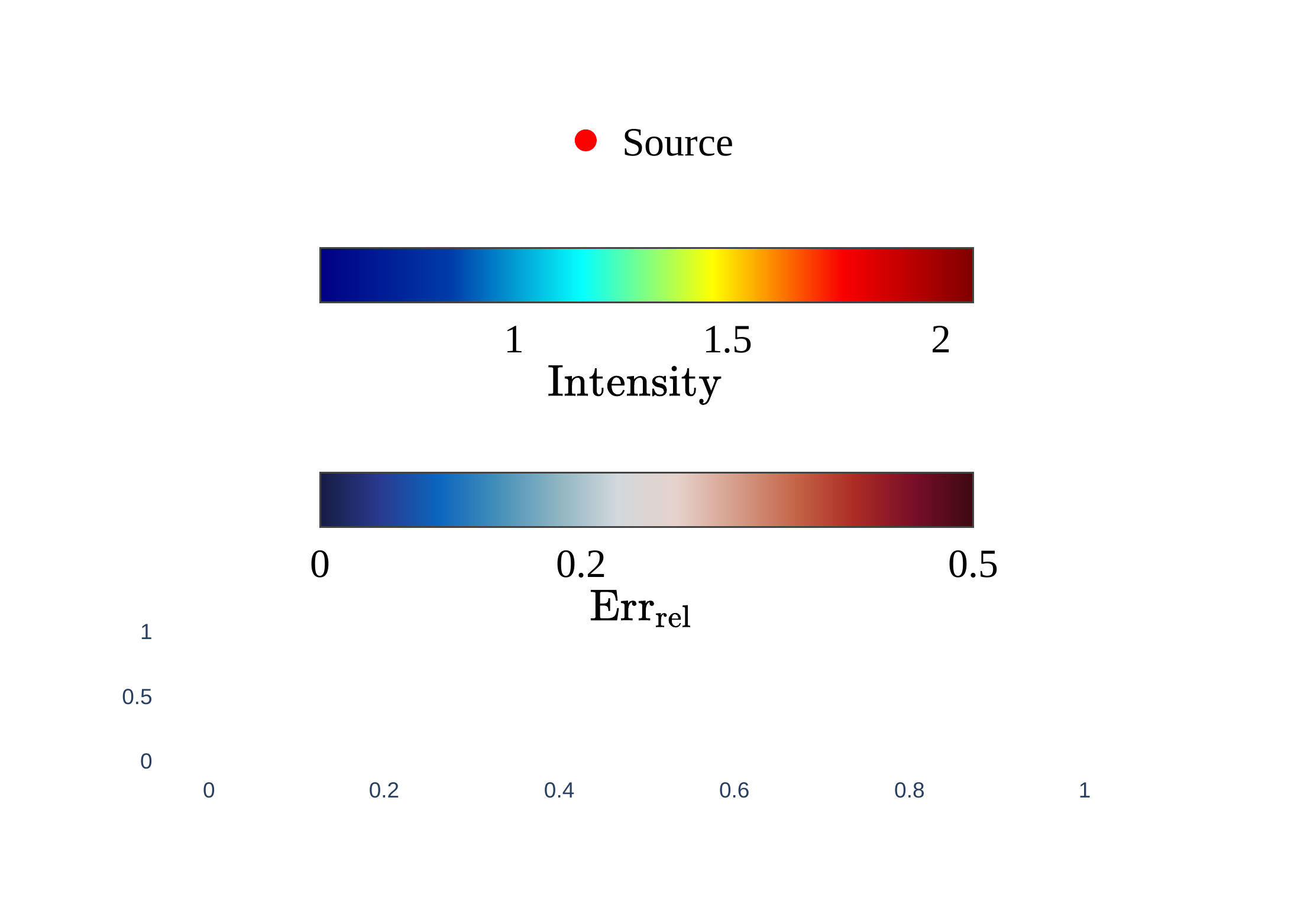}
        \end{subfigure}
    \end{tabular}
    \caption{Qualitative results on the Laplace Dirichlet problem with Point Transformer v3 \citep{wu2024point} and ScaGNN.}
    \label{fig:quali_laplace_dirichlet}
\end{figure}

\begin{figure}[htbp]
    \centering

    \begin{tabular}{c c c}
        &
        \textbf{log-Amplitude} &
        \textbf{Relative amplitude error} \\[4pt]

        \raisebox{0.5cm}{\rotatebox{90}{\textbf{Point Transformer v3}}} &
        \begin{subfigure}[b]{0.38\textwidth}
            \centering
            \includegraphics[width=\linewidth,trim={3cm 1cm 4.5cm 5cm},clip]{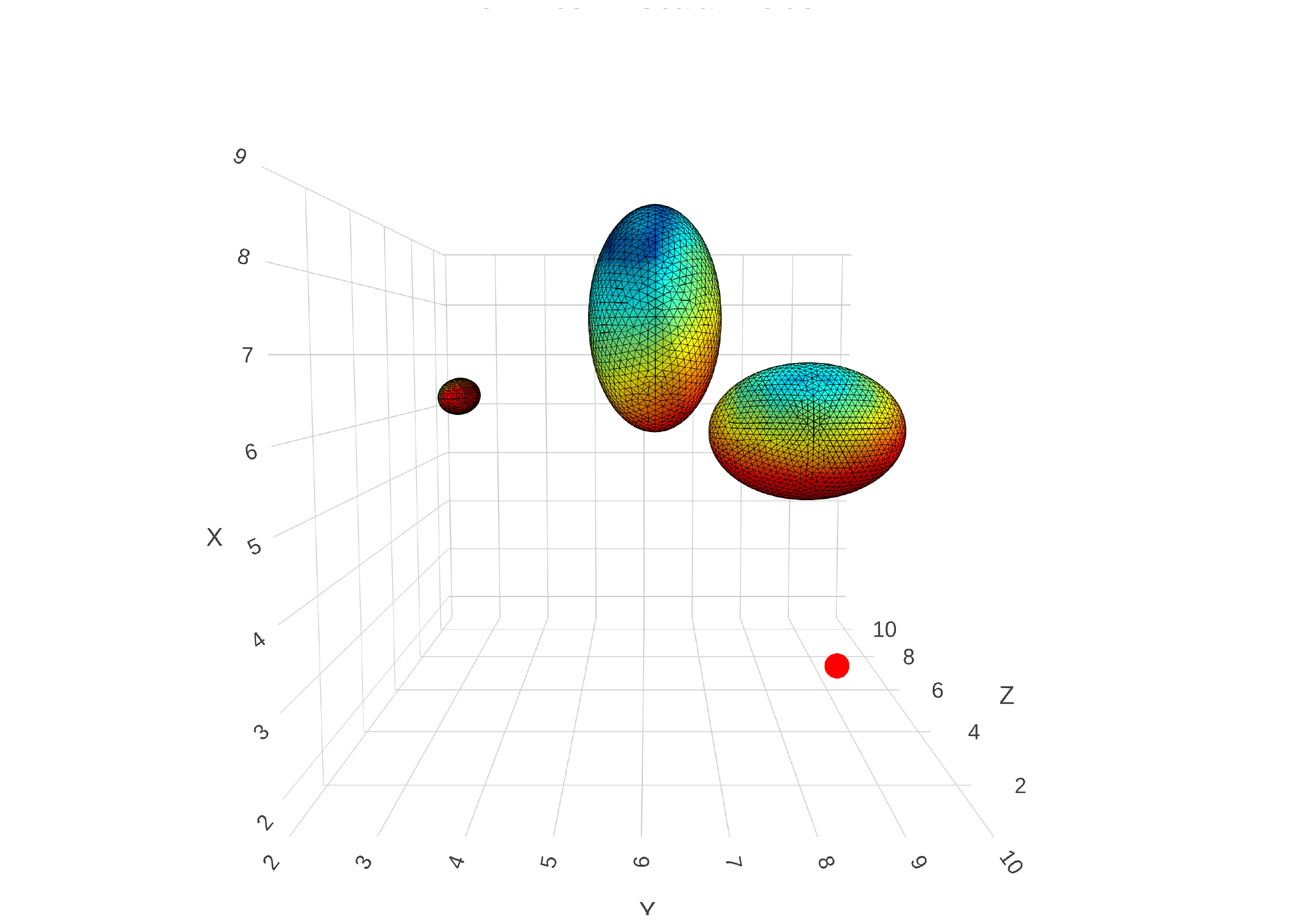}
        \end{subfigure} &
        \begin{subfigure}[b]{0.38\textwidth}
            \centering
            \includegraphics[width=\linewidth,trim={3cm 1cm 4.5cm 5cm},clip]{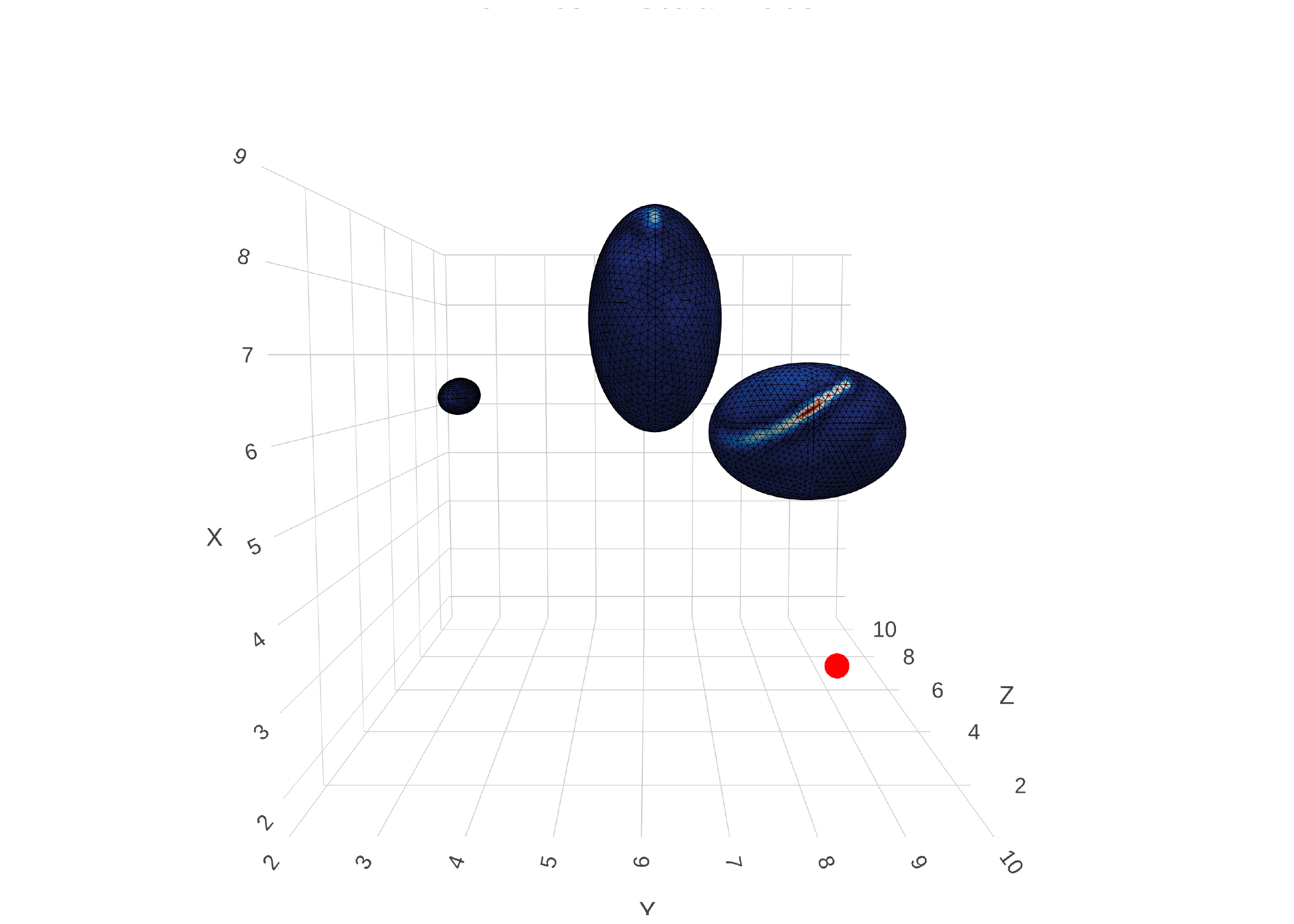}
        \end{subfigure} \\[12pt]

        \raisebox{1.5cm}{\rotatebox{90}{\textbf{ScaGNN}}} &
        \begin{subfigure}[b]{0.38\textwidth}
            \centering
            \includegraphics[width=\linewidth,trim={3cm 1cm 4.5cm 5cm},clip]{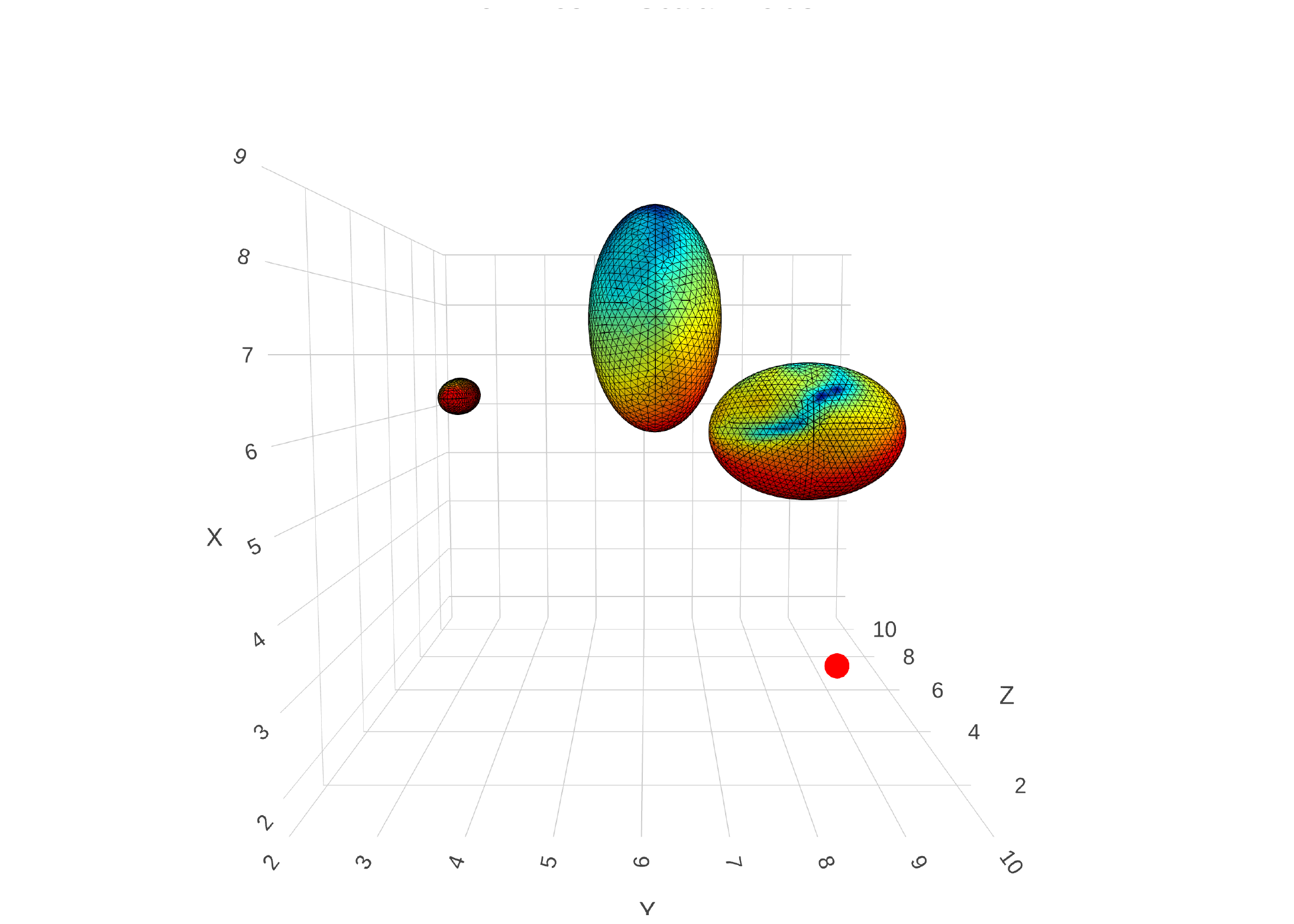}
        \end{subfigure} &
        \begin{subfigure}[b]{0.38\textwidth}
            \centering
            \includegraphics[width=\linewidth,trim={3cm 1cm 4.5cm 5cm},clip]{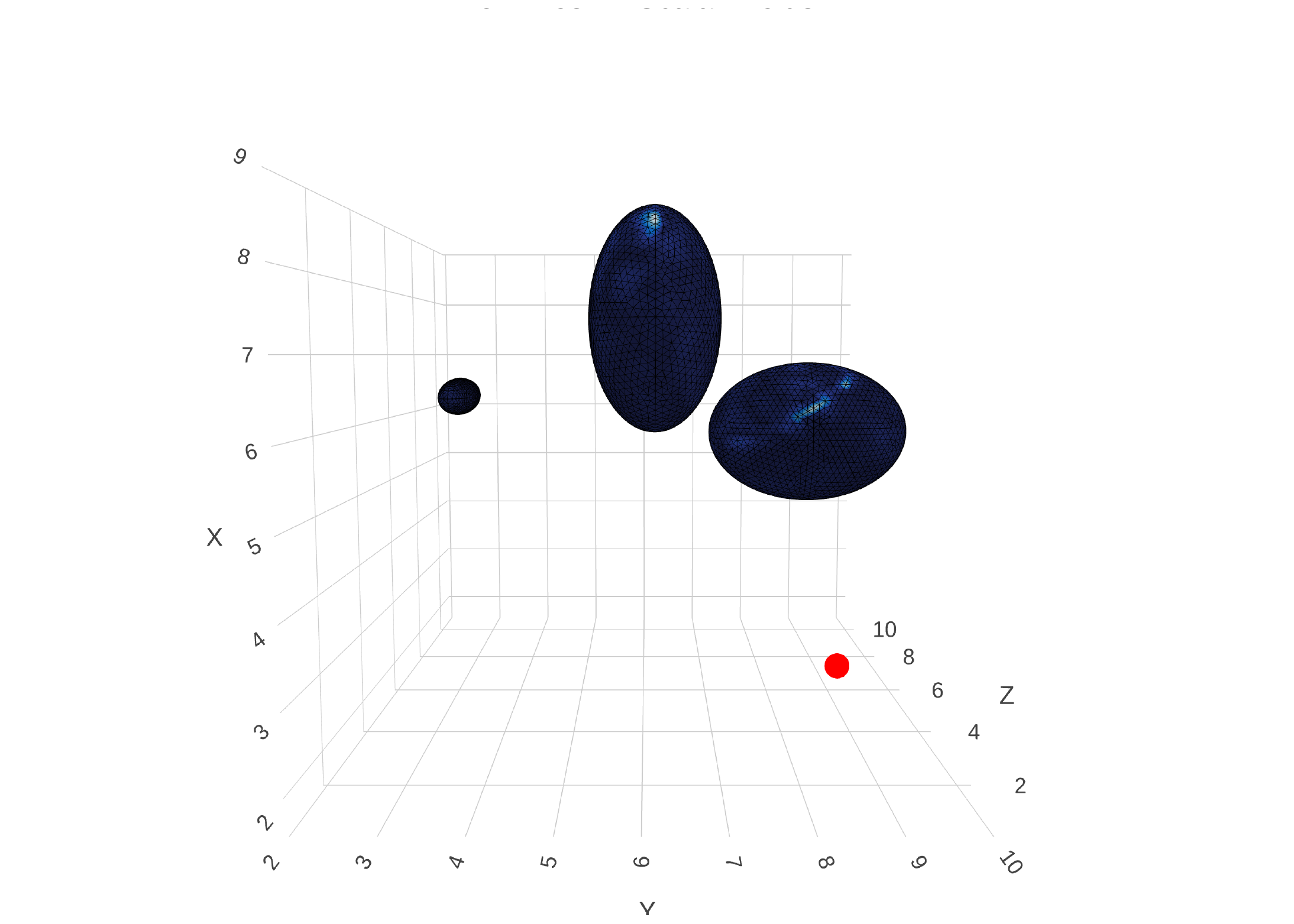}
        \end{subfigure} \\[12pt]

        \raisebox{1.2cm}{\rotatebox{90}{\textbf{Groundtruth}}} &
        \begin{subfigure}[b]{0.38\textwidth}
            \centering
            \includegraphics[width=\linewidth,trim={3cm 1cm 4.5cm 5cm},clip]{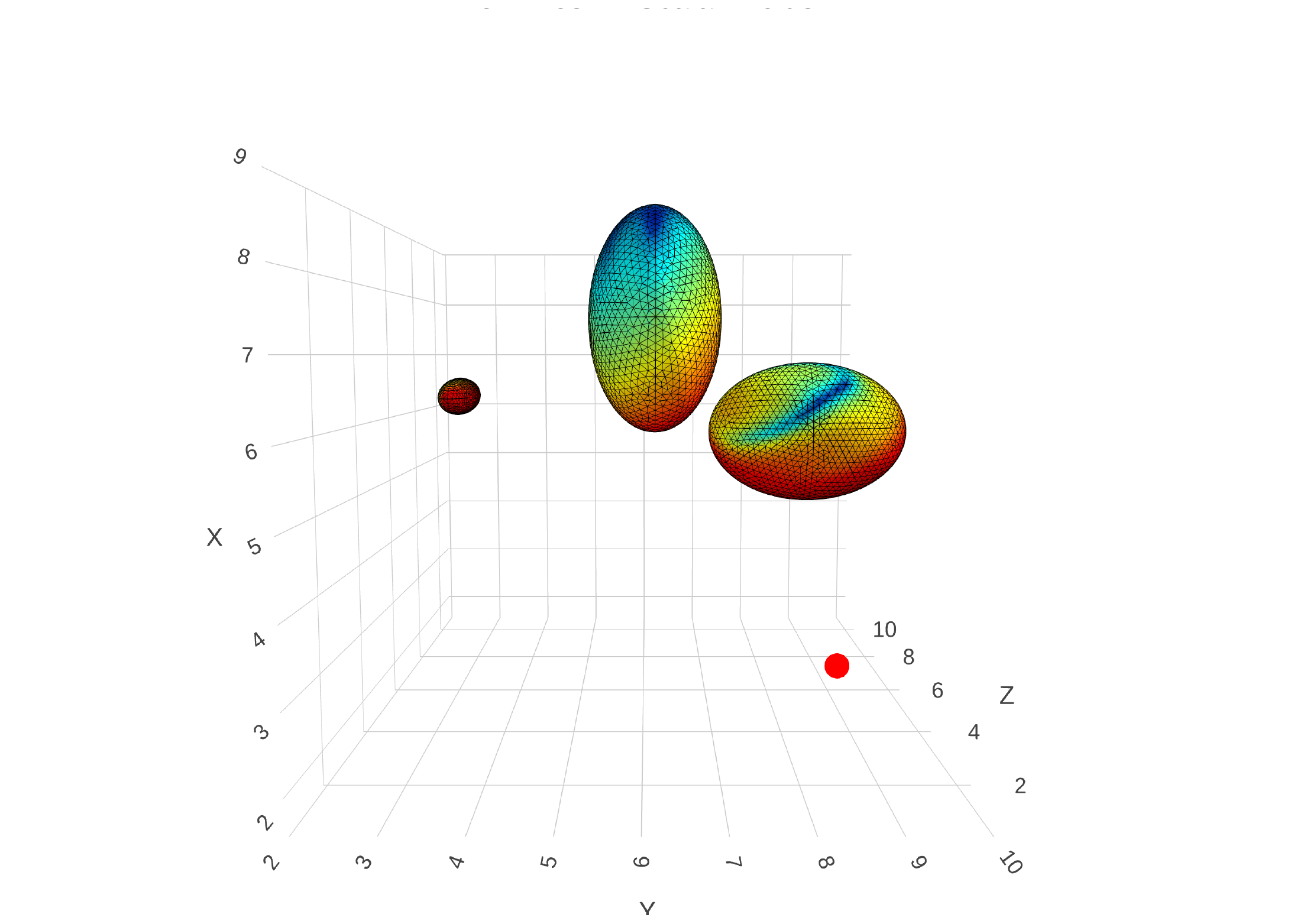}
        \end{subfigure} &
        \begin{subfigure}[b]{0.38\textwidth}
            \centering
            \hspace*{0.8cm}\includegraphics[width=\linewidth,trim={5cm 5cm 0cm 0cm},clip]{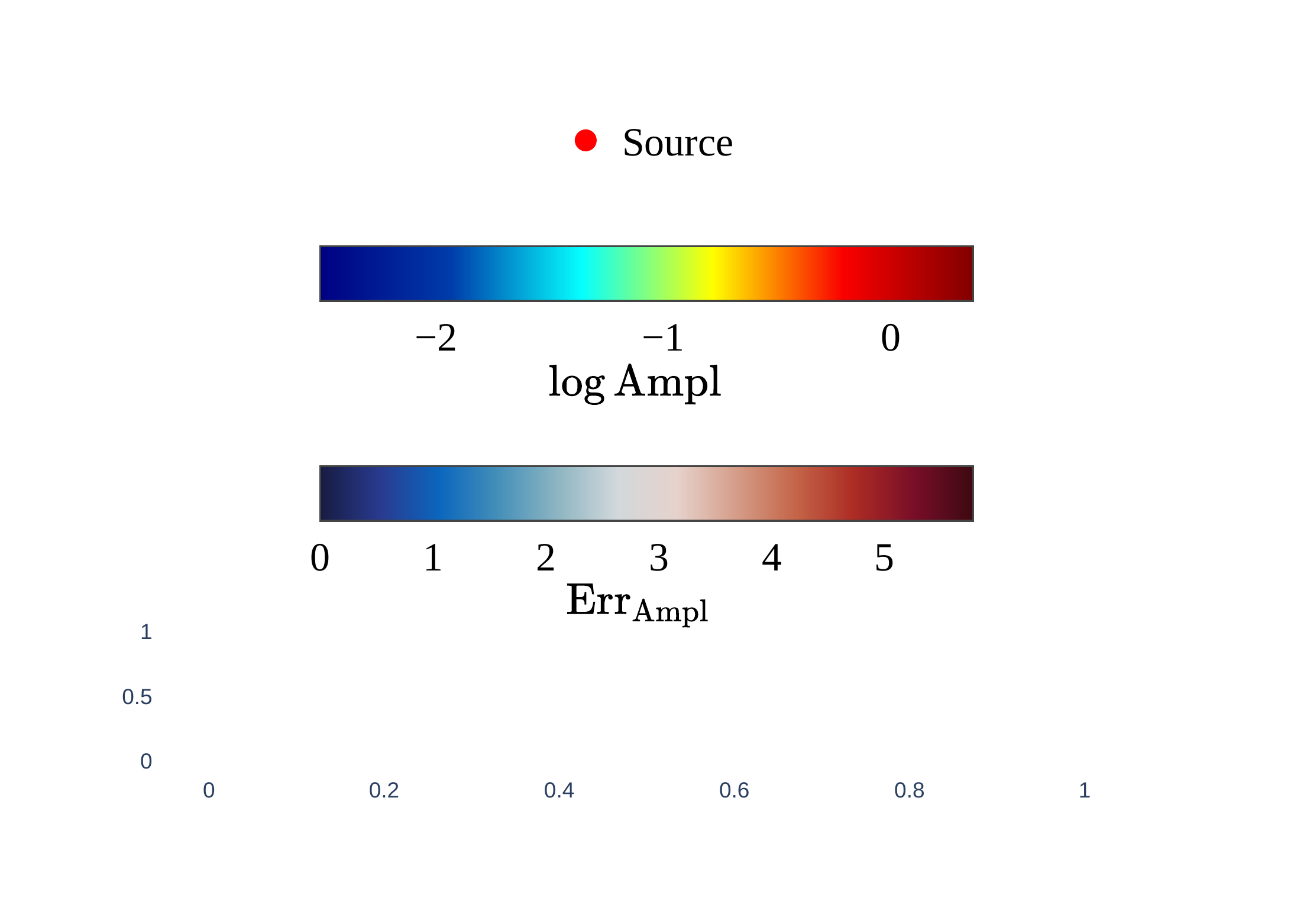}
        \end{subfigure}
    \end{tabular}
    \caption{Qualitative results for amplitude predictions on the Helmholtz Dirichlet problem with Point Transformer v3 \citep{wu2024point} and ScaGNN.}
    \label{fig:quali_helmholtz_dirichlet_ampl}
\end{figure}

\begin{figure}[htbp]
    \centering

    \begin{tabular}{c c c}
        &
        \textbf{Angle} &
        \textbf{Angle error} \\[4pt]

        \raisebox{0.5cm}{\rotatebox{90}{\textbf{Point Transformer v3}}} &
        \begin{subfigure}[b]{0.38\textwidth}
            \centering
            \includegraphics[width=\linewidth,trim={3cm 1cm 4.5cm 5cm},clip]{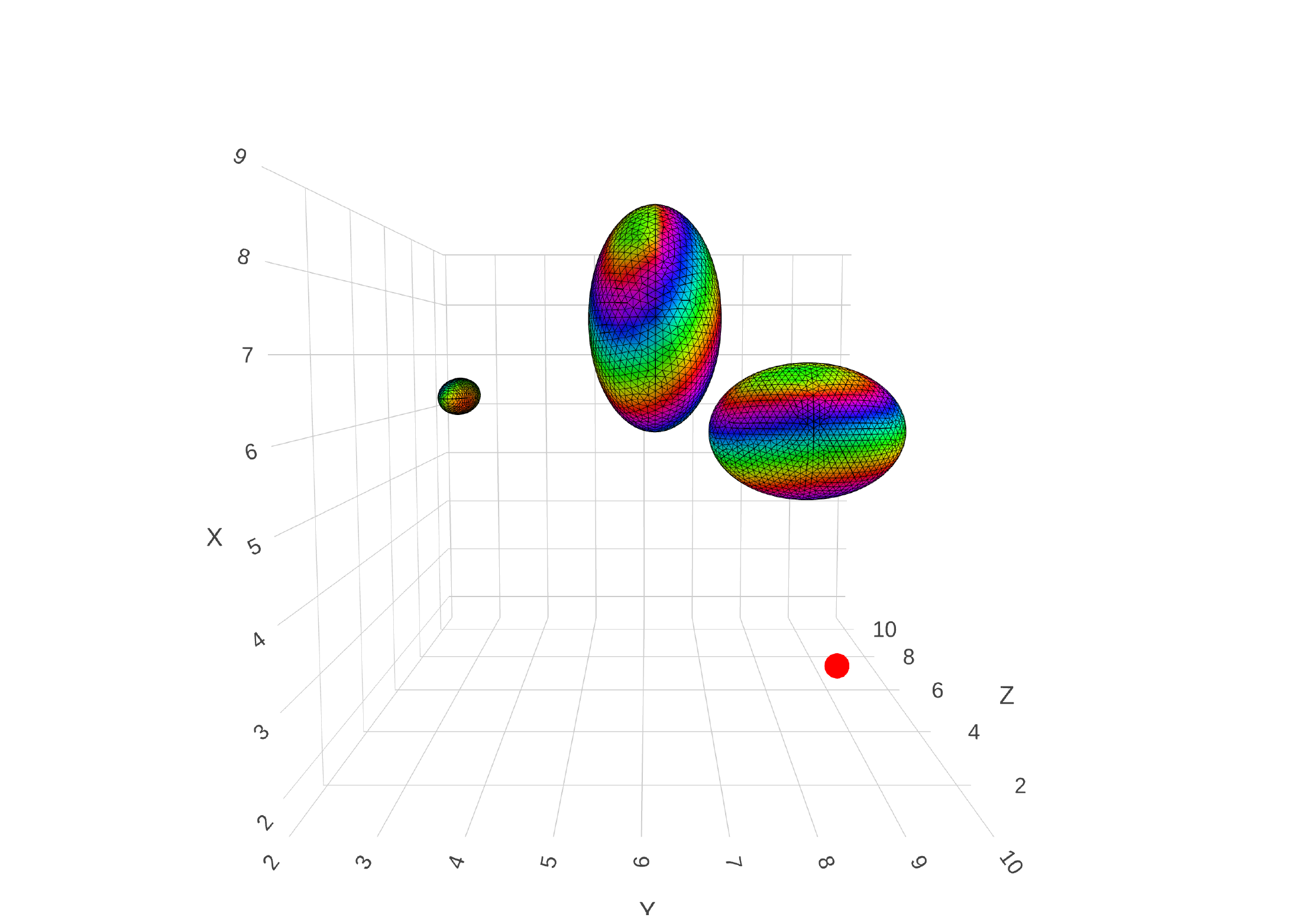}
        \end{subfigure} &
        \begin{subfigure}[b]{0.38\textwidth}
            \centering
            \includegraphics[width=\linewidth,trim={3cm 1cm 4.5cm 5cm},clip]{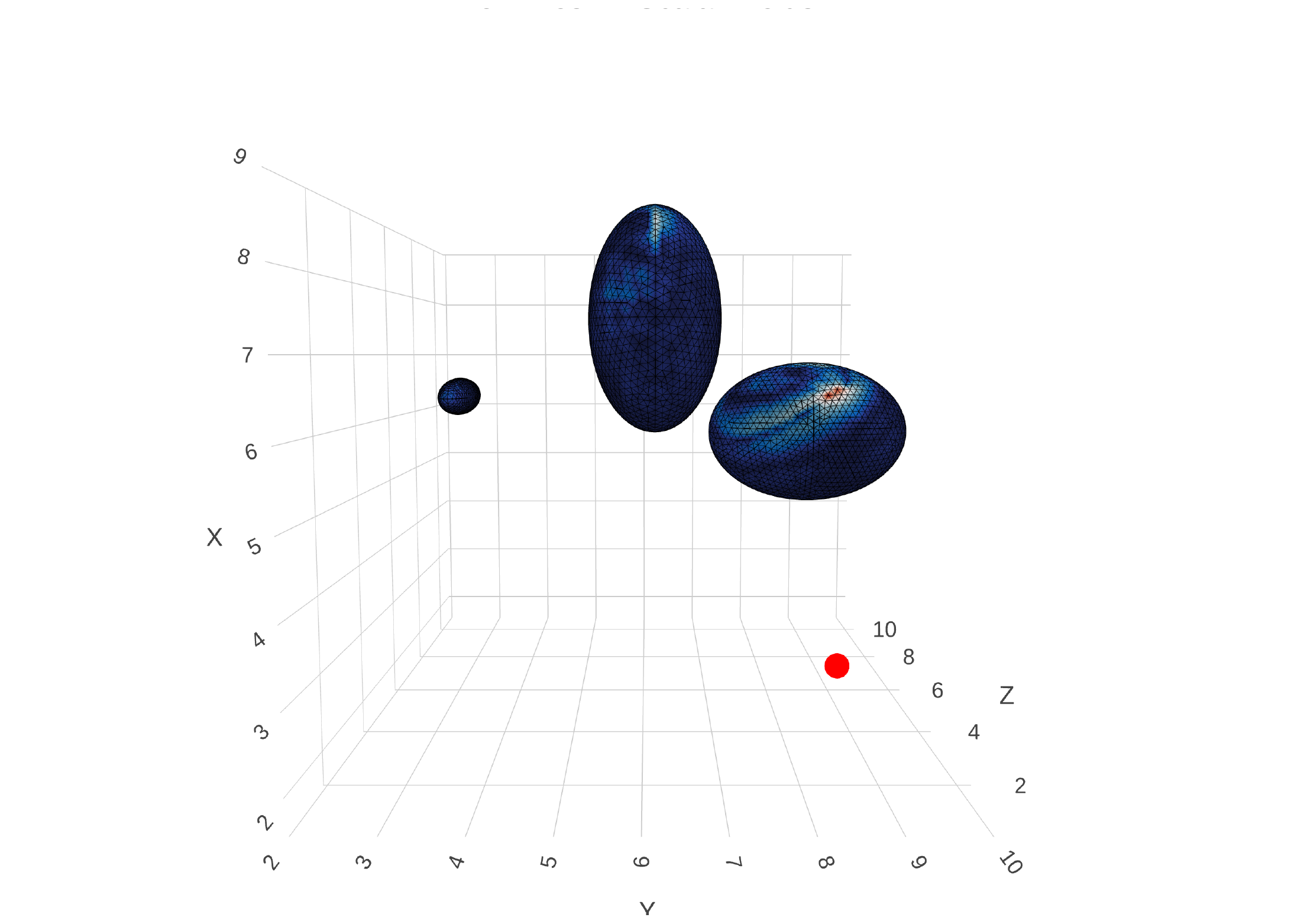}
        \end{subfigure} \\[12pt]

        \raisebox{1.5cm}{\rotatebox{90}{\textbf{ScaGNN}}} &
        \begin{subfigure}[b]{0.38\textwidth}
            \centering
            \includegraphics[width=\linewidth,trim={3cm 1cm 4.5cm 5cm},clip]{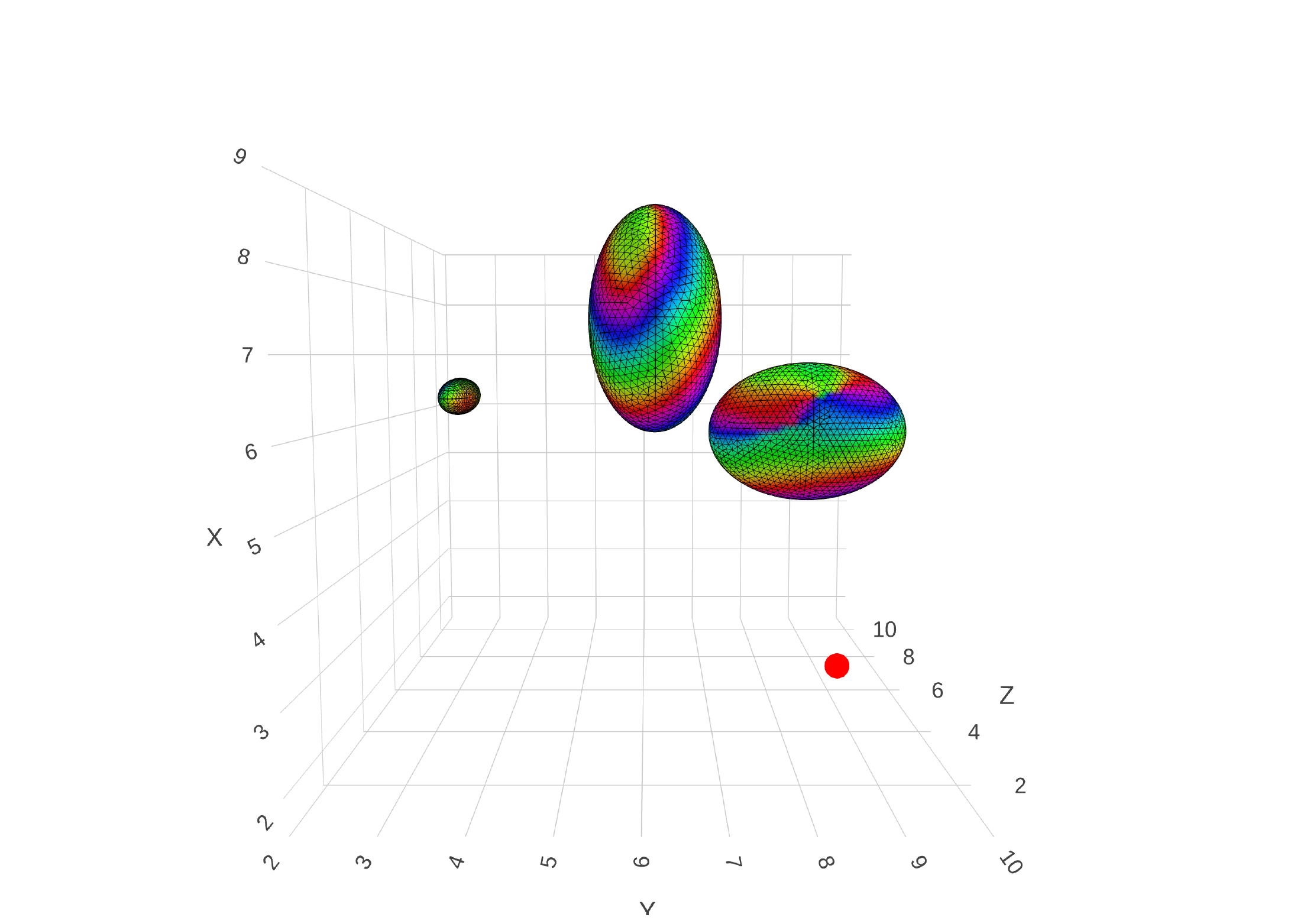}
        \end{subfigure} &
        \begin{subfigure}[b]{0.38\textwidth}
            \centering
            \includegraphics[width=\linewidth,trim={3cm 1cm 4.5cm 5cm},clip]{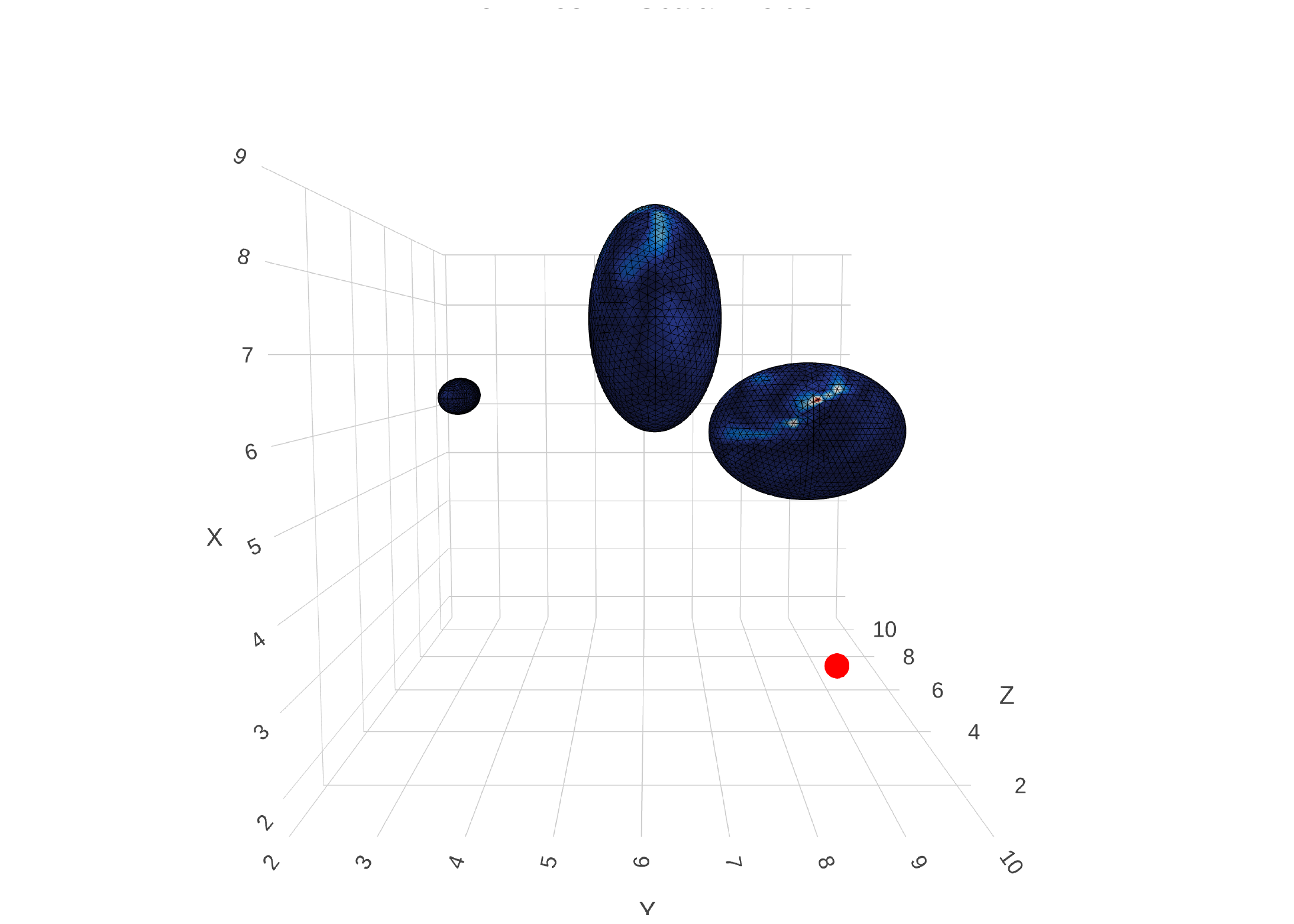}
        \end{subfigure} \\[12pt]

        \raisebox{1.2cm}{\rotatebox{90}{\textbf{Groundtruth}}} &
        \begin{subfigure}[b]{0.38\textwidth}
            \centering
            \includegraphics[width=\linewidth,trim={3cm 1cm 4.5cm 5cm},clip]{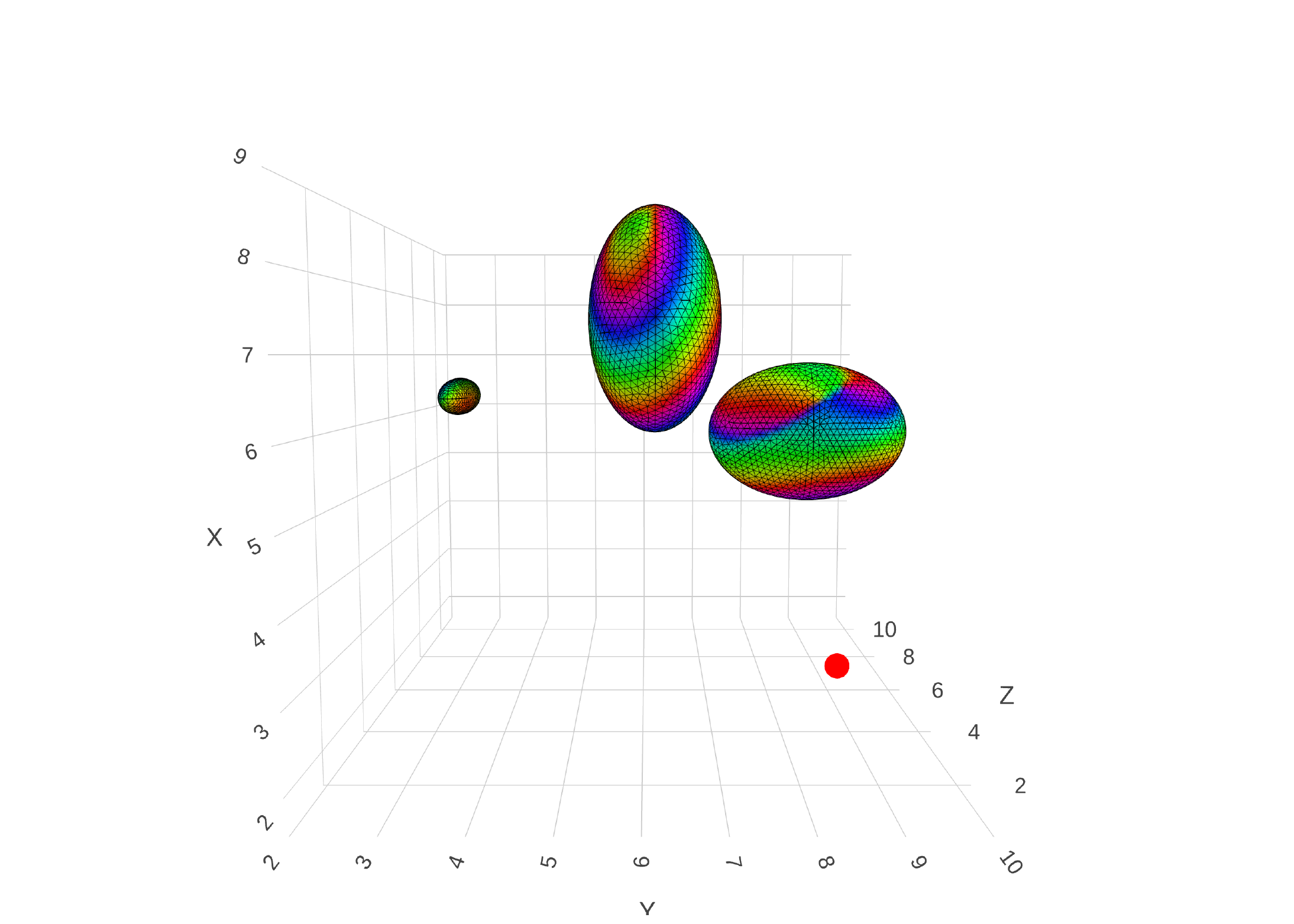}
        \end{subfigure} &
        \begin{subfigure}[b]{0.38\textwidth}
            \centering
            \hspace*{0.8cm}\includegraphics[width=\linewidth,trim={5cm 5cm 0cm 0cm},clip]{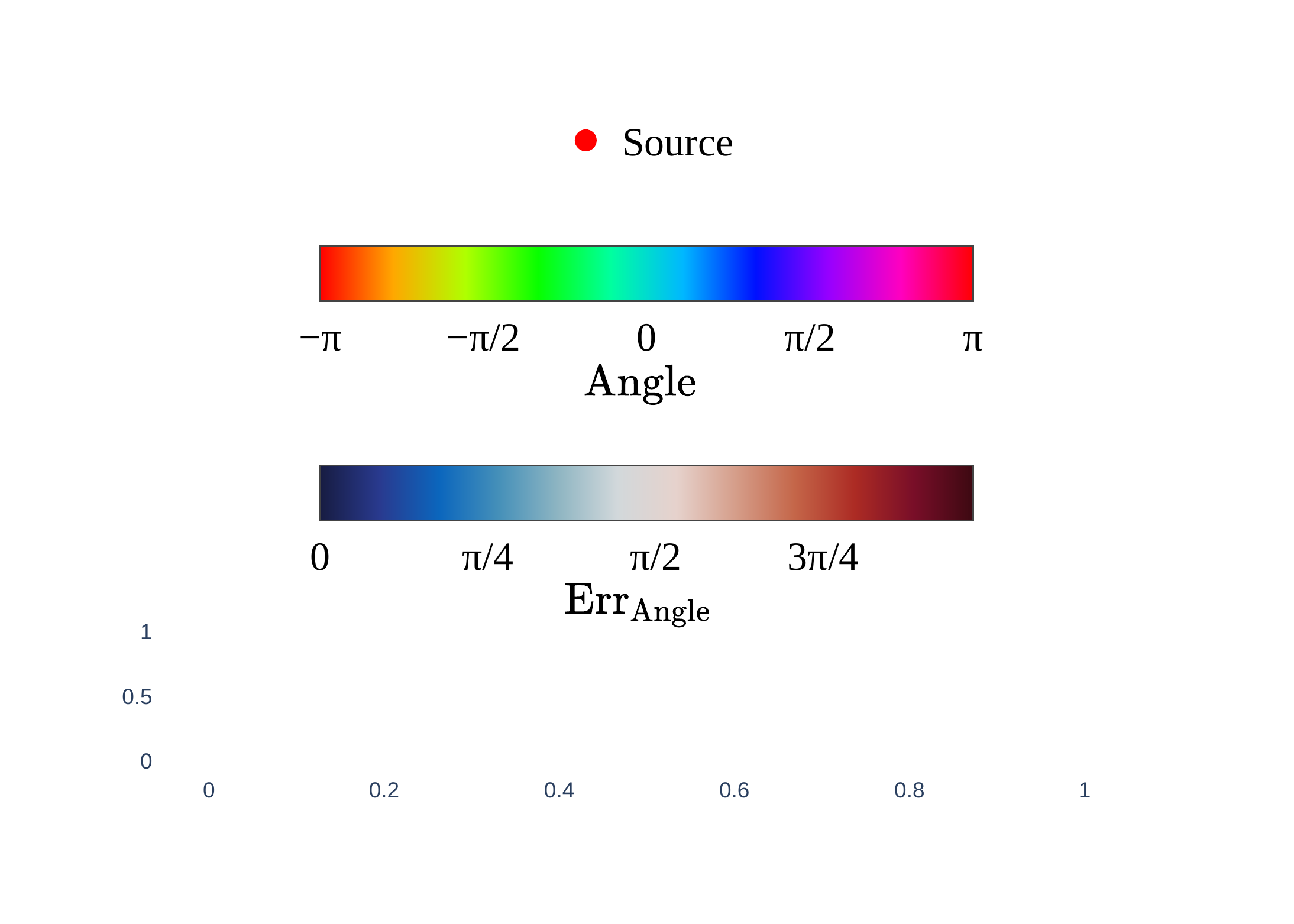}
        \end{subfigure}
    \end{tabular}
    \caption{Qualitative results for angle predictions on the Helmholtz Dirichlet problem with Point Transformer v3 \citep{wu2024point} and ScaGNN.}
    \label{fig:quali_helmholtz_dirichlet_angle}
\end{figure}

\begin{figure}[htbp]
    \centering

    \begin{tabular}{c c c}
        &
        \textbf{log-Amplitude} &
        \textbf{Relative amplitude error} \\[4pt]

        \raisebox{0.5cm}{\rotatebox{90}{\textbf{Point Transformer v3}}} &
        \begin{subfigure}[b]{0.38\textwidth}
            \centering
            \includegraphics[width=\linewidth,trim={3cm 1cm 4.5cm 5cm},clip]{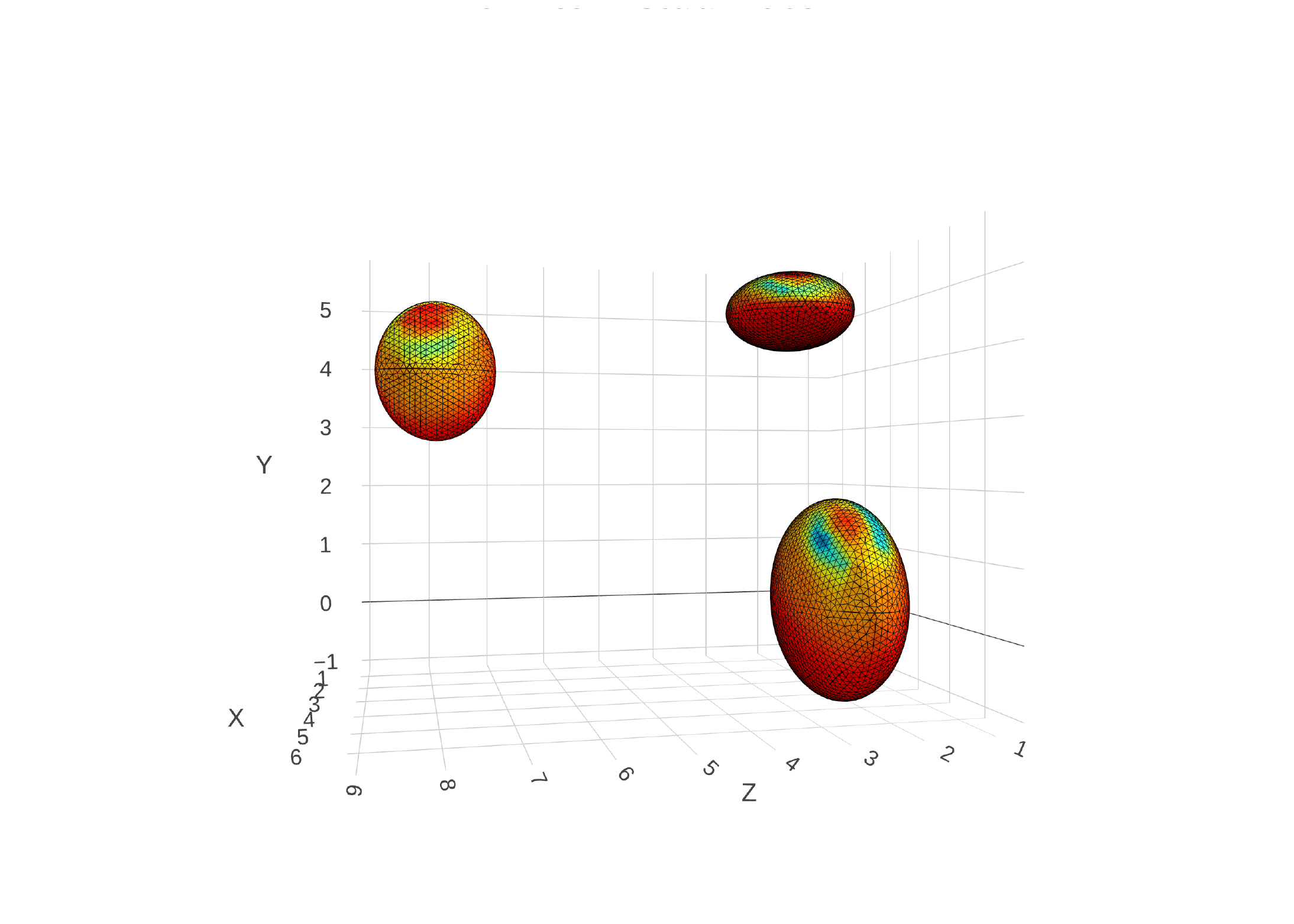}
        \end{subfigure} &
        \begin{subfigure}[b]{0.38\textwidth}
            \centering
            \includegraphics[width=\linewidth,trim={3cm 1cm 4.5cm 5cm},clip]{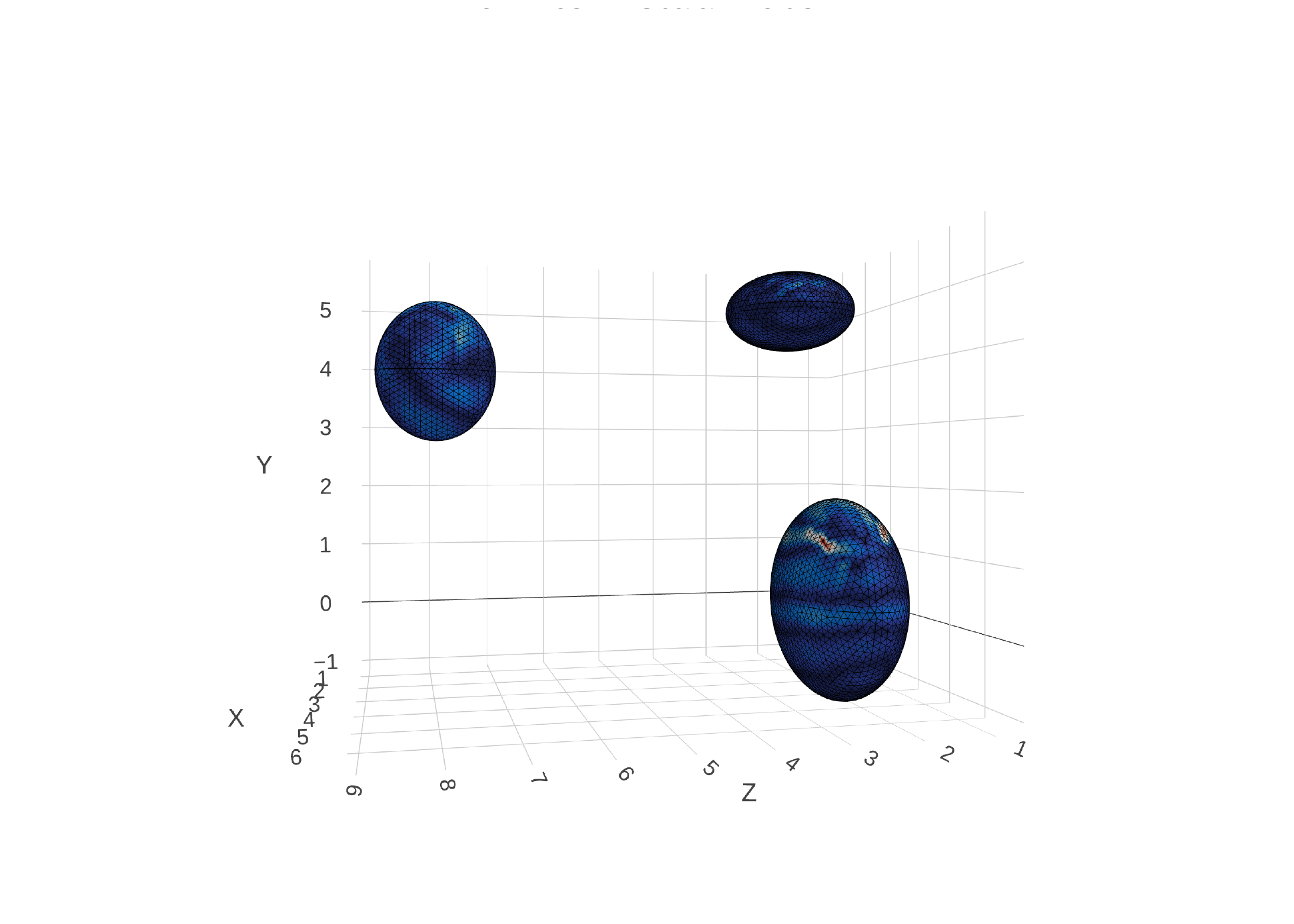}
        \end{subfigure} \\[12pt]

        \raisebox{1.5cm}{\rotatebox{90}{\textbf{ScaGNN}}} &
        \begin{subfigure}[b]{0.38\textwidth}
            \centering
            \includegraphics[width=\linewidth,trim={3cm 1cm 4.5cm 5cm},clip]{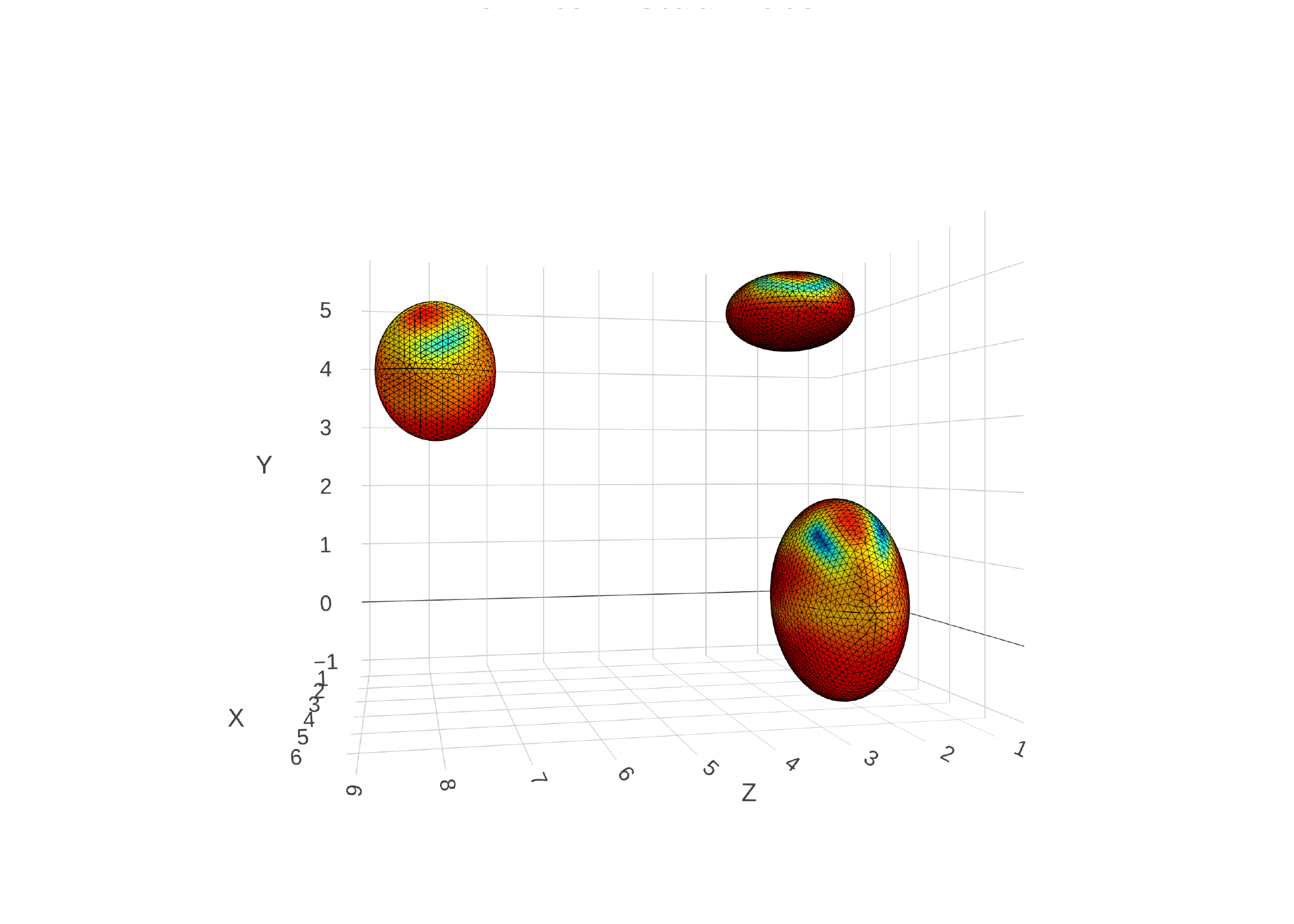}
        \end{subfigure} &
        \begin{subfigure}[b]{0.38\textwidth}
            \centering
            \includegraphics[width=\linewidth,trim={3cm 1cm 4.5cm 5cm},clip]{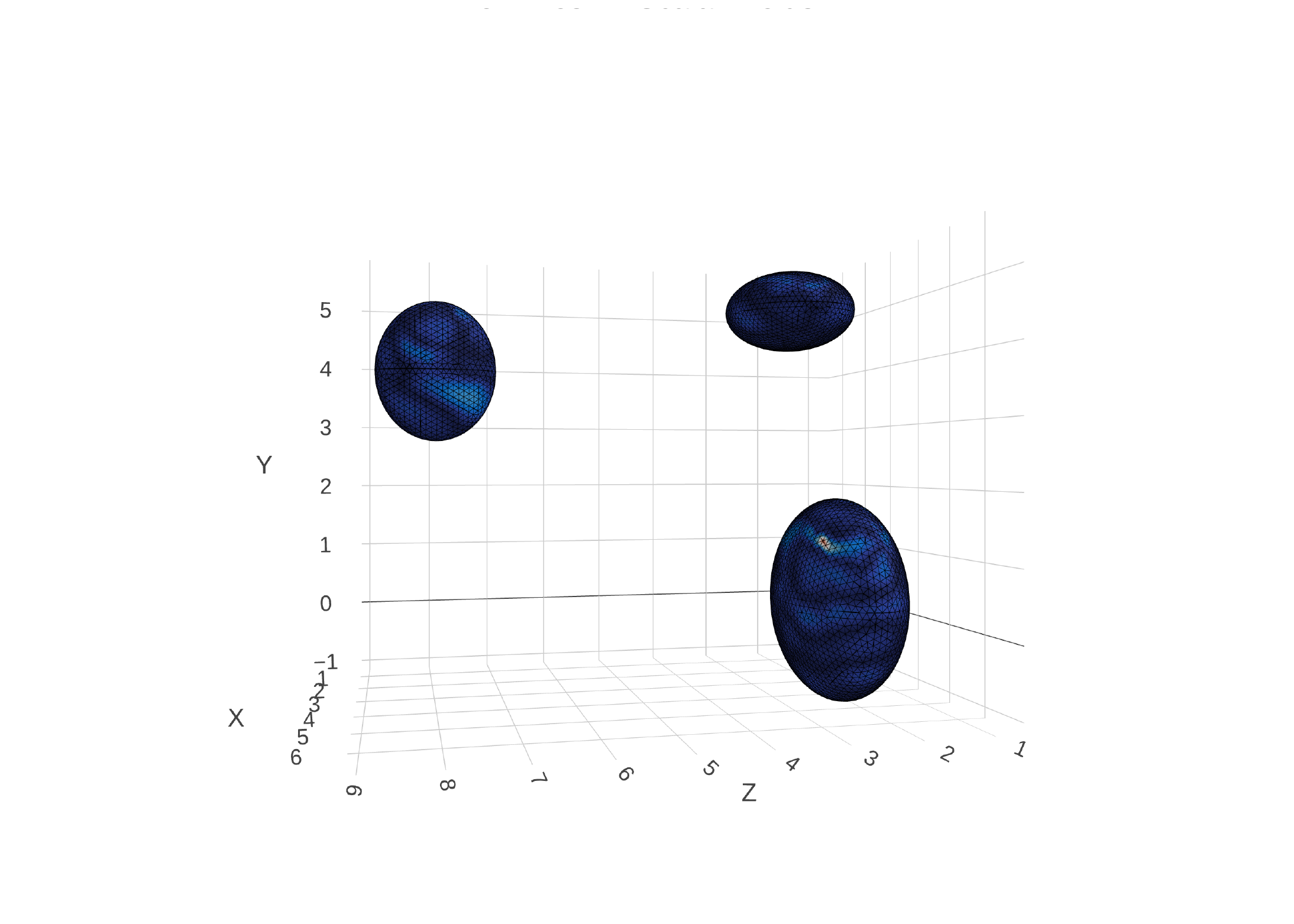}
        \end{subfigure} \\[12pt]

        \raisebox{1.2cm}{\rotatebox{90}{\textbf{Groundtruth}}} &
        \begin{subfigure}[b]{0.38\textwidth}
            \centering
            \includegraphics[width=\linewidth,trim={3cm 1cm 4.5cm 5cm},clip]{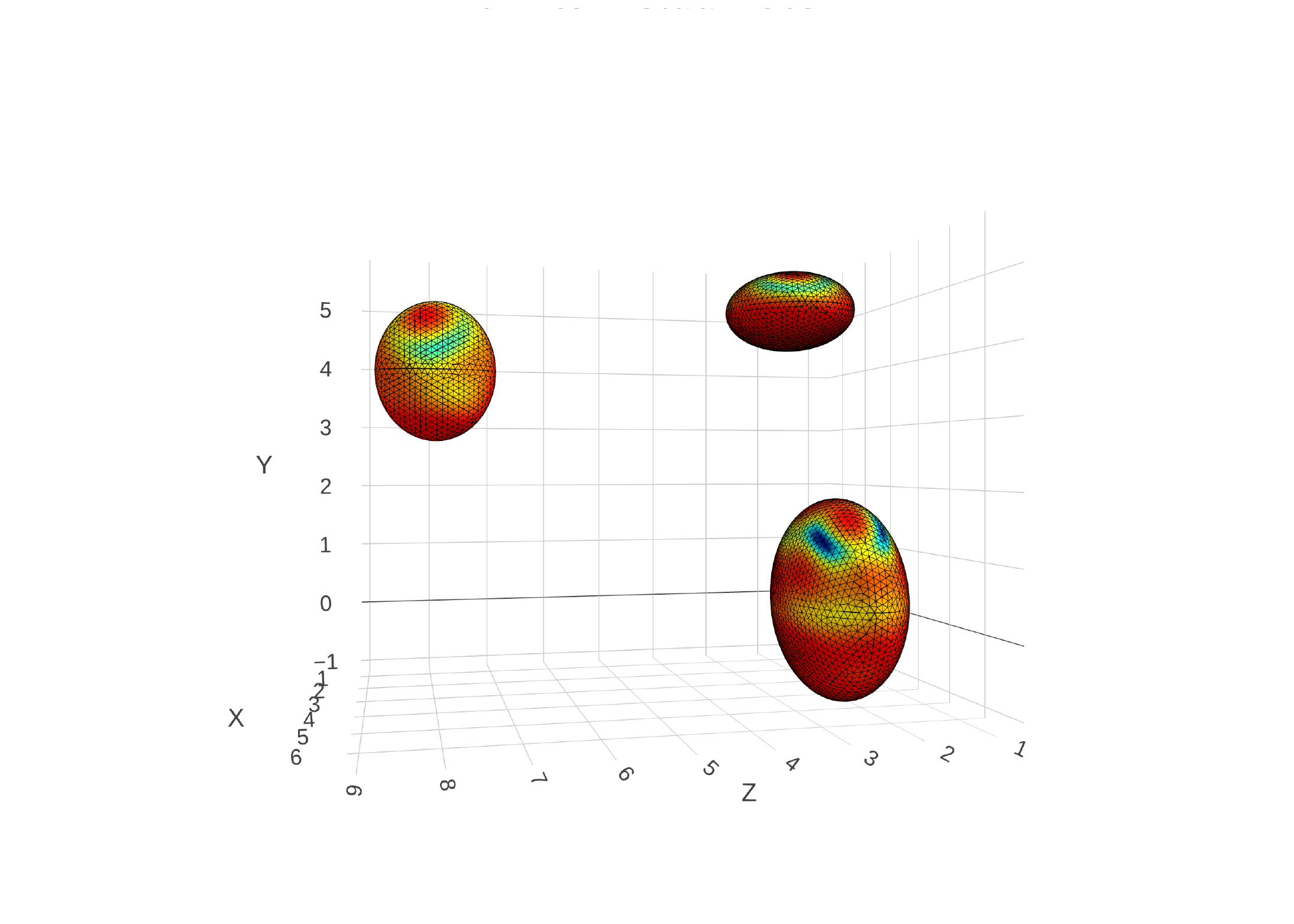}
        \end{subfigure} &
        \begin{subfigure}[b]{0.38\textwidth}
            \centering
            \hspace*{0.8cm}\includegraphics[width=\linewidth,trim={5cm 5cm 0cm 0cm},clip]{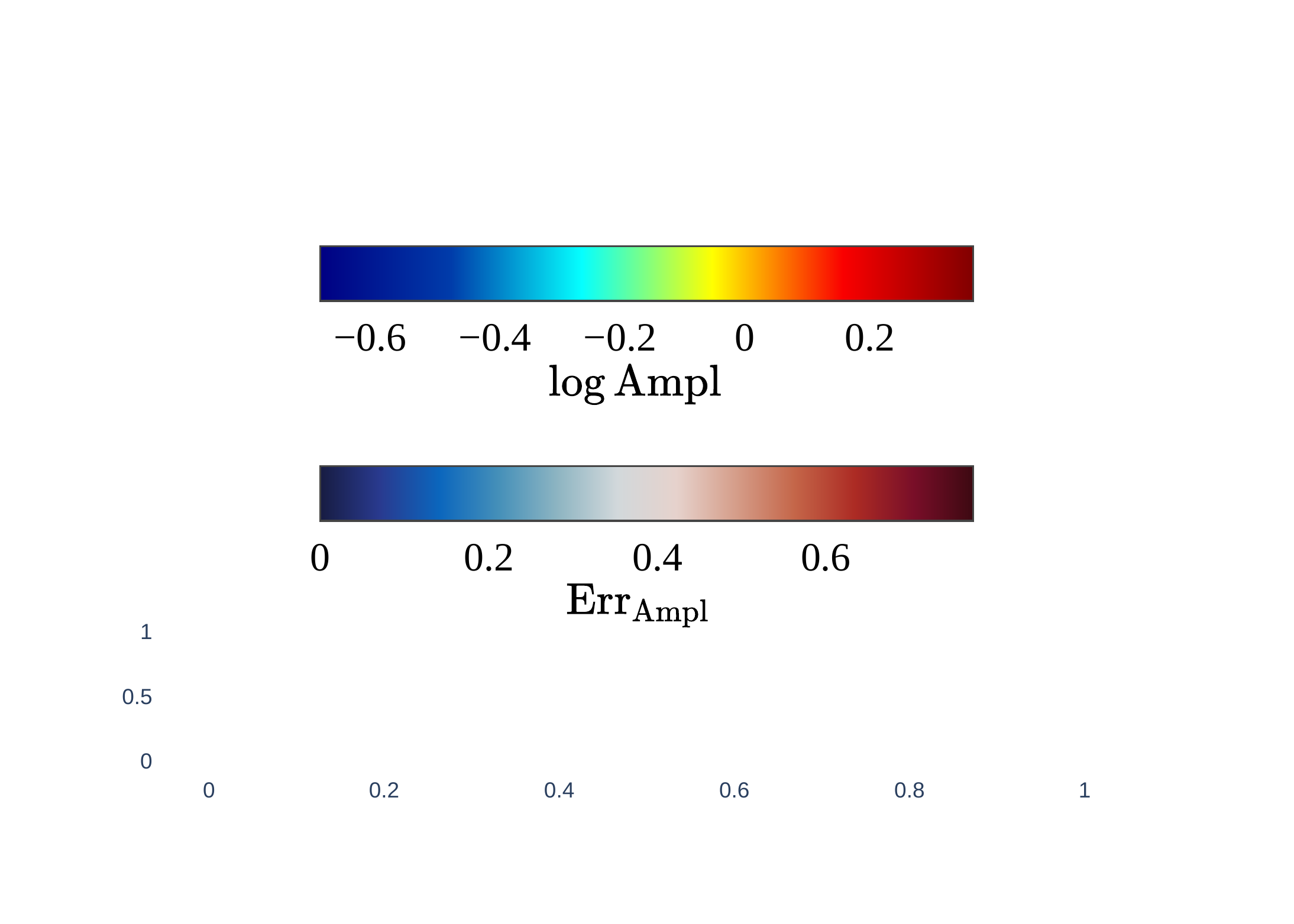}
        \end{subfigure}
    \end{tabular}
    \caption{Qualitative results for amplitude predictions on the Helmholtz Neumann problem with Point Transformer v3 \citep{wu2024point} and ScaGNN.}
    \label{fig:quali_helmholtz_neumann_ampl}
\end{figure}

\begin{figure}[htbp]
    \centering

    \begin{tabular}{c c c}
        &
        \textbf{Angle} &
        \textbf{Angle error} \\[4pt]

        \raisebox{0.5cm}{\rotatebox{90}{\textbf{Point Transformer v3}}} &
        \begin{subfigure}[b]{0.38\textwidth}
            \centering
            \includegraphics[width=\linewidth,trim={3cm 1cm 4.5cm 5cm},clip]{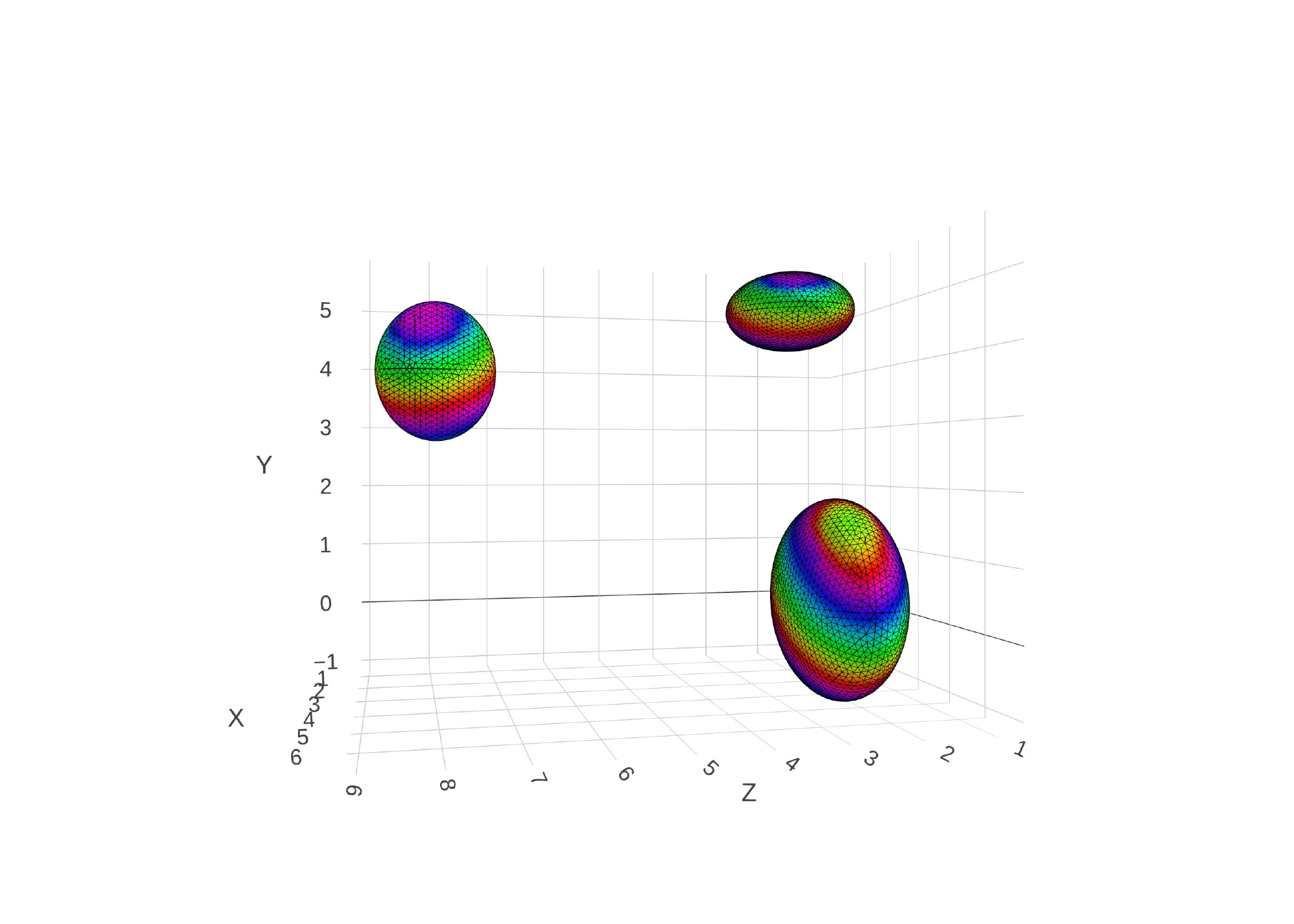}
        \end{subfigure} &
        \begin{subfigure}[b]{0.38\textwidth}
            \centering
            \includegraphics[width=\linewidth,trim={3cm 1cm 4.5cm 5cm},clip]{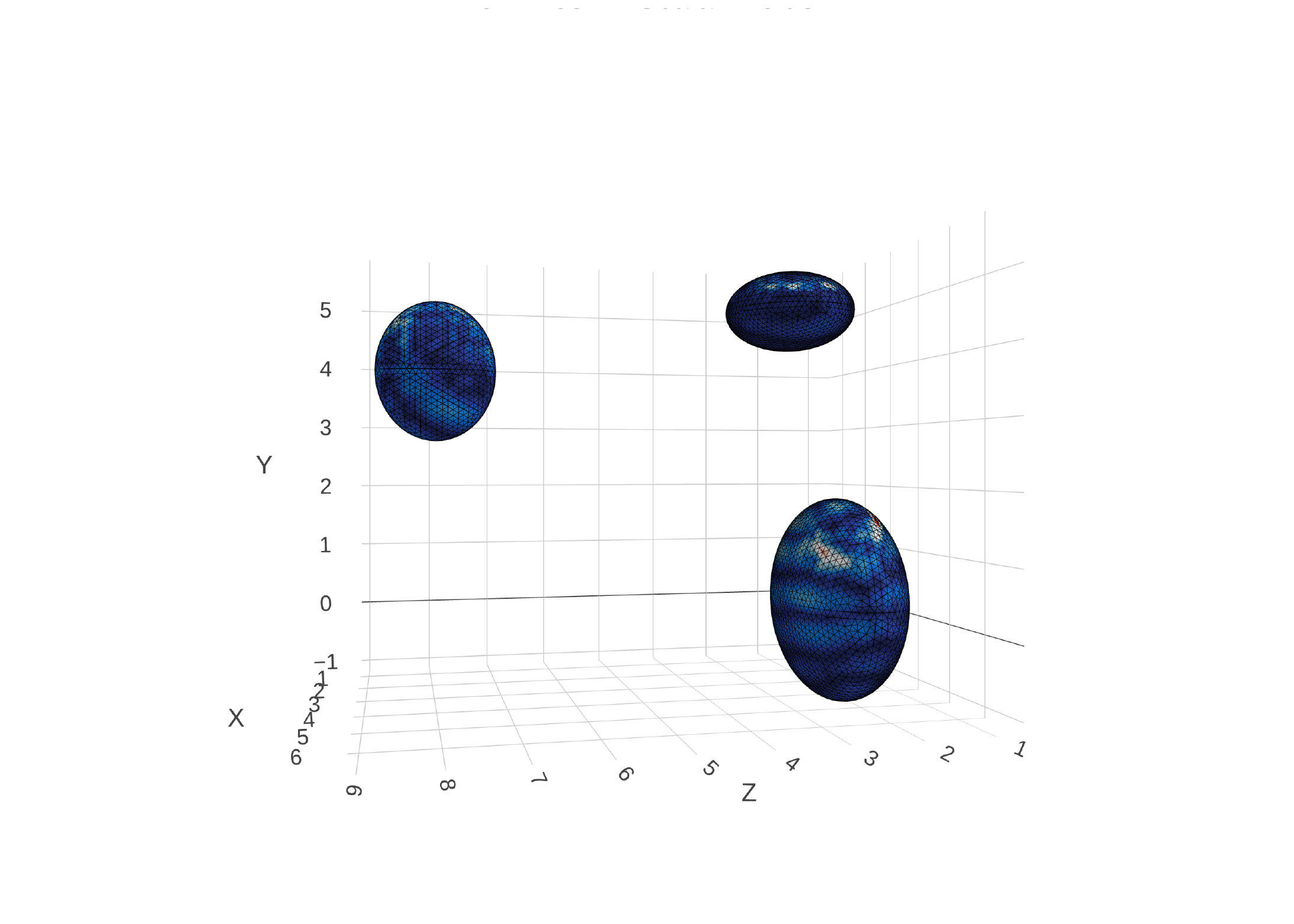}
        \end{subfigure} \\[12pt]

        \raisebox{1.5cm}{\rotatebox{90}{\textbf{ScaGNN}}} &
        \begin{subfigure}[b]{0.38\textwidth}
            \centering
            \includegraphics[width=\linewidth,trim={3cm 1cm 4.5cm 5cm},clip]{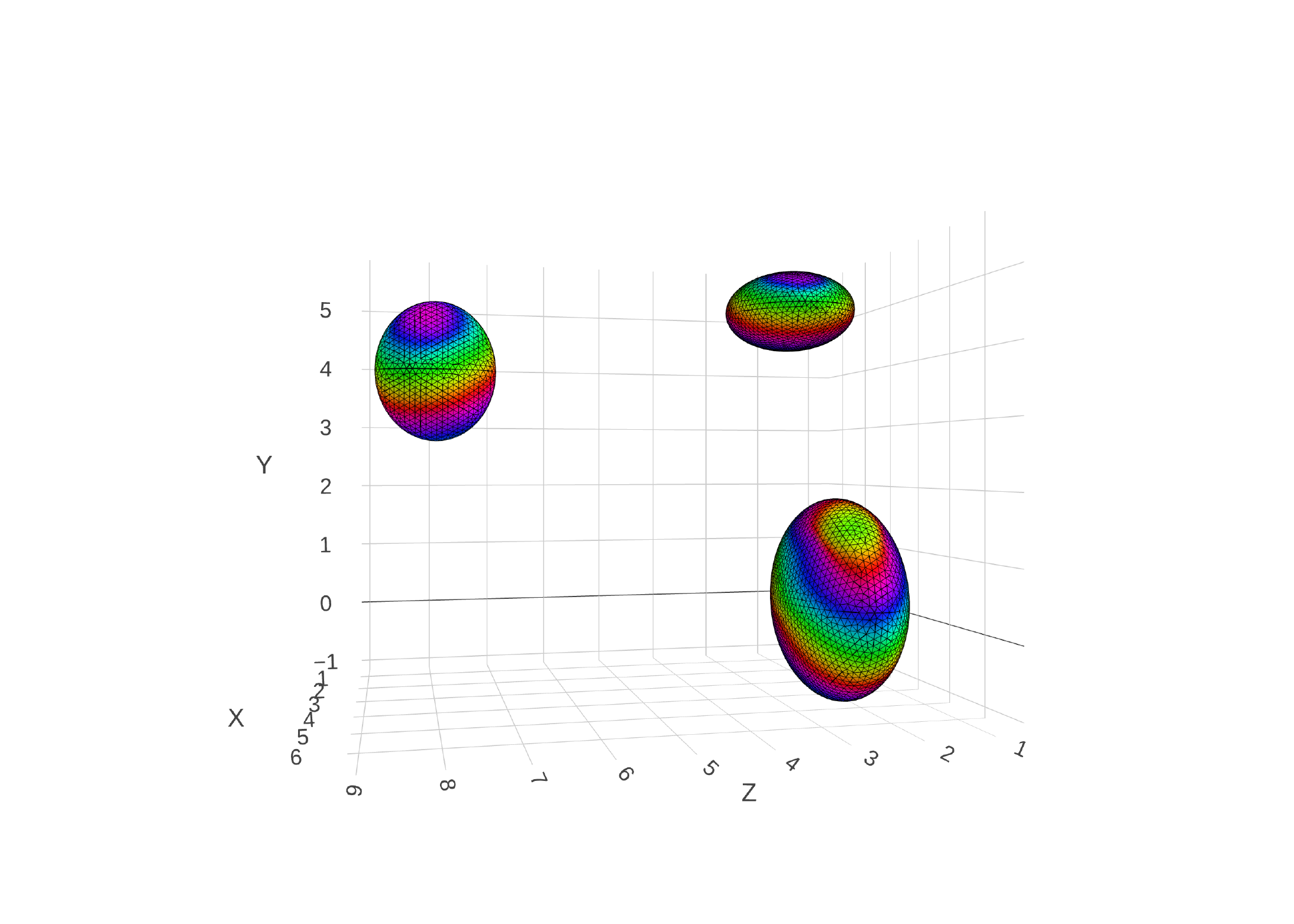}
        \end{subfigure} &
        \begin{subfigure}[b]{0.38\textwidth}
            \centering
            \includegraphics[width=\linewidth,trim={3cm 1cm 4.5cm 5cm},clip]{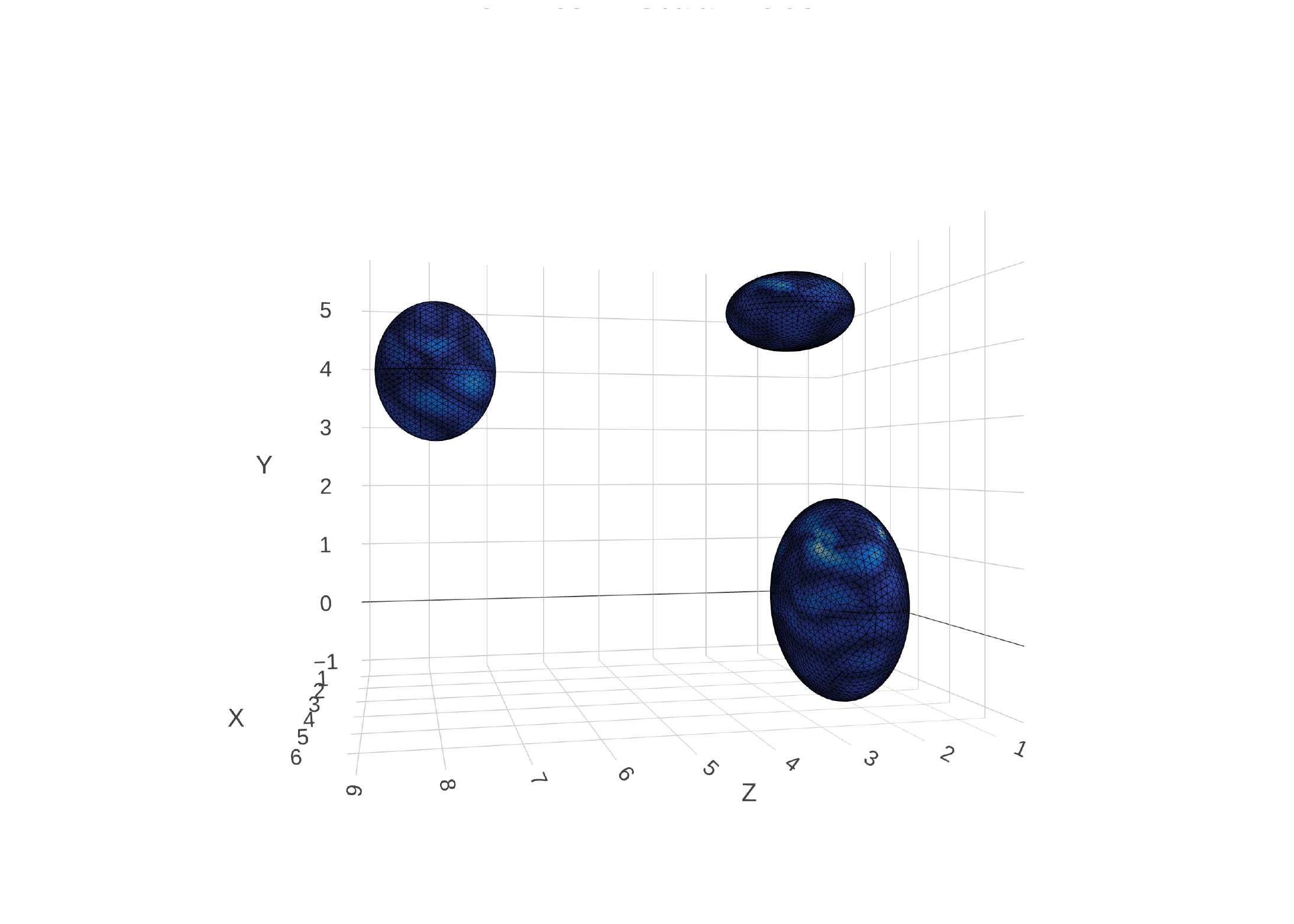}
        \end{subfigure} \\[12pt]

        \raisebox{1.2cm}{\rotatebox{90}{\textbf{Groundtruth}}} &
        \begin{subfigure}[b]{0.38\textwidth}
            \centering
            \includegraphics[width=\linewidth,trim={3cm 1cm 4.5cm 5cm},clip]{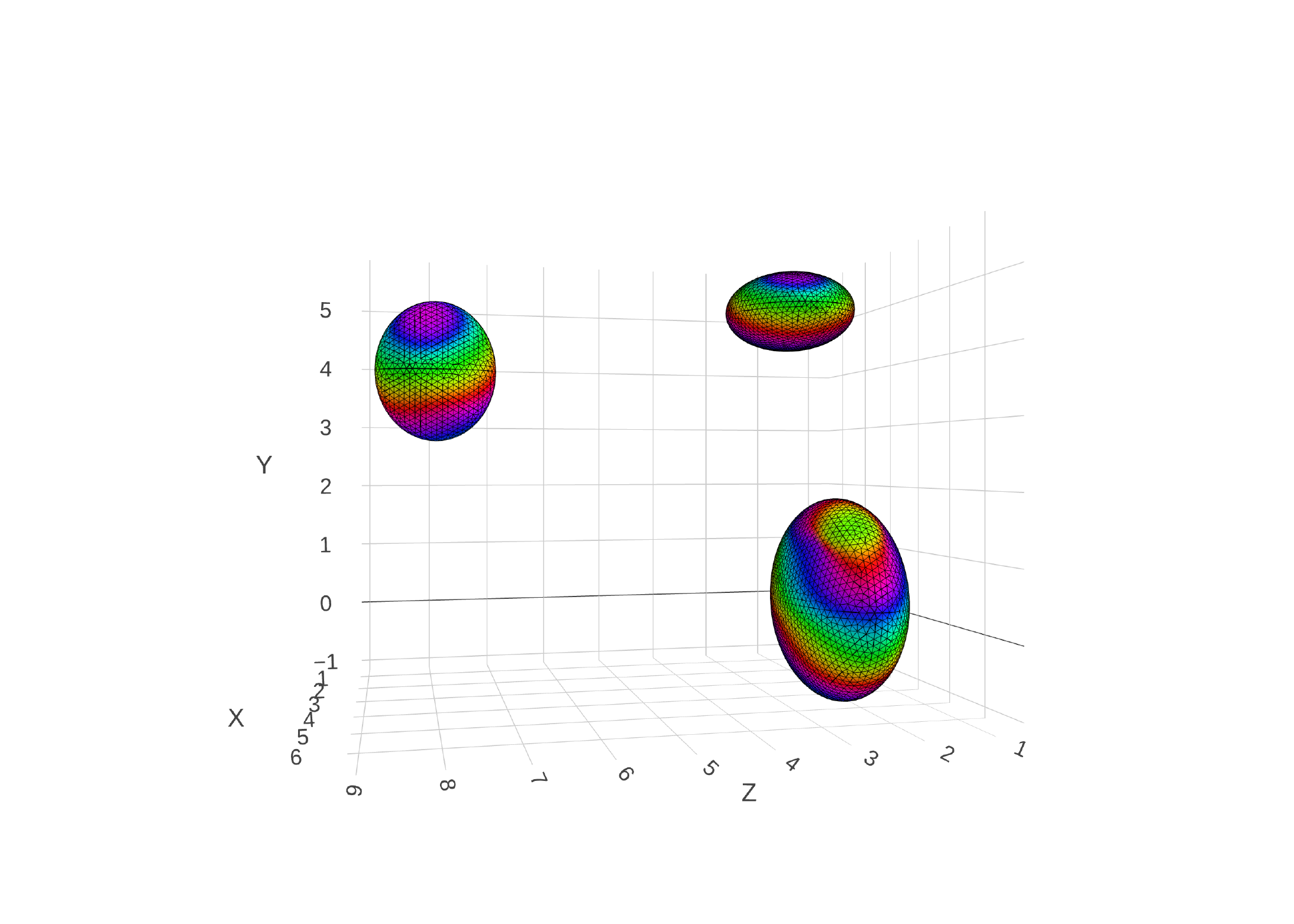}
        \end{subfigure} &
        \begin{subfigure}[b]{0.38\textwidth}
            \centering
            \hspace*{0.8cm}\includegraphics[width=\linewidth,trim={5cm 5cm 0cm 0cm},clip]{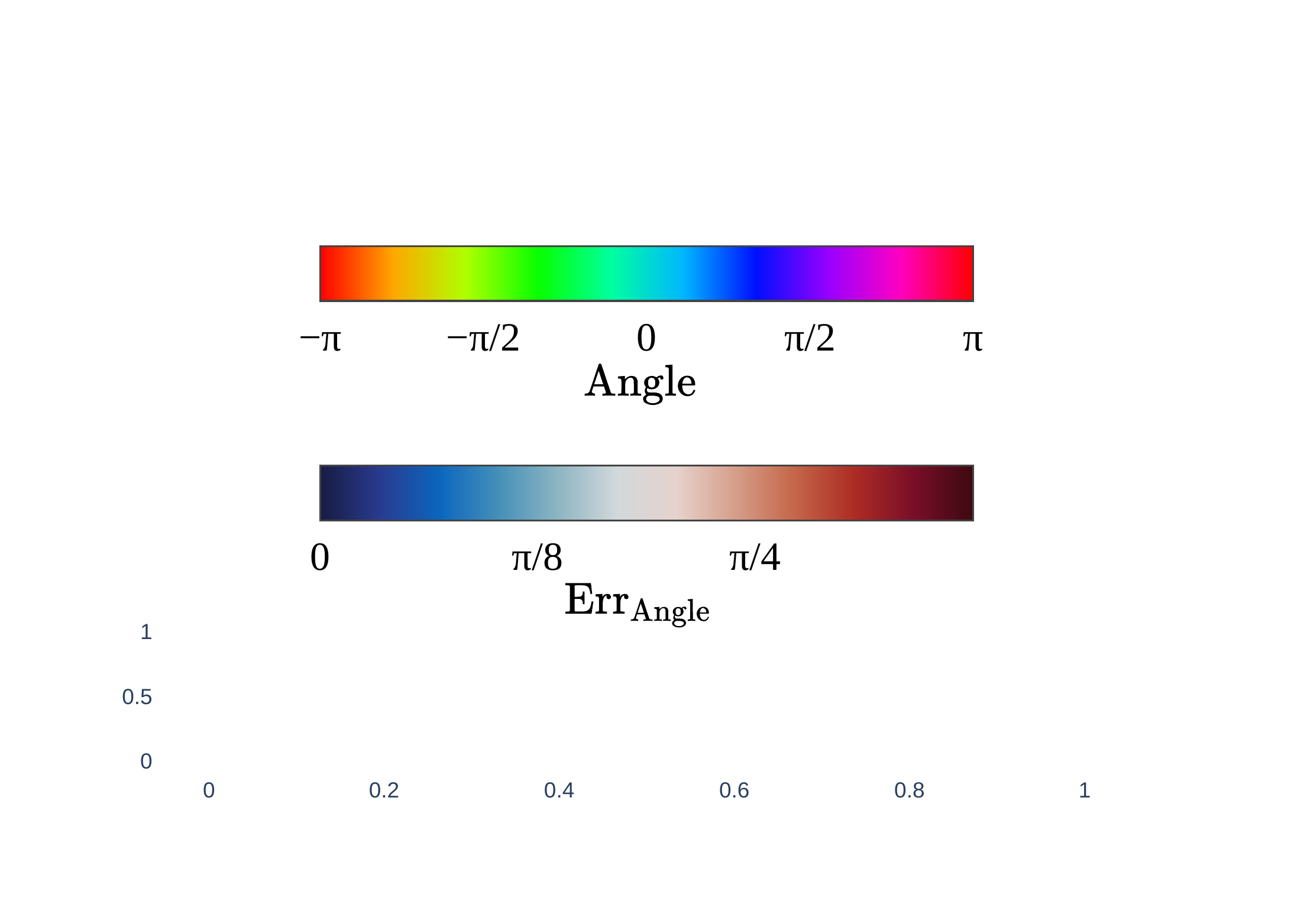}
        \end{subfigure}
    \end{tabular}
    \caption{Qualitative results for angle predictions on the Helmholtz Neumann problem with Point Transformer v3 \citep{wu2024point} and ScaGNN.}
    \label{fig:quali_helmholtz_neumann_angle}
\end{figure}

\cref{fig:quali_laplace_dirichlet,fig:quali_helmholtz_dirichlet_ampl,fig:quali_helmholtz_dirichlet_angle,fig:quali_helmholtz_neumann_ampl,fig:quali_helmholtz_neumann_angle} show qualitative results of the boundary solution for the Laplace Dirichlet, the Helmholtz Dirichlet and the Helmholtz Neumann problems with our ScaGNN approach and Point Transformer v3 \citep{wu2024point}, the best baseline.

\newpage
\subsection{Volumetric solution}

\begin{figure*}[ht]
    \includegraphics[width=\textwidth]{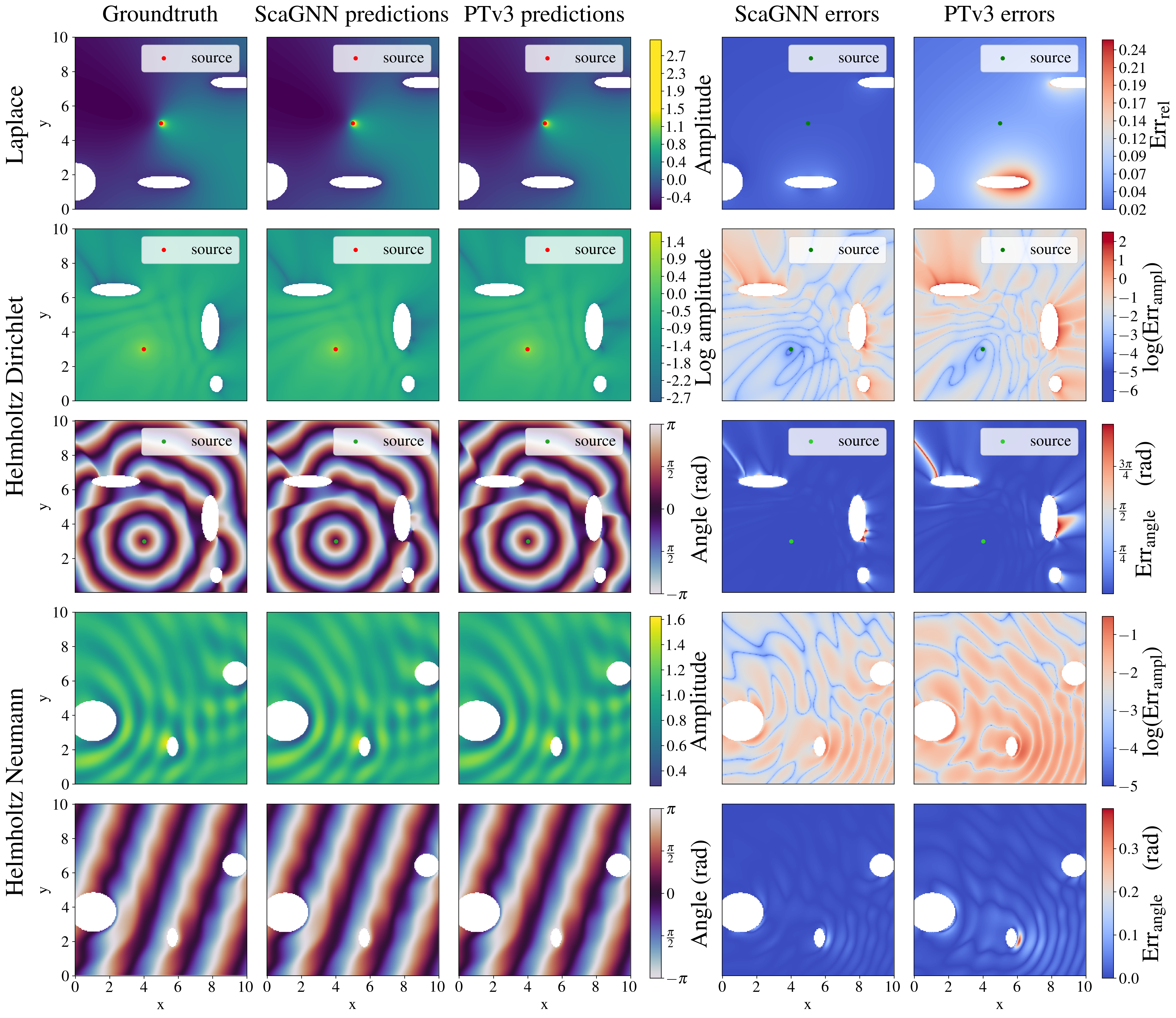}
    \caption{From left to right are the volumetric solutions of the total field obtained with GMRES (the groundtruth), with ScaGNN and with PTv3 for estimating the trace solution on the boundary, respectively, and the corresponding errors relative to the groundtruth with ScaGNN and PTv3, respectively. For each problem, the volumetric solutions of the total field and their associated errors are sampled within a square domain of side length 10 on the plane $z=0$, and the obstacles are represented in white.}
    \label{fig:quali}
\end{figure*}

In \cref{fig:quali}, we provide qualitative results relative to the volumetric solution of the total field, which is the sum of the incident and the scattered field, for both ScaGNN and the best baseline, Point Transformer v3 \citep{wu2024point}.

\end{document}